\documentclass[journal]{IEEEtran}

\usepackage{xcolor}
\usepackage{amsmath}
\usepackage{amssymb}
\usepackage{graphicx}
\usepackage{mathtools}
\usepackage{nicefrac} 
\usepackage[algo2e]{algorithm2e} 
\usepackage{algorithmic}
\usepackage{algorithm}
\usepackage{booktabs} 

\newtheorem{Theorem}{Theorem}

\newtheorem{Lemma}{Lemma}

\newtheorem{Prop}{Proposition}

\usepackage{stackengine}
\usepackage{caption}
\usepackage{subfigure}

\begin{document}

\title{Low-rank Characteristic Tensor Density Estimation Part I: Foundations}

\author{Magda~Amiridi,
        Nikos~Kargas,
        and~Nicholas D. Sidiropoulos,~\IEEEmembership{Fellow,~IEEE}
\thanks{
M. Amiridi and N.D. Sidiropoulos are with the Department of ECE, University of Virginia, Charlottesville, VA 22904. Author e-mails: (ma7bx,nikos)@virginia.edu. N. Kargas was with the Department of ECE, University of Minnesota; he is now with Amazon, Cambridge, U.K. Author e-mail: karga005@umn.edu}}

\maketitle

\begin{abstract}
Effective non-parametric density estimation is a key challenge in high-dimensional multivariate data analysis. In this paper, we propose a novel approach that builds upon tensor factorization tools. Any multivariate density can be represented by its characteristic function, via the Fourier transform. If the sought density is compactly supported, then its characteristic function can be approximated, within controllable error, by a finite tensor of leading Fourier coefficients, whose size depends on the smoothness of the underlying density. This tensor can be naturally estimated from observed and possibly incomplete realizations of the random vector of interest, via sample averaging. In order to circumvent the curse of dimensionality, we introduce a low-rank model of this {\em characteristic tensor}, which significantly improves the density estimate especially for high-dimensional data and/or in the sample-starved regime. By virtue of uniqueness of low-rank tensor decomposition, under certain conditions, our method enables learning the true data-generating distribution. We demonstrate the very promising performance of the proposed method using several toy, measured, and image datasets. 
\end{abstract}


\IEEEpeerreviewmaketitle

\section{Introduction}
\label{intro}
Density estimation is a fundamental yet challenging problem in data analysis and machine learning. Density estimation is the task of learning the joint Probability Density Function (PDF) from a set of observed data points, sampled from an unknown underlying data-generating distribution. A model of the density function of a continuous random vector provides a complete description of the joint statistical properties of the data and can be used to perform tasks such as computing the most likely value of a subset of elements (``features'') conditioned on others, computing any marginal or conditional distribution, and deriving optimal estimators, such as the minimum mean squared error (conditional mean) estimator. Density estimation has a wide range of applications including classification \cite{schmah2009generative,ray2017infectious}, clustering~\cite{ester1996density}, data synthesis~\cite{zen2014deep}, data completion~\cite{titterington1989imputation} and reconstruction related applications~\cite{balle2015density}, as well as learning statistical regularities such as skewness, tail behavior, multi-modality or other structures present in the data~\cite{silverman2018density}. 

Existing work on density estimation can be mainly categorized into parametric approaches such as Gaussian Mixture Models (GMM) \cite{pearson1894contributions}, and non-parametric approaches such as Histogram~\cite{rosenblatt1956} and Kernel Density Estimation (KDE)~\cite{parzen1962estimation}. A density model must be expressive -- flexible enough to represent a wide class of distributions, and tractable and scalable (computationally and memory-wise) at the same time (expressivity-tractability trade-off). Over the last several years, explicit feed-forward neural network based density estimation methods~\cite{germain2015made, uria2016neural,papamakarios2017masked} have gained increasing attention as they provide a tractable way to evaluate high-dimensional densities point-wise. On the other hand implicit {\em generative} models such as generative adversarial networks \cite{goodfellow2014generative} and variational autoencoders \cite{kingma2013auto} can be used to obtain models which allow effective and efficient sampling. 

In this paper, we develop a novel non-parametric method for multivariate PDF estimation based on tensor rank decomposition -- known as {\em Canonical Polyadic Decomposition} (CPD)~\cite{hitchcock1927expression,harshman1970foundations}. CPD is a powerful model that can parsimoniously represent high-order data tensors exactly or approximately, and its distinguishing feature is that under certain reasonable conditions it is unique -- see~\cite{sidiropoulos2017tensor} for a recent tutorial overview. We show that any compactly supported continuous density can be approximated, within controllable error, by a finite {\em characteristic tensor} of leading complex Fourier coefficients, whose size depends on the smoothness of the density. This  characteristic tensor can be naturally estimated via sample averaging from realizations of the random vector of interest. 

The main challenge, however, lies in the fact that the size of this tensor (the number of model parameters in the Fourier domain) grows exponentially with the number of random variables -- the length of the random vector of interest. In order to circumvent this ``curse of dimensionality'' (CoD) and further denoise the naive sample averaging estimates, we introduce a low-rank model of the characteristic tensor, whose degrees of freedom (for fixed rank) grow linearly with the random vector dimension. Low-rank modeling significantly improves the density estimate especially for high-dimensional data and/or in the sample-starved regime. By virtue of uniqueness of low-rank tensor decomposition, under certain conditions, our method enables learning the true data-generating distribution.

In order to handle incomplete data (vector realizations with missing entries) as well as scaling up to high-dimensional vectors, we further introduce coupled low-rank decomposition of lower-order characteristic tensors corresponding to smaller subsets of variables that share `anchor' variables, and show that this still enables recovery of the global density, under certain conditions. As an added benefit, our approach yields a generative model of the sought density, from which it is very easy to sample from. This is because our low-rank model of the characteristic tensor admits a latent variable naive Bayes interpretation. A corresponding result for finite-alphabet random vectors was first pointed out in~\cite{Kargas2018Kolmogorov}. In contrast to~\cite{Kargas2018Kolmogorov}, our approach applies to continuous random vectors possessing a compactly supported multivariate density function. From an algorithmic standpoint, we formulate a constrained coupled tensor factorization problem and develop a Block Coordinate Descent (BCD) algorithm. 

The main results and  contributions of this paper can be summarized as follows:
\begin{itemize}
\item We show that any smooth compactly supported multivariate PDF can be approximated by a finite tensor model, without using any prior or data-driven discretization process. We also show that truncating the sampled multivariate characteristic function of a random vector is equivalent to using a finite separable mixture model for the underlying distribution. Under these relatively mild assumptions, the proposed model can approximate any high dimensional PDF with approximation guarantees.By virtue of  uniqueness of CPD, assuming low-rank in the Fourier domain, the underlying multivariate density is identifiable. 

\item We show that high dimensional joint PDF recovery is possible under low tensor-rank conditions, even if we only observe subsets (triples) of variables. This is a key point that enables one to handle incomplete realizations of the random vector of interest. To the best of our knowledge, no other generic density estimation approach allows this. To tackle this more challenging version of the problem, we propose an optimization framework based on coupled tensor factorization. Our approach jointly learns lower-order ($3$-dimensional) characteristic functions, and then assembles tensor factors to synthesize the full characteristic function model.

\item The proposed model allows efficient and low-complexity inference, sampling, and density evaluation. In that sense, it a more comprehensive solution that neural density evaluation or neural generative models. We provide convincing experimental results on toy, image, and measured datasets that corroborate the effectiveness of the proposed method.
\end{itemize}

This is the first of a two-part paper. The second part builds on this foundation to develop a joint compression (nonlinear dimensionality reduction) and compressed density estimation framework that offers additional flexibility and scalability, but does not provide a density estimate in the original space as the ``baseline'' method in this first part does. Each approach has its own advantages, but the second builds upon the first. It is therefore natural to present them as Part I and Part II.   

\section{Background}
\subsection{Related work}
\label{sec:related_work}
Density estimation has been the subject of extensive research in statistics and the machine learning community. Methods for density estimation can broadly be classified as either parametric or non-parametric. Parametric density estimation assumes that the data are drawn from a known parametric family of distributions, parametrized by a fixed number of tunable parameters. Parameter estimation is usually performed by maximizing the likelihood of the observed data. One of the most widely used parametric models is the Gaussian Mixture Model (GMM). GMMs can approximate any density function if the number of components is large enough~\cite{mclachlan1988mixture}. 
However, a very large number of components may be required for good approximation of the unknown density, especially in high dimensions. Increasing the number of components introduces computational challenges and may require a large amount of data~\cite{chen1995optimal}. Misspecification and inconsistent estimation is less likely to occur with nonparametric density estimation. 

Nonparametric density estimation is more unassuming and in that sense ``universal'', but the flip-side is that it does not scale beyond a small number of variables (dimensions). The most widely-used approach for nonparametric density estimation is Kernel Density Estimation (KDE)~\cite{rosenblatt1956, parzen1962estimation}. The key idea of KDE is to estimate the density by means of a sum of kernel functions centered at the given observations. However, worst-case theoretical results show that its performance worsens exponentially with the dimension of the data vector~\cite{scott1991feasibility}. 

Our approach falls under nonparametric methods, and is motivated by Orthogonal Series Density Estimation (OSDE) \cite{girolami2002orthogonal, efromovich2010orthogonal, tsybakov2008introduction}, a powerful non-parametric estimation methodology. OSDE approximates a probability density function using a truncated sum of orthonormal basis functions, which may be trigonometric, polynomial, wavelet etc. However, OSDE becomes computationally intractable even for as few as $10$ dimensions, since the number of parameters grows exponentially with the number of dimensions. Unlike OSDE, our approach is able to scale to much higher dimensions.

Recently, several density evaluation and modeling methods that rely on neural networks have been proposed. The Real-valued Neural Autoregressive Distribution Estimator (RNADE)~\cite{uria2013rnade} is among the best-performing neural density estimators and has shown great potential in scaling to high-dimensional distribution settings. These so-called autoregressive models (not to be confused with classical AR models for time-series) decompose the joint density as a product of one-dimensional conditionals of increasing conditioning order, and model each conditional density with a parametric model. Normalizing Flows (NF)~\cite{pmlr-v37-rezende15} models start with a base density e.g., standard Gaussian, and stack a series of invertible transformations with tractable Jacobian to approximate the target density. Masked Autoregressive Flow (MAF)~\cite{papamakarios2017masked} is a type of NF model, where the transformation layer is built as an autoregressive neural network. These methods do not construct a joint PDF model but rather serve for point-wise density {\em evaluation}. That is, for any given input vector (realization), they output an estimate of the density evaluated at that particular input vector (point). For small vector dimensions, e.g., two or three, it is possible to evaluate all inputs on a dense grid, thereby obtaining a histogram-like density estimate; but the curse of dimensionality kicks in for high vector dimensions, where this is no longer an option. Additionally, these methods cannot impute more than very few missing elements in the input, for the same reason (grid search becomes combinatorial).

\subsection{Notation}
In this section we briefly present notation conventions and some tensor algebra preliminaries. We use the symbols $\mathbf{x}$, $\mathbf{X}$, $\underline{\mathbf{X}}$ for vectors, matrices and tensors respectively. We use the notation $\mathbf{x}(n)$, $\mathbf{X}(:,n)$, ${\underline{\mathbf{X}}(:,:,n)}$ to refer to a particular element of a vector, a column of a matrix and a slab of a tensor. Symbols $\|\mathbf{x}\|_2$, $\|\mathbf{X}\|_F$, and $\|\mathbf{x}\|_{\infty}$ correspond to $L_2$ norm, Frobenius norm, and infinity norm.
Symbols $\circ$, $\circledast$, $\odot$  denote the outer, Hadamard and Khatri-Rao product respectively. The vectorization operator is denoted as $\textrm{vec}(\mathbf{X})$, $\textrm{vec}(\underline{\mathbf{X}})$ for a matrix and tensor respectively. Additionally, ${\rm diag}(\mathbf{x})\in \mathbb{C}^{K \times K}$ denotes the diagonal matrix with the elements of vector $\mathbf{x}\in \mathbb{C}^K$ on its diagonal. The set of integers $\{1,\ldots, K\}$ is denoted as $[K]$. 

\subsection{Relevant tensor algebra}
\label{sec:tensors}
\begin{figure}
\begin{center}
\includegraphics[width=80mm]{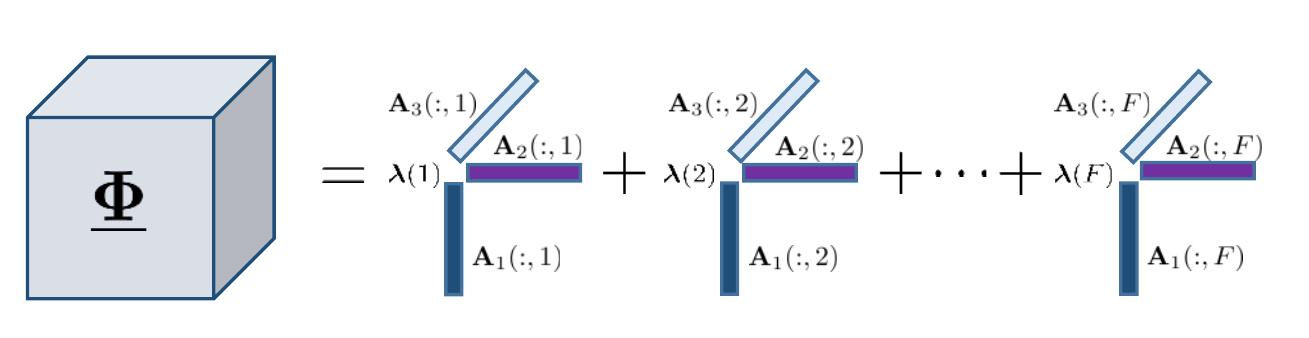}
\caption{CPD model of a 3-way tensor.}
\label{CPD}
\end{center}
\end{figure}

 An $N$-way tensor ${\underline{\boldsymbol{\Phi}} \in \mathbb{C}^{K_1 \times K_2 \times \cdots \times K_N}}$ is a multidimensional array whose elements are indexed by $N$ indices. Any tensor can be decomposed as a sum of $F$ rank-$1$ tensors
 \begin{equation}
 \underline{\boldsymbol{\Phi}} = \sum_{h=1}^F{ \boldsymbol{\lambda}}(h) \mathbf{A}_1(:,h) \circ \mathbf{A}_2(:,h) \circ \cdots \circ \mathbf{A}_N(:,h),
\end{equation}
where $\mathbf {A}_n \in \mathbb{C}^{K_{n}\times F}$ and constraining the columns ${\mathbf{A}_n(:,h)}$ to have unit norm, the real scalar ${\boldsymbol{\lambda}}(h)$ absorbs the $h$-th rank-one tensor’s scaling. A visualization is shown in Figure~\ref{CPD} for the case of $N=3$. 

We use ${\underline{\boldsymbol{\Phi}} = [\![\boldsymbol{\lambda}, \mathbf{A}_1,\ldots,\mathbf{A}_N]\!]}$ to denote the decomposition. When $F$ is minimal, it is called the rank of tensor $\underline{\boldsymbol{\Phi}}$, and the decomposition is called {\em Canonical} Polyadic Decomposition (CPD). A particular element of the tensor is given by $\underline{\boldsymbol{\Phi}}(k_1,k_2,\ldots,k_N) = \sum_{h=1}^F \boldsymbol{\lambda}(h) \prod_{n=1}^N \mathbf{A}_n(k_n,h)$. The vectorized form of the tensor can be expressed as $\textrm{vec}(\underline{\boldsymbol{\Phi}})= \left(\odot_{n=1}^N \mathbf{A}_n\right)\boldsymbol{\lambda}$. We can express the mode-$n$ matrix unfolding which is a concatenation of all mode-$n$ `fibers' of the tensor as  ${\underline{\boldsymbol{\Phi}}^{(n)} = (\odot_{k \neq n} \mathbf{A}_k) {\rm diag}(\boldsymbol{\lambda}){\mathbf{A}_n}^T}$,
where ${(\odot_{k \neq n} \mathbf{A}_k)= \mathbf{A}_N \odot \cdots \odot \mathbf{A}_{n+1} \odot \mathbf{A}_{n-1} \odot \cdots \odot \mathbf{A}_1}$. 

A key property of the CPD  is that the rank-1 components are unique under mild conditions. For learning probabilistic latent variable models and latent representations, the uniqueness of tensor decomposition can be interpreted as model identifiability. A model is identifiable, if and only iff there is a unique set of parameters that is consistent with what we have observed. 
\begin{Theorem}
\label{thm:tensor_ident_1}~\cite{sidiropoulos2000uniqueness}: 
{Let $k_\mathbf{A}$ be the Kruskal rank of $\mathbf{A}$, defined as the largest integer $k$ such that every $k$ columns of $\mathbf{A}$ are linearly independent. Given  $\underline{\boldsymbol{\Phi}}=[\![\boldsymbol{\lambda},\mathbf{A}_1,\ldots,\mathbf{A}_N]\!]$, if $\sum_{n=1}^N k_{\mathbf{A}_n}\geq2F+N-1$, then the rank of $\underline{\boldsymbol{\Phi}}$ is $F$ and the decomposition of $\underline{\boldsymbol{\Phi}}$ in rank-one terms is unique.} 
\end{Theorem}
Better results allowing for higher tensor rank are available for generic tensors of given rank.
\begin{Theorem}\label{thm:tensor_ident_2} \cite{chiantini2012generic}: Given $\underline{\boldsymbol{\Phi}}~=[\![\boldsymbol{\lambda},\mathbf{A}_1,\mathbf{A}_2,\mathbf{A}_3]\!]$, assume, without loss of generality, that $I_1\leq I_2\leq I_3$. Let $\alpha,\beta$ be the largest integers such that $2^{\alpha}\leq I_1$ and $2^{\beta}\leq I_2$. If $F\leq2^{\alpha+\beta-2}$ the decomposition of $\underline{\boldsymbol{\Phi}}$ in rank-one terms is unique almost surely.
\end{Theorem}

\section{A Characteristic Function Approach}
The characteristic function of a random variable $X$ is the Fourier transform of its PDF, and it conveys all information about $X$. The characteristic function can be interpreted as an expectation: the Fourier transform at frequency $\nu \in \mathbb{R}$ is $E\left[e^{j {\nu} X}\right]$. Similarly, the multivariate characteristic function is the multidimensional Fourier transform of the density of a random vector ${\boldsymbol X}$, which can again be interpreted as the expectation $E[e^{j {\boldsymbol \nu}^T {\boldsymbol X}}]$, where ${\boldsymbol \nu}$ is a vector of frequency variables. The expectation interpretation is crucial, because ensemble averages can be estimated via sample averages; and whereas direct nonparametric density estimation at point $x$ requires samples around $x$, estimating the characteristic function enables reusing all samples globally, thus enabling better sample averaging and generalization. This point is the first key to our approach. The difficulty, however, is that pinning down the characteristic function seemingly requires estimating an uncountable set of parameters. We need to reduce this to a {\em finite} parameterization with controllable error, and ultimately distill a parsimonious model that can learn from limited data and still generalize well. In order to construct an accurate joint distribution estimate that is scalable to high dimensions without making explicit and restrictive prior assumptions (such as a GMM model) on the nature of the density, and without requiring huge amounts of data, we encode the following key ingredients into our model.
\begin{itemize}
\item \textbf{Compactness of support}. In most cases, the random variables of interest are bounded, and these bounds are known or can be relatively easily estimated.  
\item \textbf{Continuity of the underlying density and its derivatives}. The joint distribution is assumed to be sufficiently smooth in some sense, which enables the use of explicit or implicit interpolation.
\item \textbf{Low-rank tensor modeling}.
We show that joint characteristic functions can be represented as higher order tensors. In practice these tensor data are not unstructured. Low-rank tensor modeling provides a concise  representation that captures the salient characteristics (the {\em principal components}) of the data distribution in the Fourier domain. 

\end{itemize}
\subsection{The Univariate Case}
\label{sec:1_d}

Before we delve into the multivariate setting, it is instructive to examine the univariate case. Given a real-valued random variable $X$ with compact support $S_X$, the Probability Density Function (PDF) $f_{X}$ and its corresponding Characteristic Function (CF) $\Phi_{X}$ form a Fourier transform pair:
\begin{align}
\Phi_{X}(\nu) &:=\int_{S_X}f_{X}(x)e^{j\nu x}dx  = E[e^{j\nu X}], \\ 
f_{X}(x)&:=\frac{1}{2\pi}\int_{-\infty}^{\infty}\Phi_{X}(\nu)e^{-j\nu x}d\nu.
\end{align}
Note that $\Phi_{X}(0)=\int_{-\infty}^{\infty}f_{X}(x)d x=1$. Without loss of generality, we can apply range normalization and mean shifting so that $sX+c \in [0,1]$ -- the transformation is invertible. We may therefore assume that $S_X = [0,1]$. Every PDF supported in $[0,1]$ can be uniquely represented over its support by an infinite Fourier series,
\begin{equation}
f_{X}(x) = {\sum_{k=-\infty}^\infty}\Phi_{X}[k]e^{-j2\pi k x},
\end{equation}
where $ \Phi_{X}[k]={\Phi_{X}(\nu)}\big\rvert_{\nu=2\pi k}, \quad k\in\mathbb{Z}$. This shows that {\em countable} parameterization through samples of the characteristic function suffices for compactly supported densities. But this is still not enough - we need a finite parametrization.  Thankfully, if $f_{X}$ is sufficiently differentiable in the sense that ${f_{X} \in C^p}$ i.e.,  all its derivatives $\frac{\partial f_{X}}{\partial x}, \frac{\partial^2 f_{X}}{\partial x^2}, \cdots, \frac{\partial^p f_{X}}{\partial x^p}$ exist and are continuous we have that
\newline
\begin{Lemma}
\label{lem:Plonka1D}~(e.g., see \cite{plonka2018numerical}): If ${f_{X} \in C^p}$, then 
\begin{align*}
    |\Phi_{X}[k]|=~\mathcal{O}\left(\frac{1}{1+|k|^{p}}\right).
\end{align*} 
\end{Lemma}
\begin{figure}
\begin{center}
\includegraphics[width=0.52 \columnwidth]{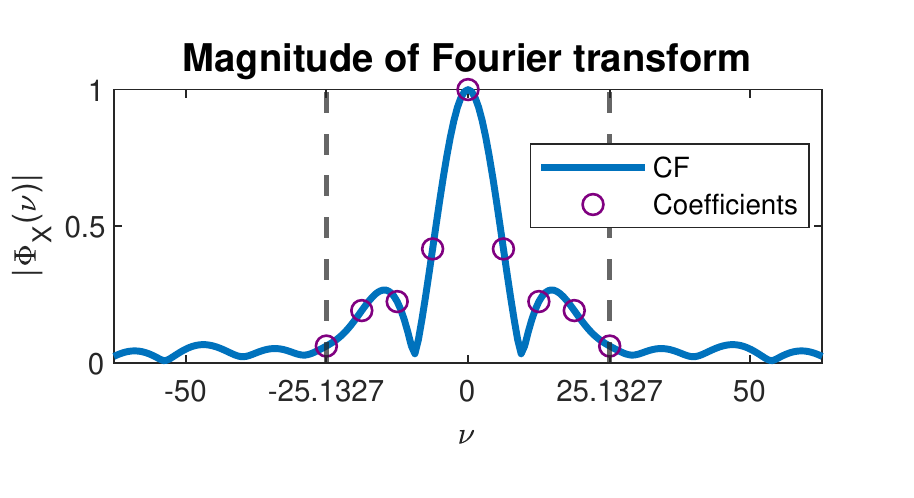}
\includegraphics[width=0.52 \columnwidth]{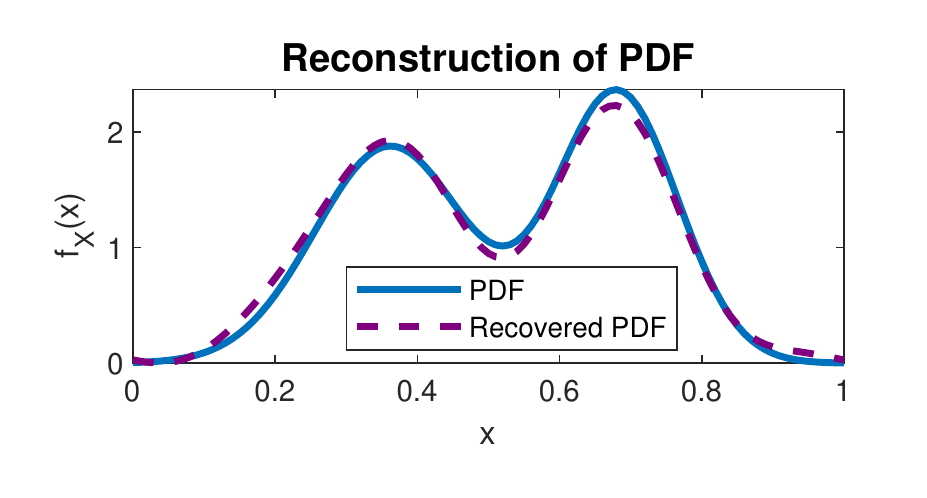}
\caption{Illustration of the key idea on a univariate Gaussian mixture. The PDF can be (approximately) recovered from only 9 uniform samples of its CF.}
\label{fig:toy_exaple_1}
\end{center}
\end{figure}
It is therefore possible to use a truncated series 
\begin{align*}
\widehat{f}_{X}(x) = {\sum_{k=-K}^K}\Phi_{X}[k]e^{-j2\pi k x},
\end{align*} 
with proper choice of $K$ that will not incur significant error. Invoking Parseval's Theorem 
\begin{align*}
{\|f_X - \widehat{f}_X \|_2^2={\sum_{|k|>K}}{|\Phi_{X}[k]|}^2},
\end{align*} 
which is controllable by the {\em smoothing parameter} $K$. The $k$-th Fourier coefficient
\begin{align*}
{\Phi_{X}[k] = \int_{0}^{1}f_{X}(x)e^{j2\pi k x}d x =E[e^{j2\pi k X}]}
\end{align*} 
can be conveniently estimated via the sample mean
\begin{align*}
\widehat{\Phi}_{X}[k]=\frac{1}{M}\sum\limits_{m=1}^M{e^{j2\pi k x_m}}
\end{align*} 
Here $M$ is the number of available realizations of the random variable $X$.

A toy example to illustrate the idea is shown in Figure~\ref{fig:toy_exaple_1}. For this example, we are given $M=500$ realizations of a random variable $X$, which is a mixture of two Gaussian distributions with means ${\mu}_1=0.35$,  ${\mu}_2=0.7$ and standard deviations ${\sigma}_1=0.1$, ${\sigma}_2=0.08$. The recovered PDF is very close to the true PDF using only $9$ coefficients of the CF.
\subsection{The Multivariate Case}\label{sec:universal}

In the multivariate case, we are interested in obtaining an estimate ${\widehat{f}}_{{\boldsymbol X}}$ of the true density ${f}_{{\boldsymbol X}}$ of a random vector ${\boldsymbol X}:= \left[X_1,\ldots, X_N\right]^T$. The {\em joint} or {\em multivariate characteristic function} of ${\boldsymbol X}$ is a function ${\Phi}_{\boldsymbol X}:{\mathbb{R}}^N\rightarrow{\mathbb{C}}$ defined as 
\begin{equation}
{\Phi}_{{\boldsymbol X}}({\boldsymbol \nu}):=
E\left[e^{j {\boldsymbol \nu}^T {\boldsymbol X}}\right], 
\end{equation}
where ${\boldsymbol \nu} := \left[\nu_1,\ldots,\nu_N\right]^T.$
For any given ${\boldsymbol\nu}$, given a set of realizations $\left\{ {\mathbf x}_m \right\}_{m=1}^M$, we can estimate the empirical characteristic function of the sequence as 
\begin{equation}
{\widehat{\Phi}}_{{\boldsymbol X}}({\boldsymbol \nu}) =
\frac{1}{M}\sum_{m=1}^M e^{j {\boldsymbol \nu}^T{{\mathbf x}_{m}}}.
\end{equation}

Under mixing conditions such that sample averages converge to ensemble averages, the corresponding PDF can be uniquely recovered via the multidimensional inverse Fourier transform
\begin{equation}
f_{{\boldsymbol X}}({\boldsymbol{x}})=\frac{1}{({2\pi})^N}\int_{\mathbb{R}^N}{{\Phi}_{{\boldsymbol X}}({\boldsymbol \nu})}e^{-j{\boldsymbol \nu}^T {\boldsymbol x}} d{\boldsymbol \nu}.
\end{equation}
If the support of the joint PDF $f_{\boldsymbol X}({\mathbf x})$ is contained within the hypercube $S_{\boldsymbol X}=[0,1]^N$, then similar to the univariate case, it can be represented by a multivariate Fourier series
\begin{equation}
\begin{aligned}
f_{{\boldsymbol X}}({\boldsymbol{x}})= {\sum_{k_1=-\infty}^\infty} \cdots{\sum_{k_N=-\infty}^\infty}{\Phi}_{\boldsymbol X}[{\boldsymbol{k}}]e^{-j2\pi {\mathbf k}^T{\mathbf x}},\\ 
\text{where } {\Phi_{{\boldsymbol X}}}[{\mathbf{k}}] = \Phi_{\boldsymbol X}(\boldsymbol{\nu})\big\rvert_{{\boldsymbol{\nu}}=2\pi \mathbf{k}}, \mathbf{k} = [k_1,\ldots,k_N]^T.
\end{aligned}
\end{equation}
\smallskip
\begin{Lemma}\label{lem:PlonkaND} (see e.g., \cite{plonka2018numerical}): {\em For any $p \in \mathbb{N}$, if the partial derivatives $\frac{\partial^{\theta_1}}{\partial x_1^{\theta_1}} \cdots \frac{\partial^{\theta_N}}{\partial x_N^{\theta_N}} f_{{\boldsymbol X}}({\mathbf x})$ exist and are absolutely integrable for all $\theta_1,\ldots,\theta_N$ with ${\sum_{n=1}^N \theta_n \leq p}$ then the rate of decay of the magnitude of the ${\mathbf k}$-th Fourier coefficient $|{{\Phi}_{{\boldsymbol X}}}[{\mathbf k}]|$ obeys }$|{\Phi_{{\boldsymbol X}}}[{\mathbf{k}}]|=~\mathcal{O}{\bigg(}{{\frac{1}{1+\|\mathbf{k} \|_2^p}}}{\bigg)}.$
\end{Lemma}
\medskip
The smoother the underlying PDF, the faster its Fourier coefficients and the approximation error tend to zero. Thus we can view the joint PDF through the lens of functions with only low frequency harmonics. Specifically,
it is known~\cite{mason1980near},~\cite[Chapter~23]{handscomb2014methods} that the approximation error of the truncated series with absolute cutoffs $\left\{K_n\right\}_{n=1}^N$ is upper bounded by
\begin{equation}
\|f_{\boldsymbol X}- \widehat{f}_{{\boldsymbol X}} \|_{\infty}\leq C  \sum_{n=1}^N \frac{\omega_n \left(\frac{\partial^{\theta_n}}{ {\partial x_{n}^{\theta_n}}} f_{\boldsymbol X},\frac{1}{1+K_n} \right)}{{(1+K_n)}^{\theta_n}}, 
\end{equation}
where $\omega_n(f_{\boldsymbol X},\delta):=$
\[
\mathrel{\stackunder{$\displaystyle\text{sup}$}{
    \stackunder{$\scriptstyle {\left|x_j-x_j'\right|\leq \delta}$}}}
\left|f_{\boldsymbol X}(x_1,\ldots,x_j,\ldots,x_N)-f_{\boldsymbol X}(x_1,\ldots,x_j',\ldots,x_N)\right|,
\]
and 
\[C = C_2 \left(1+ C_1  \prod_{n=1}^N \log K_n\right).
\] 
$C_1,C_2$ are constants independent of $K_n$. The  smoother  the  underlying  PDF,  the  smaller  the  obtained  finite parametrization error. It follows that we can approximate $f_{\boldsymbol X}$
\begin{equation}
{\widehat{f}}_{{\boldsymbol X}}({\mathbf{x}})={\sum_{k=-{K_1}}^{K_1}}\cdots{\sum_{k_N=-K_N}^{K_N}}{\Phi}_{\boldsymbol X}[{\mathbf{k}}]e^{-j2\pi {\mathbf k}^T{\mathbf x}}.
\label{eq:truncated_mult}
\end{equation}
The truncated Fourier coefficients can be naturally represented using an $N$-way tensor $\underline{\boldsymbol{\Phi}}$ where
\begin{equation}
\underline{\boldsymbol{\Phi}}(k_1, \ldots, k_N) = \Phi_{\boldsymbol{X}}[\mathbf{k}].
\end{equation}

\section{Proposed Approach: Breaking the Curse of Dimensionality}
\label{sec:Proposed_app}
We have obtained a finite parameterization with controllable and bounded error, but the number of parameters ${(2K_1 + 1) \times \cdots \times (2K_N+1)}$  obtained by truncating ${\Phi}_{{\boldsymbol X}}$ as above grows exponentially with $N$. This curse of dimensionality can be circumvented by focusing on the {\em principal components} of the resulting tensor, i.e., introducing a low-rank parametrization of the {\em Characteristic Tensor} obtained by truncating the multidimensional Fourier series. Keeping the first $F$ principal components, the number of parameter reduces from order of $K_1 \times \cdots \times K_N$ to order of $(K_1 + \cdots + K_N)F$. Introducing the rank-$F$ CPD in Equation~\eqref{eq:truncated_mult}, one obtains the approximate model
\begin{multline}
{\tilde{f}}_{{\boldsymbol X}}({\mathbf x})=
{\sum_{k_1=-K}^{K}}\cdots{\sum_{k_N=-K}^{K}}\sum_{h=1}^F p_H(h)\prod_{n=1}^N\Phi_{X_n|H=h}[k_n]\\e^{-j2\pi k_n x_n},
\end{multline}
where $H$ can be interpreted as a latent ($H$ for `hidden') random variable, $\Phi_{X_n|H=h}[k_n]$ is the characteristic function of $X_n$ conditioned on $H=h$
\begin{align}
 \Phi_{X_n|H=h}[k_n]&:={\Phi}_{X_n|H=h}(\nu|h)\big\rvert_{\nu=2\pi k_n}
  \nonumber\\
  &=\mathbb{E}_{X_n|H=h}\left[e^{j2\pi k_n X_n}\right],
\end{align}
and we stress that for high-enough $F$, this representation is without loss of generality -- see, e.g.,~\cite{sidiropoulos2017tensor}. For the rest of the paper, we consider $K=K_1=\cdots=K_N$ for brevity. 

\begin{figure}
\begin{center}
\includegraphics[width=0.4 \columnwidth]{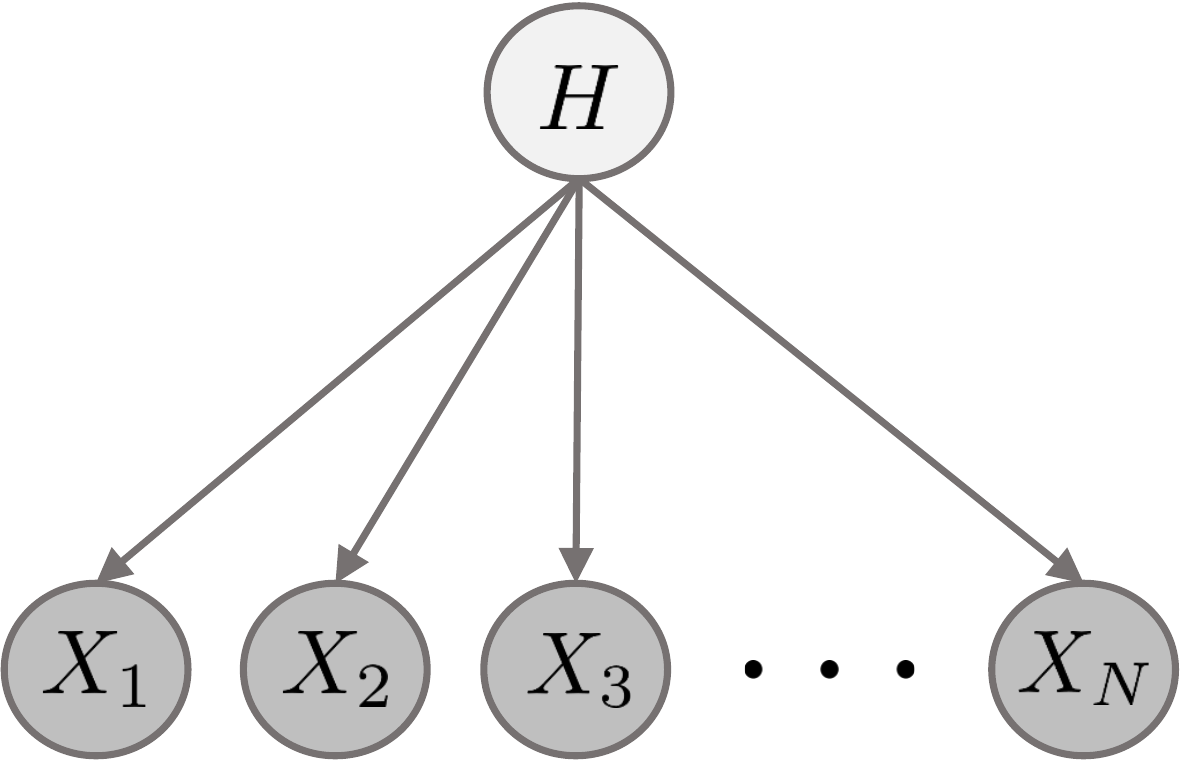}
\caption{The proposed generative model admits a latent variable naive Bayes interpretation.}
\label{NB}
\end{center}
\end{figure}

\begin{figure*} 
\begin{center}
\subfigure{\includegraphics[width=.33\columnwidth]{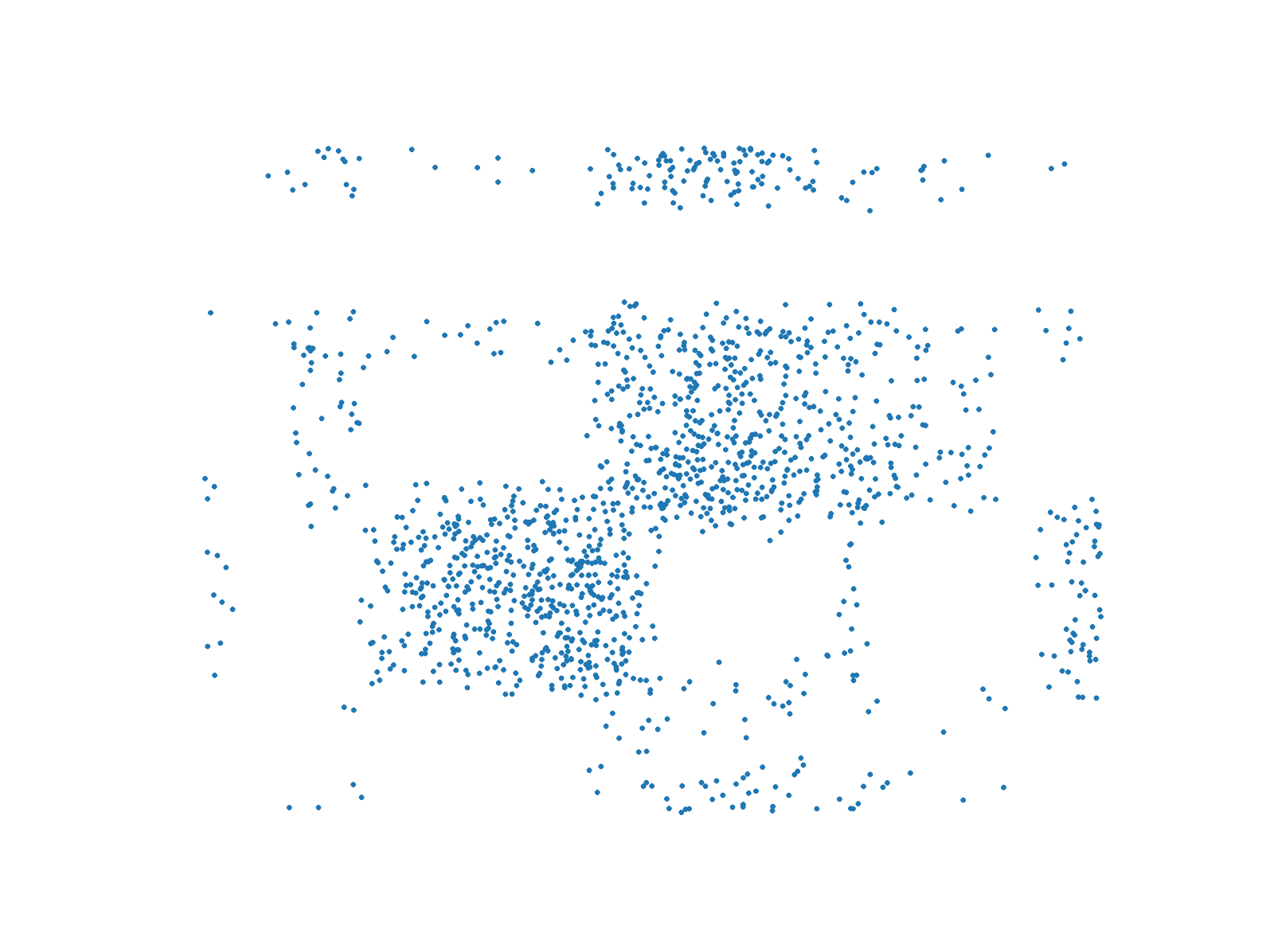}}
\subfigure{\includegraphics[width=.33\columnwidth]{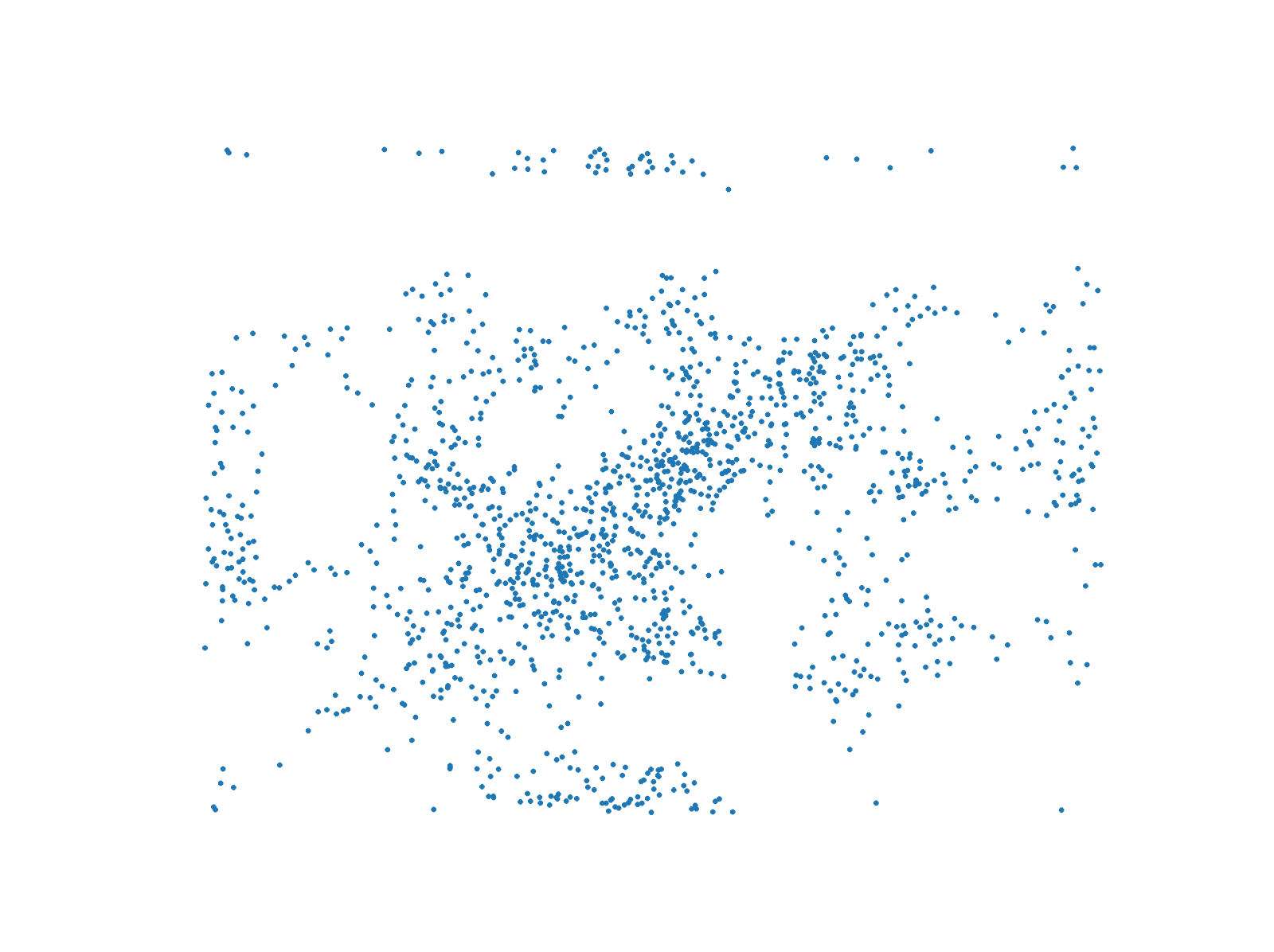}}
\subfigure{\includegraphics[width=.33\columnwidth]{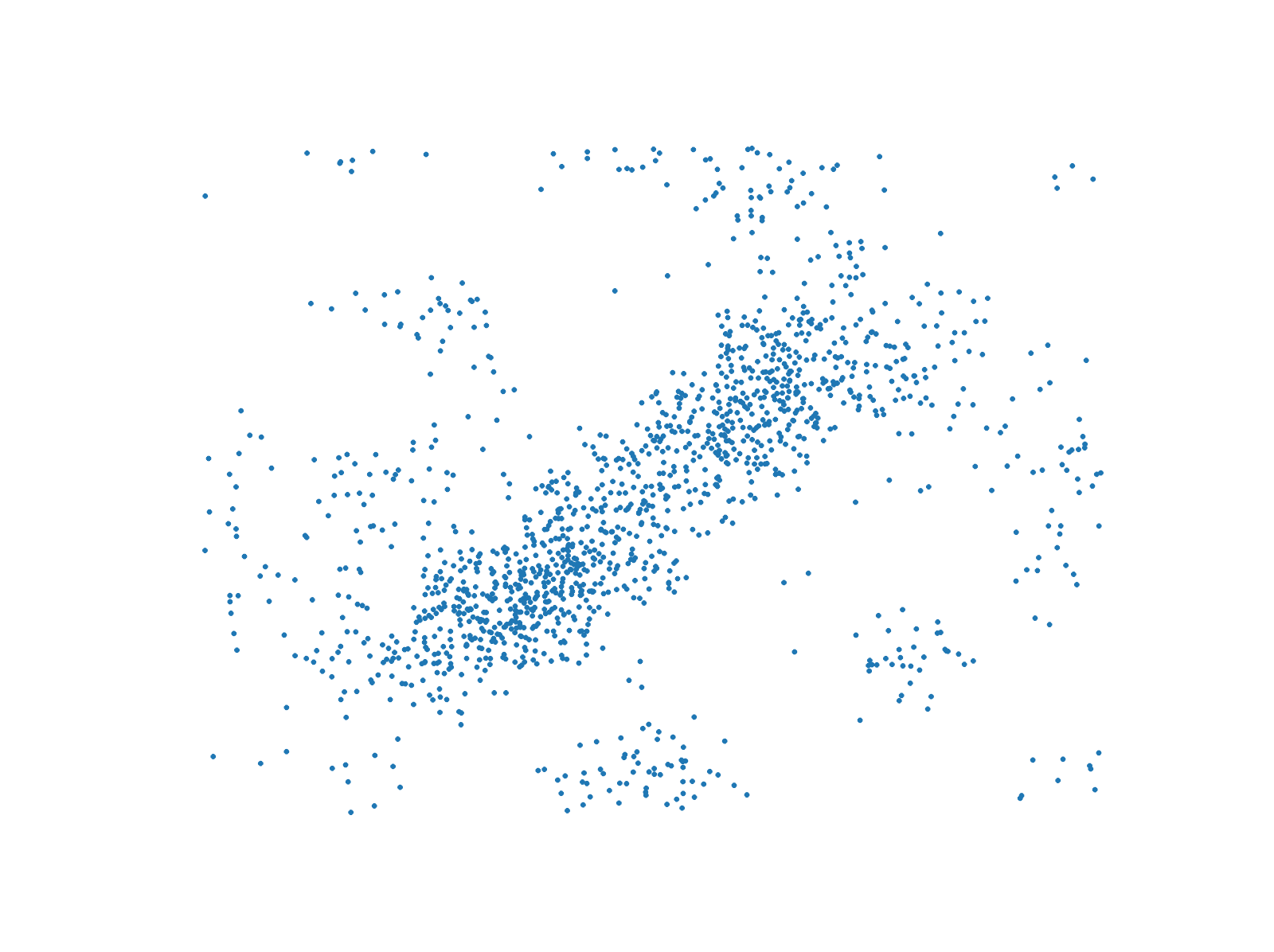}}
\subfigure{\includegraphics[width=.33\columnwidth]{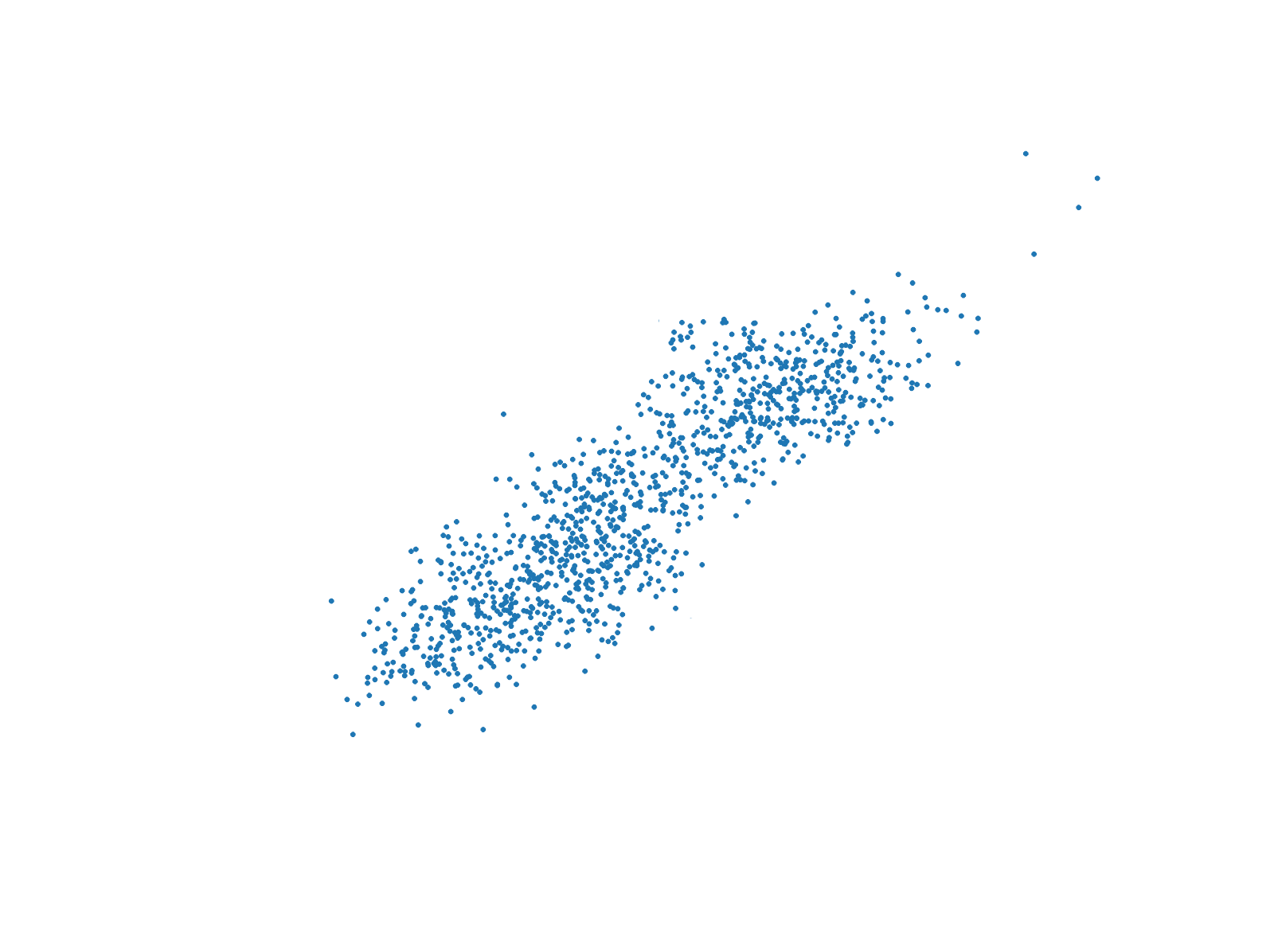}}
\subfigure{\includegraphics[width=.33\columnwidth]{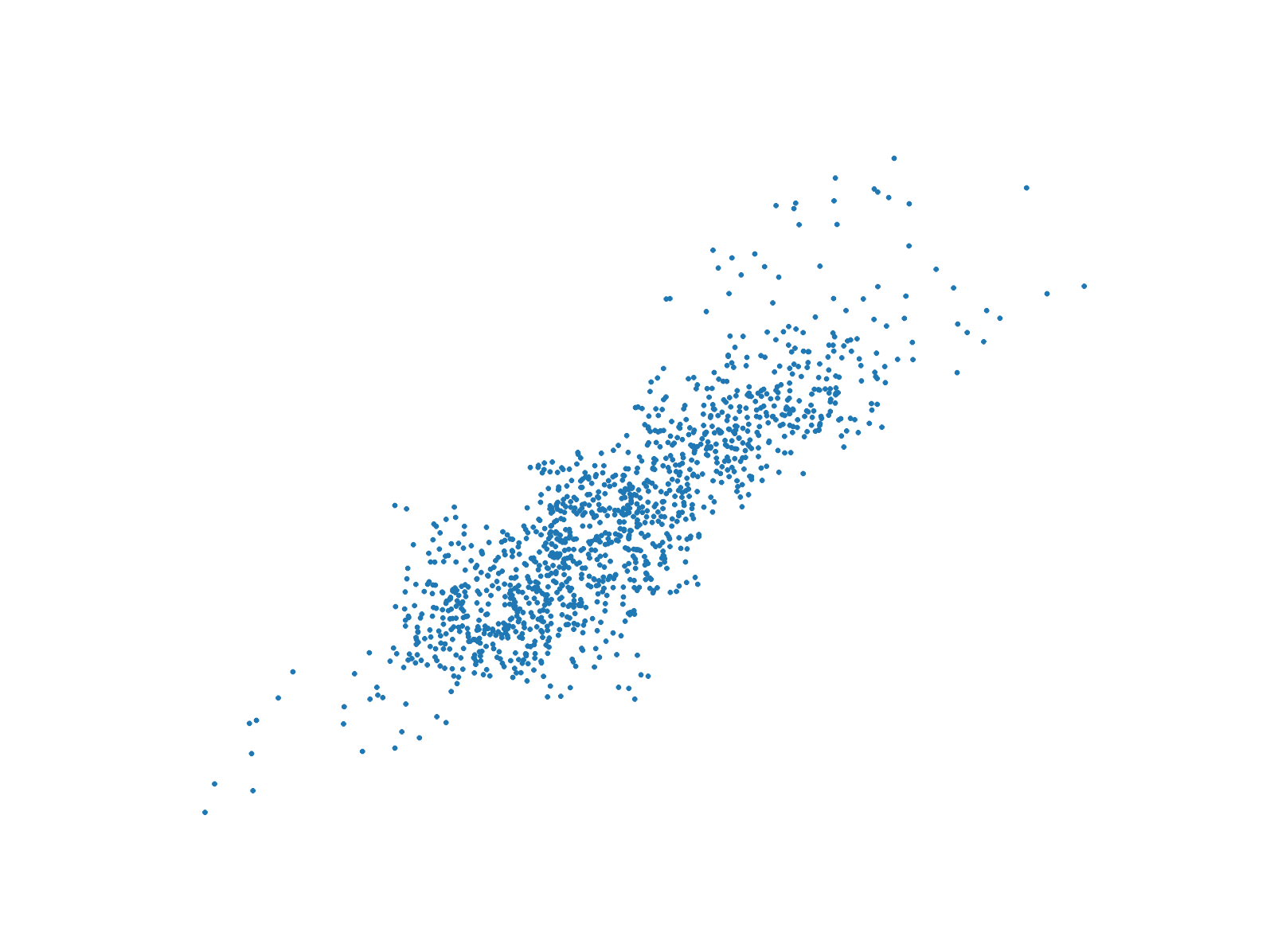}}
\subfigure{\includegraphics[width=.33\columnwidth]{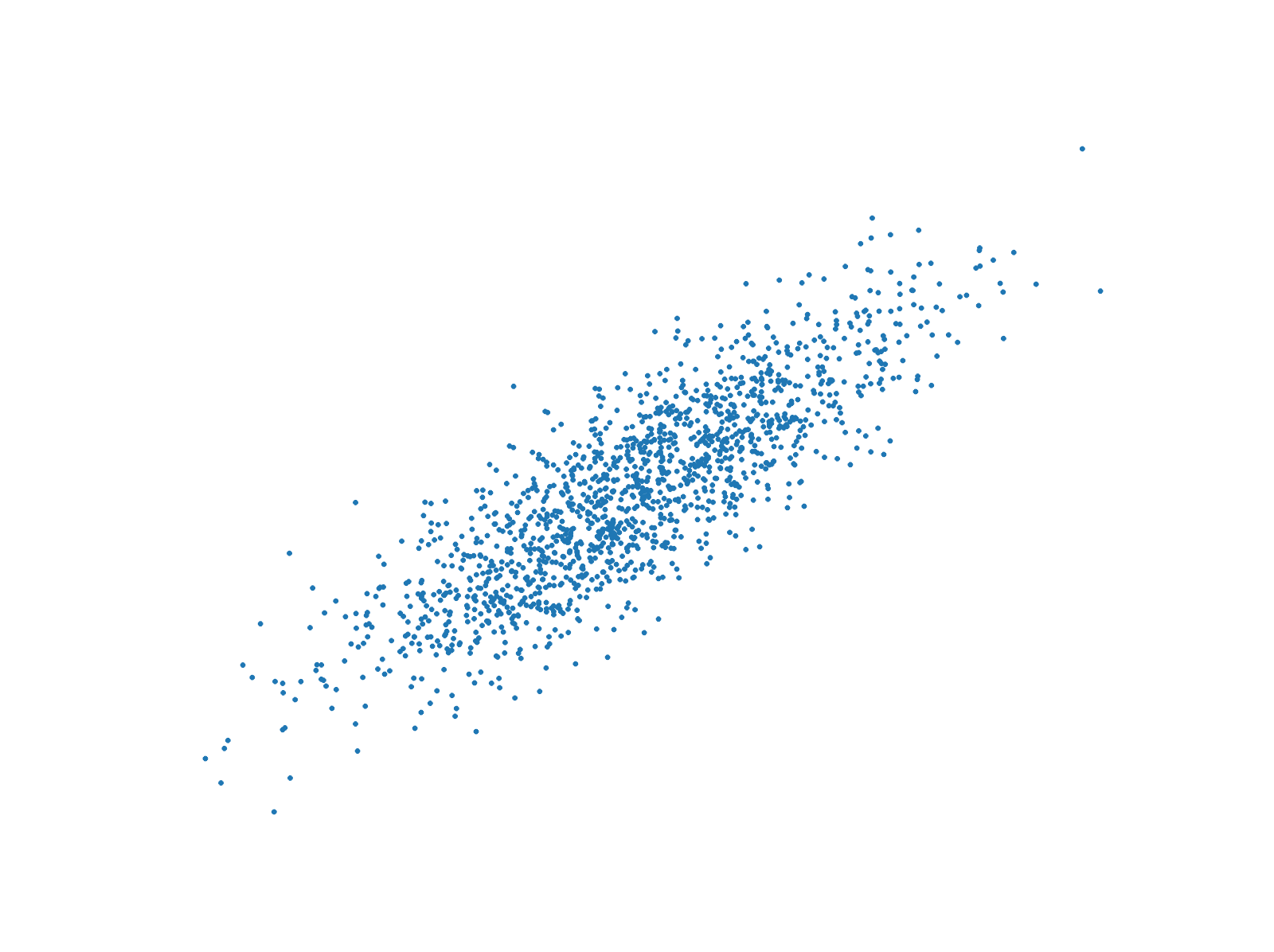}}

\subfigure{\includegraphics[width=.33\columnwidth]{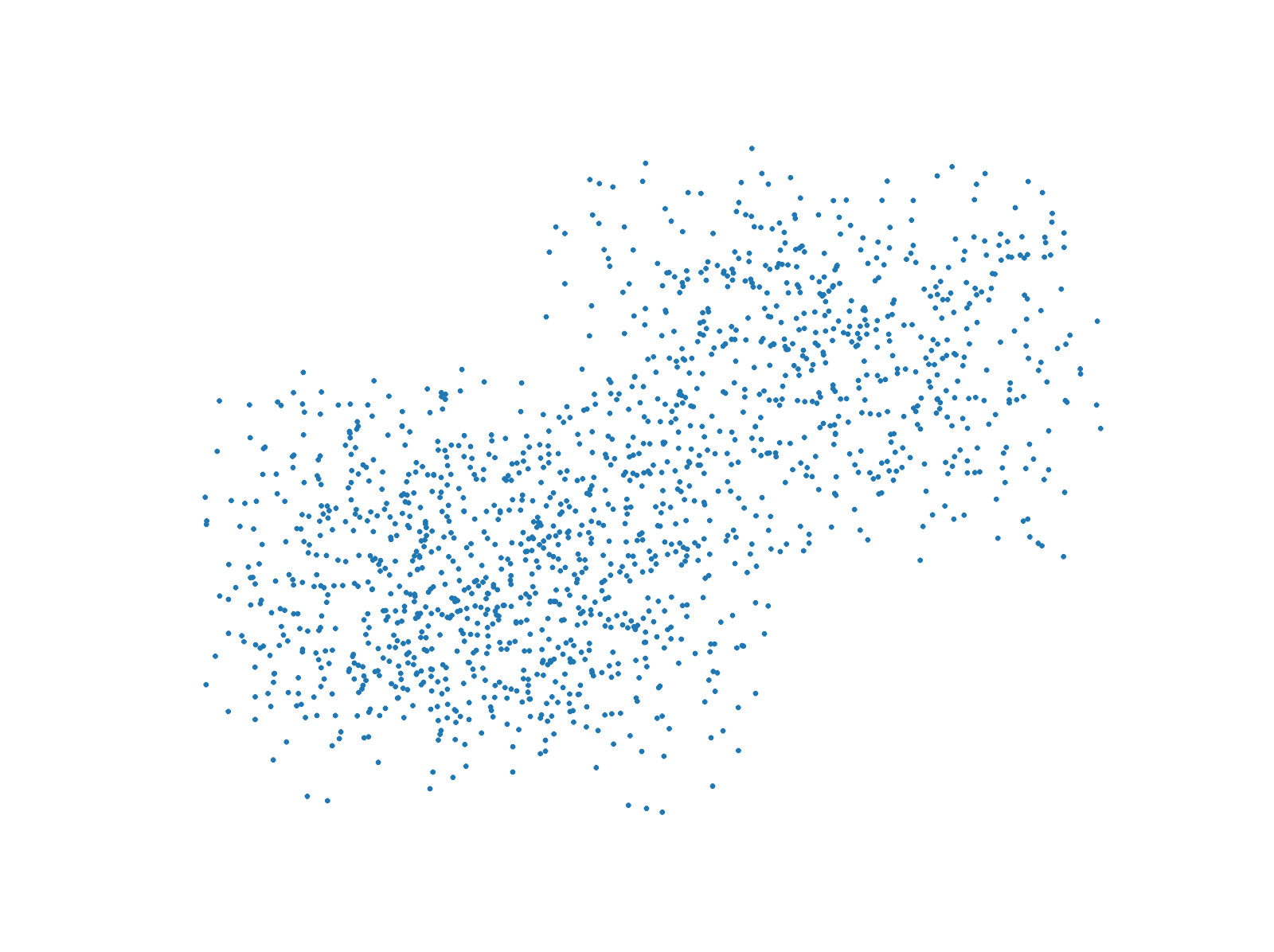}}
\subfigure{\includegraphics[width=.33\columnwidth]{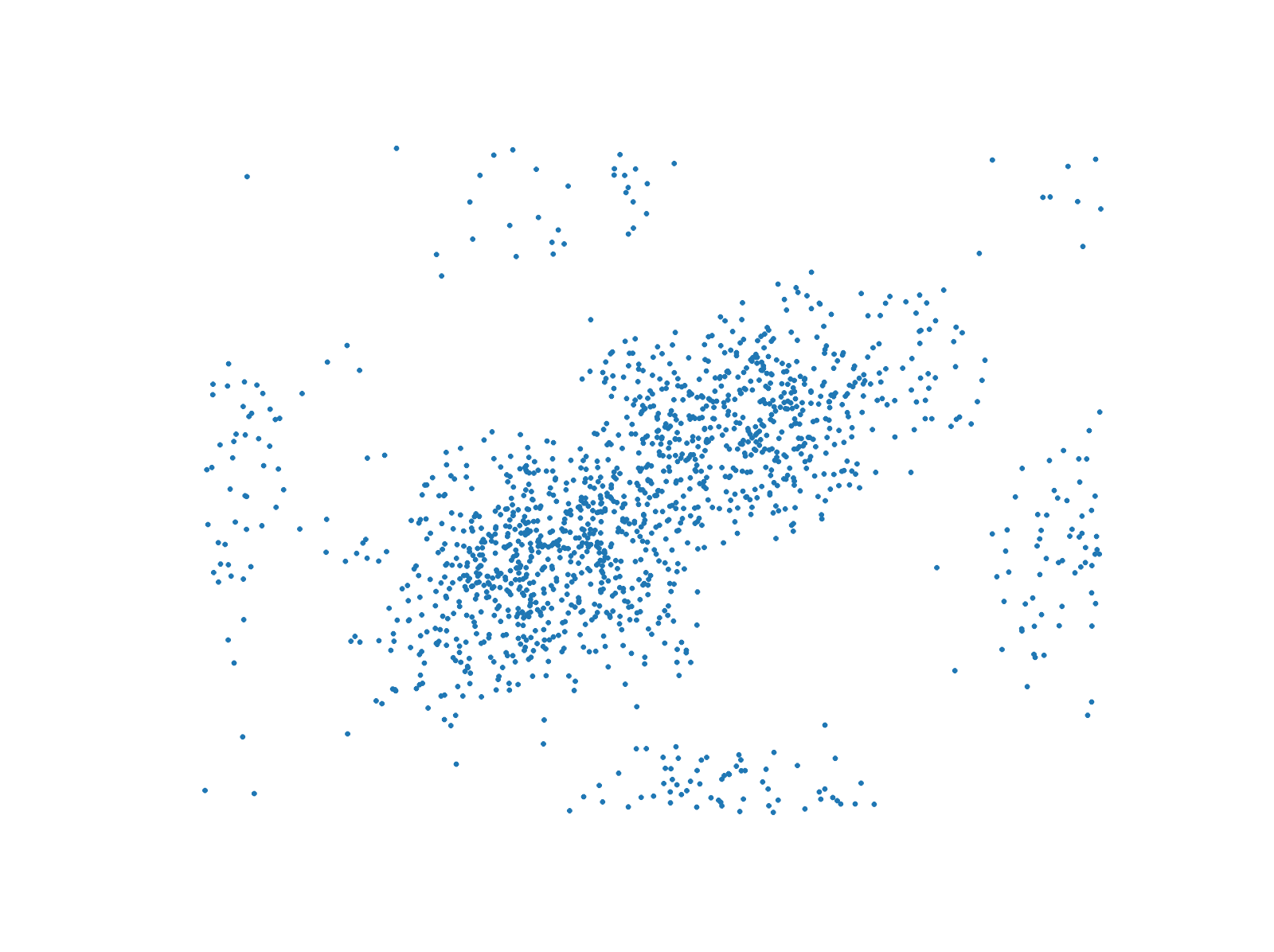}}
\subfigure{\includegraphics[width=.33\columnwidth]{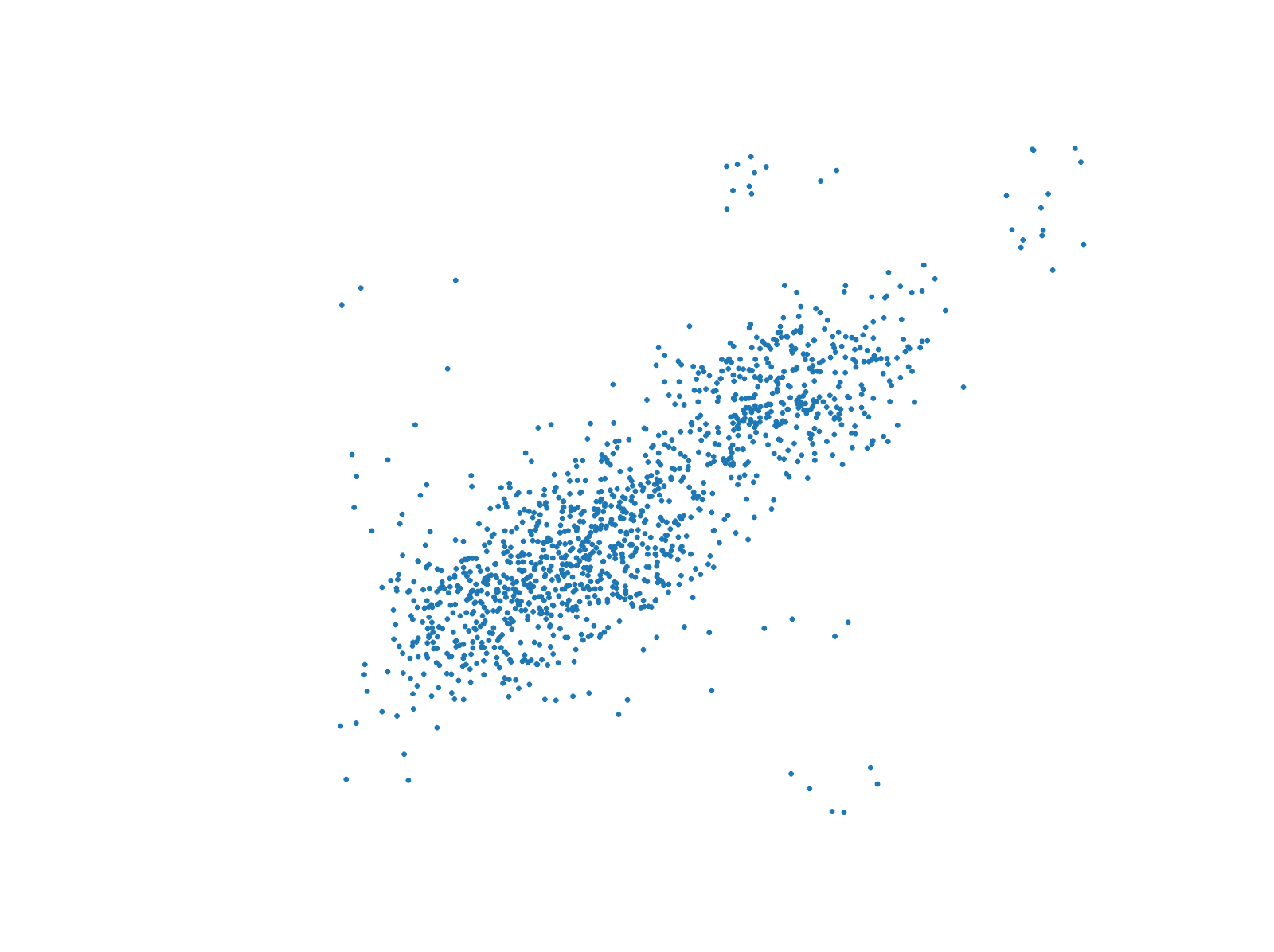}}
\subfigure{\includegraphics[width=.33\columnwidth]{./F_12_K_9}}
\subfigure{\includegraphics[width=.33\columnwidth]{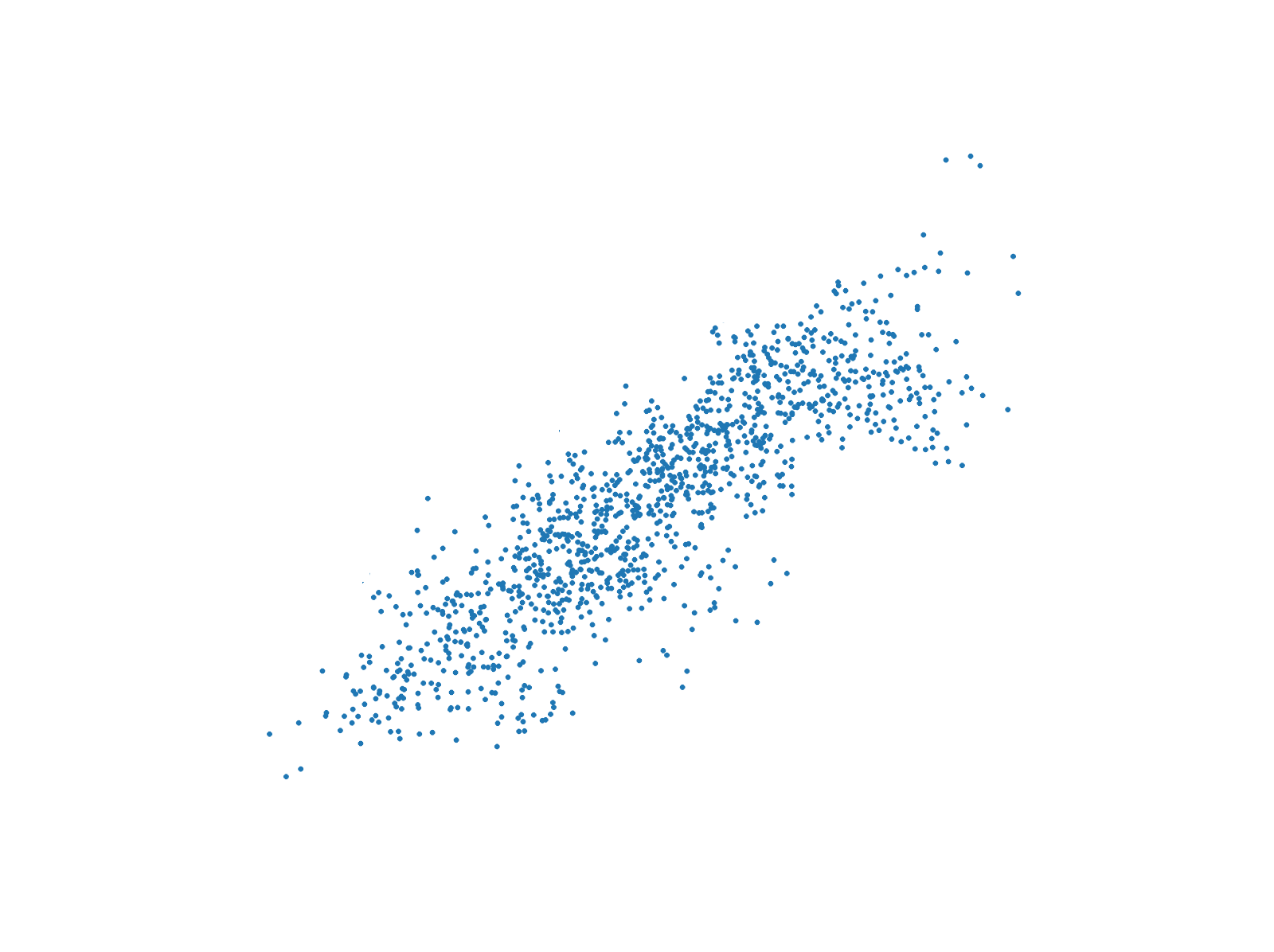}}
\subfigure{\includegraphics[width=.33\columnwidth]{./gt}}
\end{center}
\caption{Visualization of synthetic $M'=1500$ samples generated from the proposed model trained on the Weight-Height dataset for different $F,K$ parameter combinations -- The rightmost figure represents the ground truth. On the first row, we fixed $K$, $K=4$, and varied $F$, $F \in[2,4,6,8,10]$ (from left to right). On the second row, we fixed $F$, $F=8$, and varied $K$,  $K \in[1,2,3,4,5]$ (from left to right).}
\label{toy_weight}
\end{figure*}

\begin{figure*}
\begin{center}
\subfigure{\includegraphics[width=.32\columnwidth]{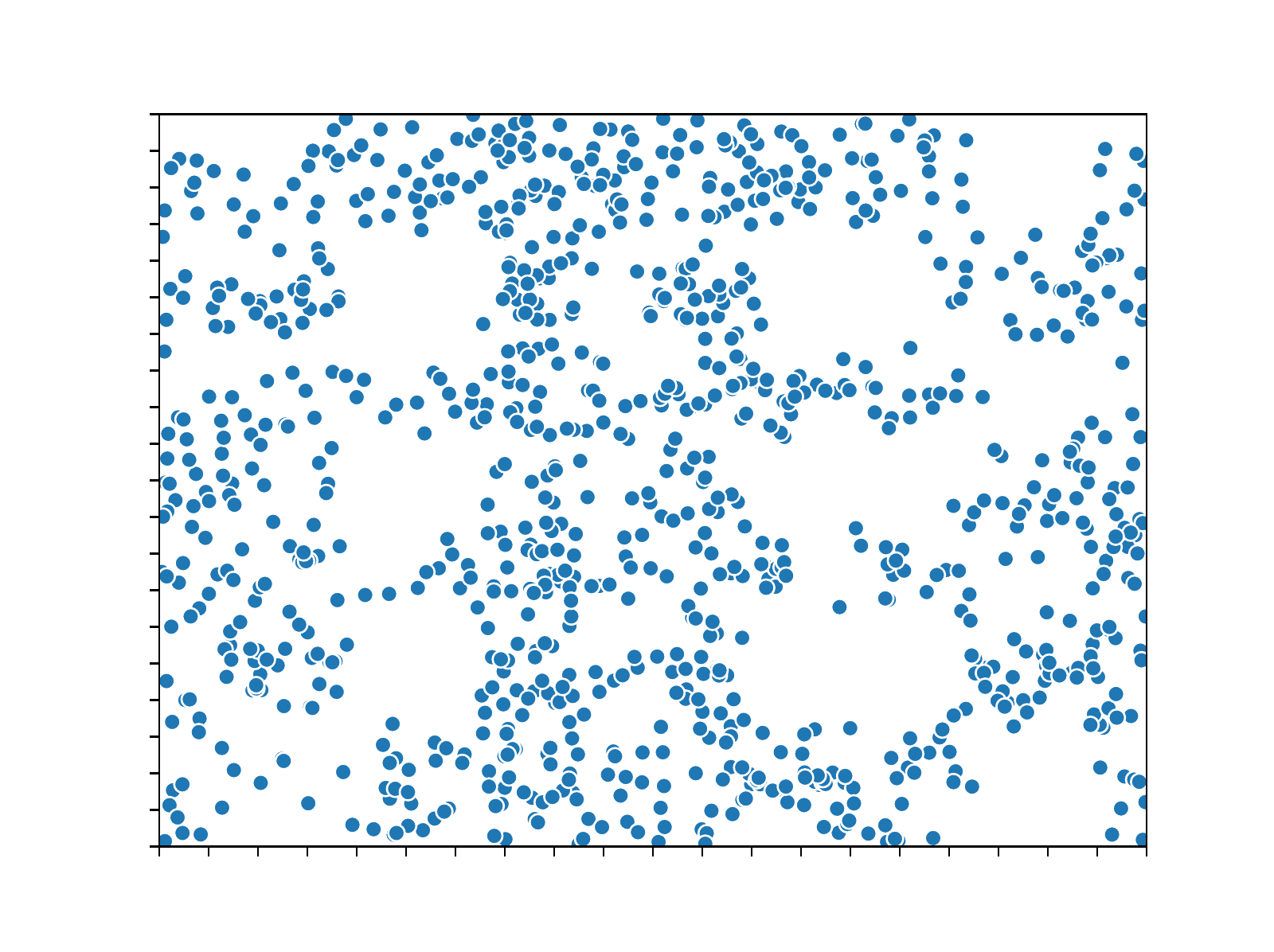}}
\subfigure{\includegraphics[width=.32\columnwidth]{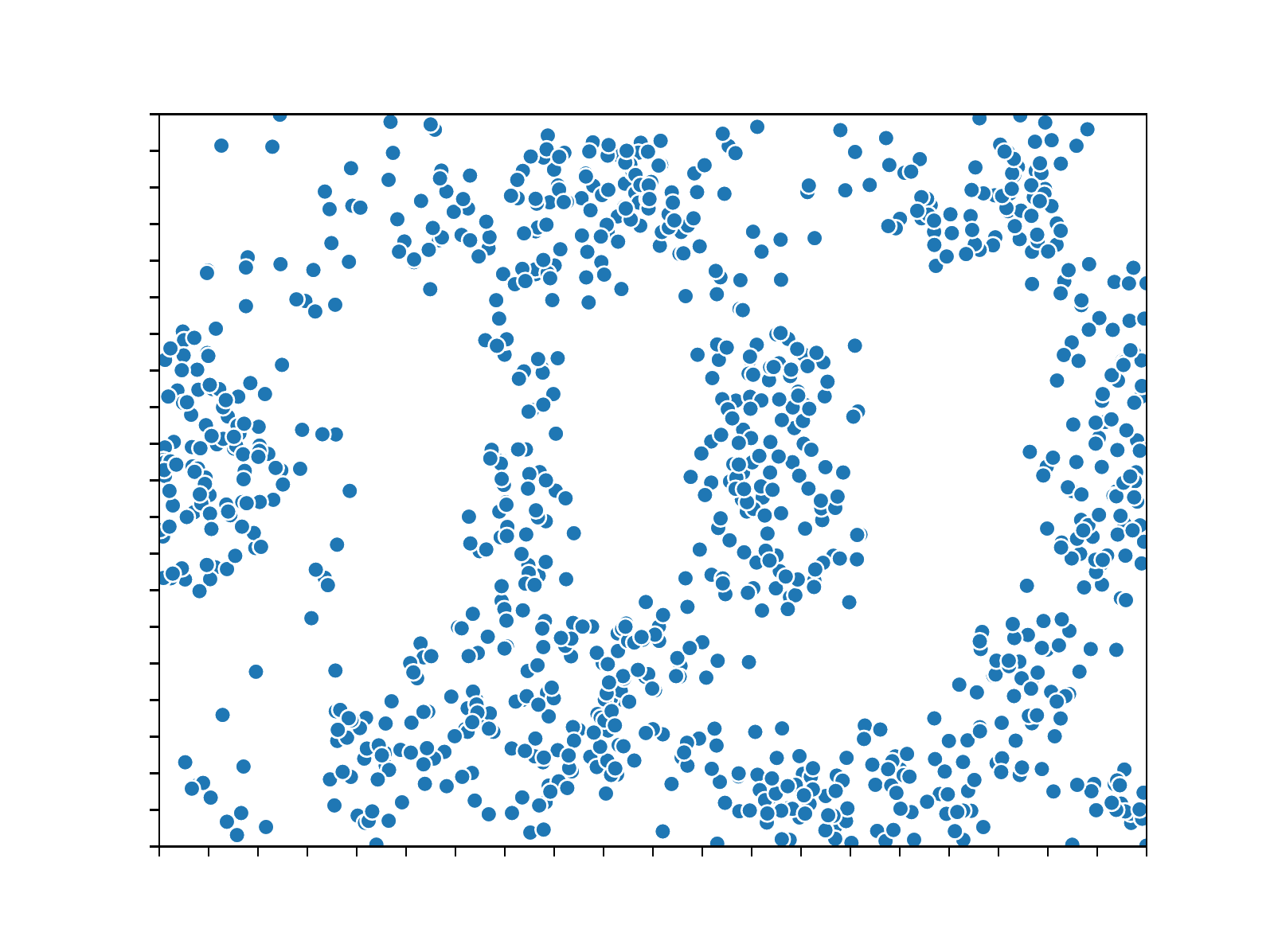}}
\subfigure{\includegraphics[width=.32\columnwidth]{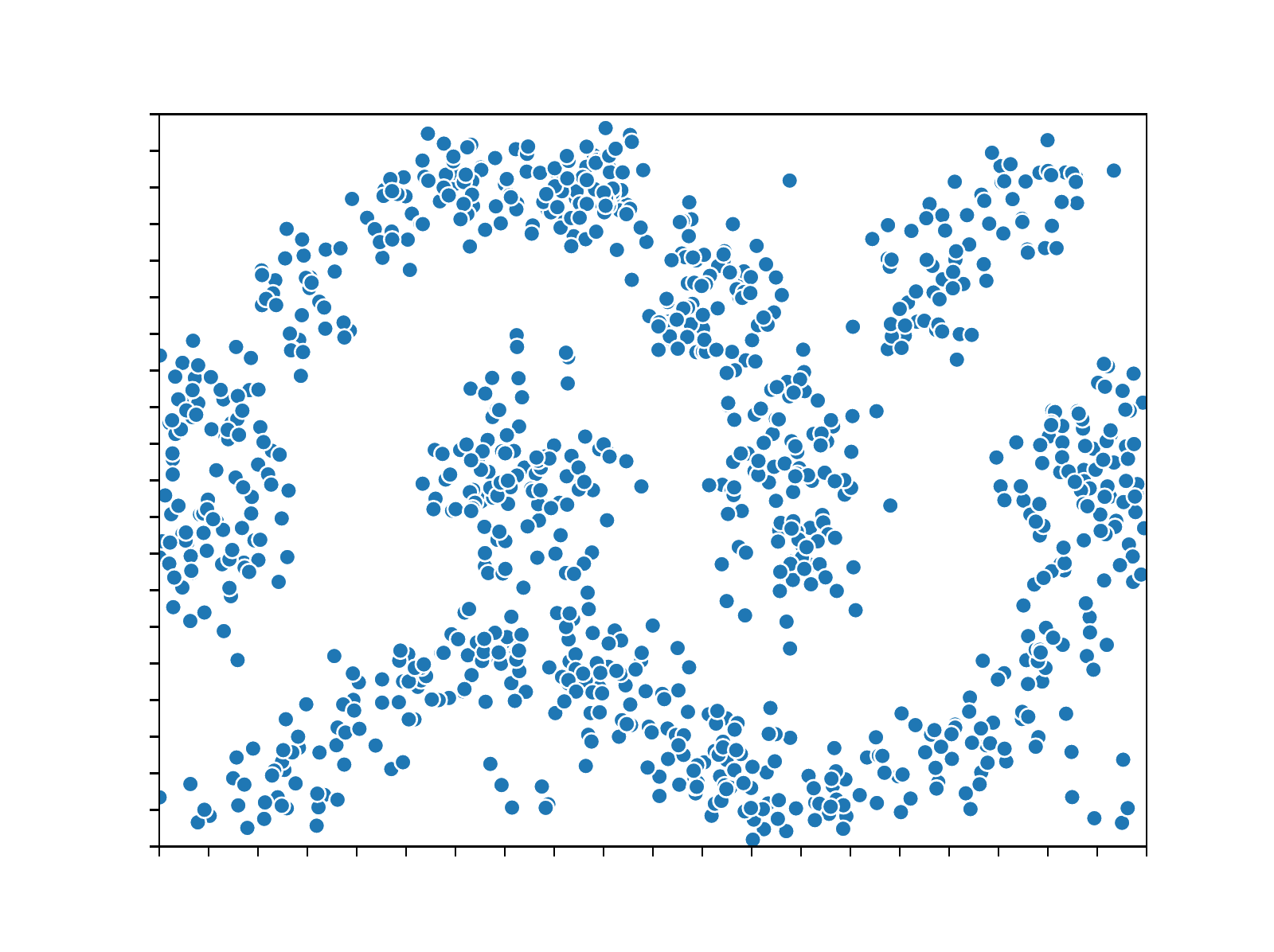}}
\subfigure{\includegraphics[width=.32\columnwidth]{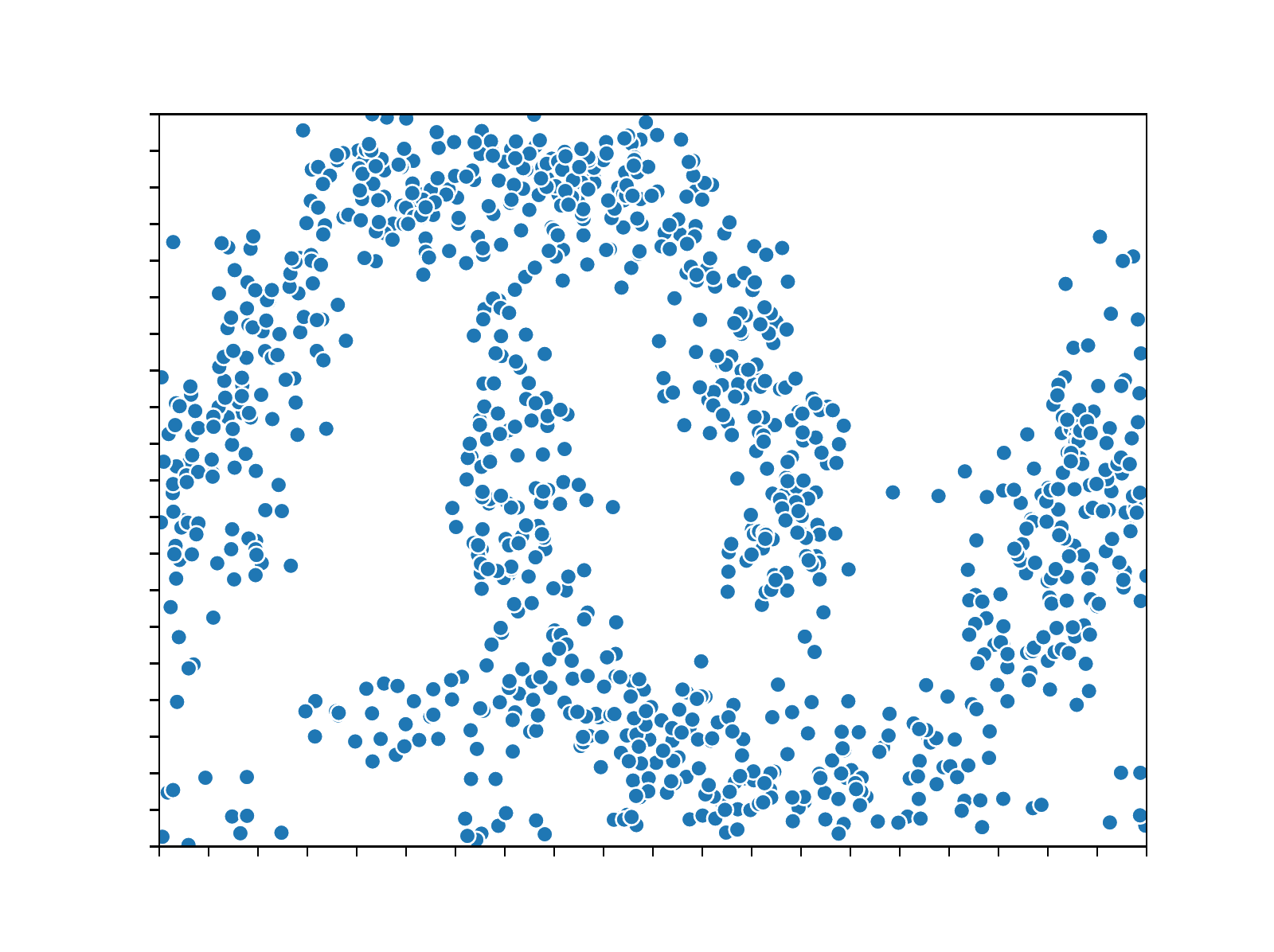}}
\subfigure{\includegraphics[width=.32\columnwidth]{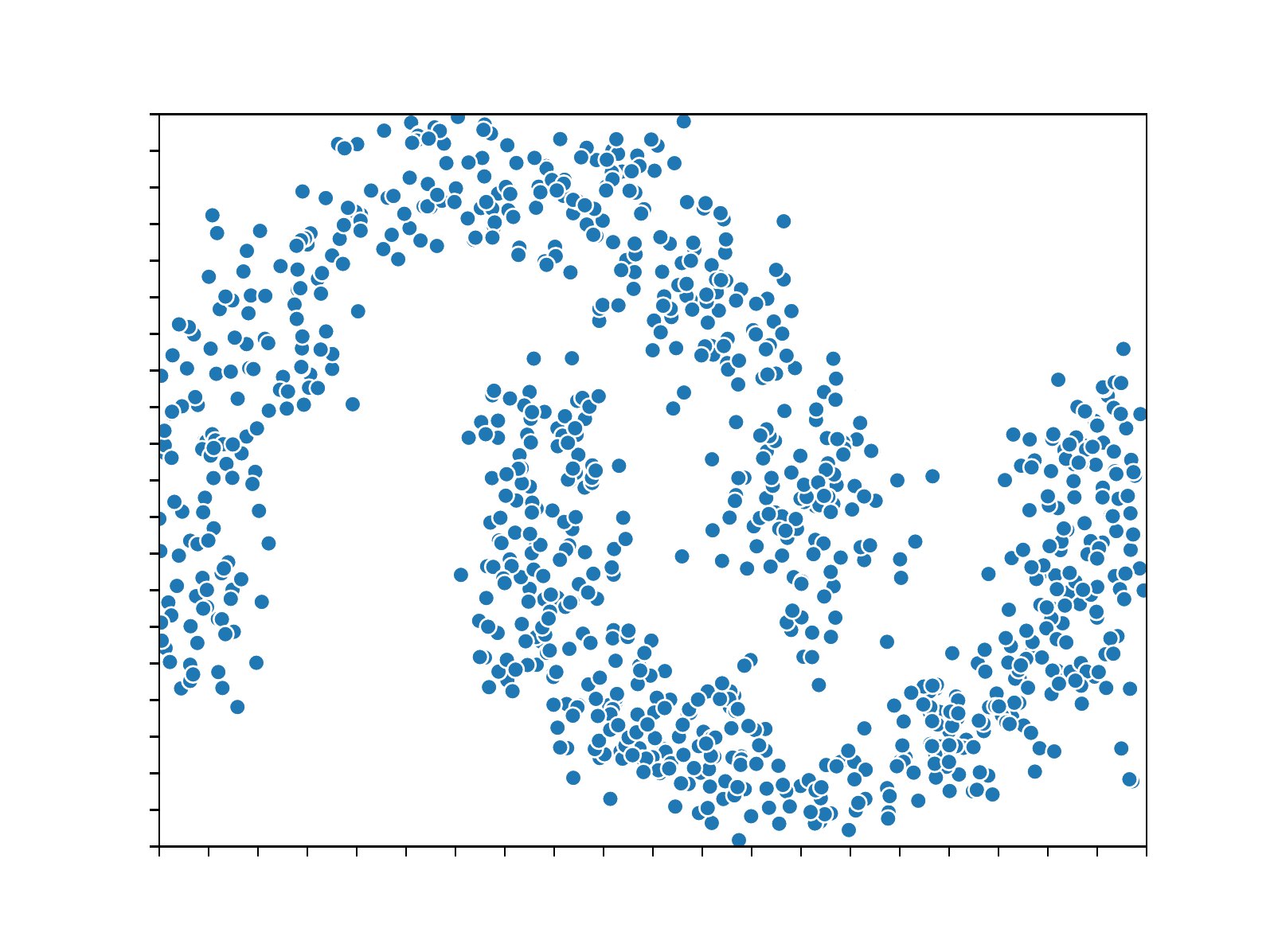}}
\subfigure{\includegraphics[width=.32\columnwidth]{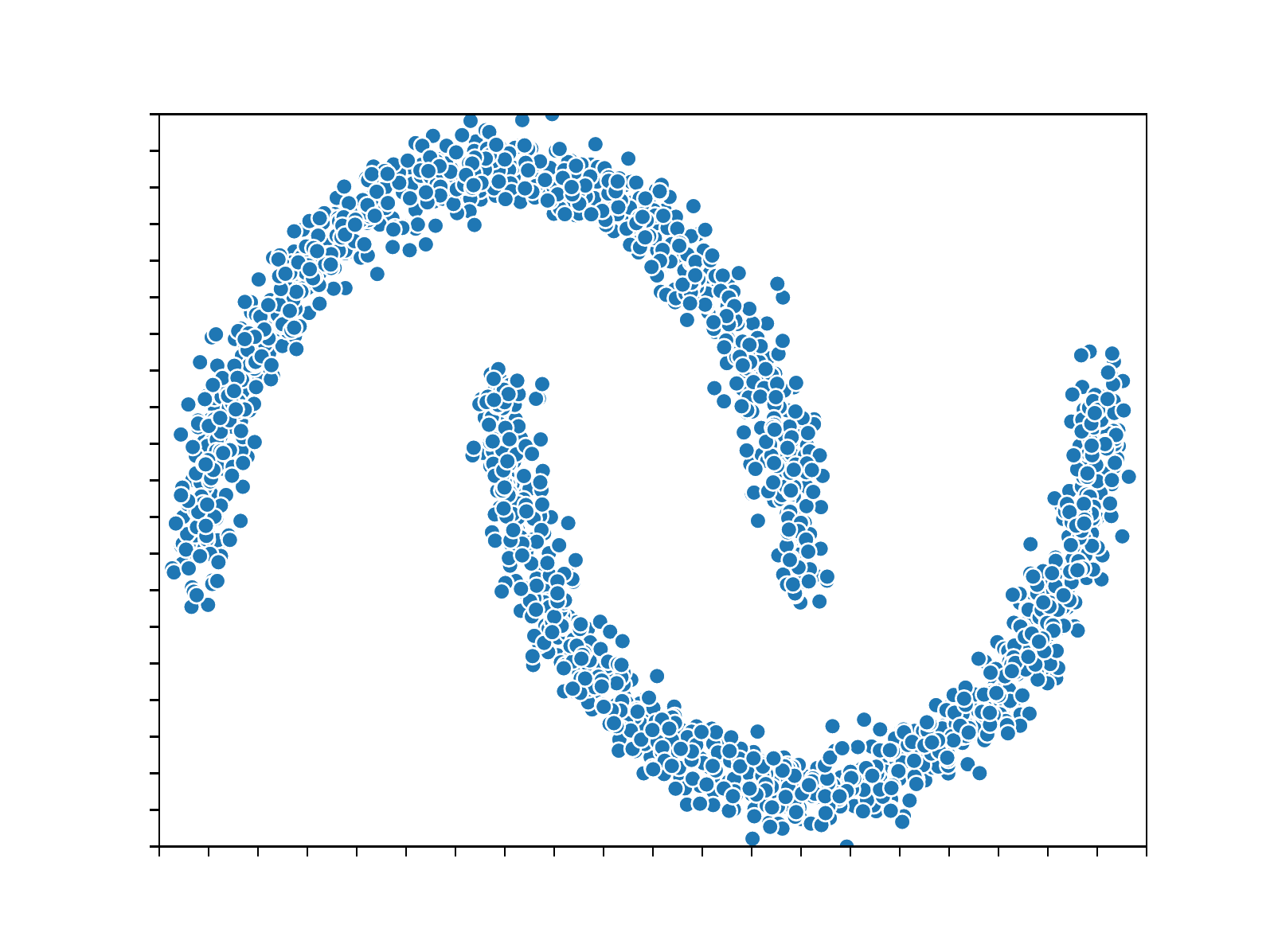}}
\subfigure{\includegraphics[width=.32\columnwidth]{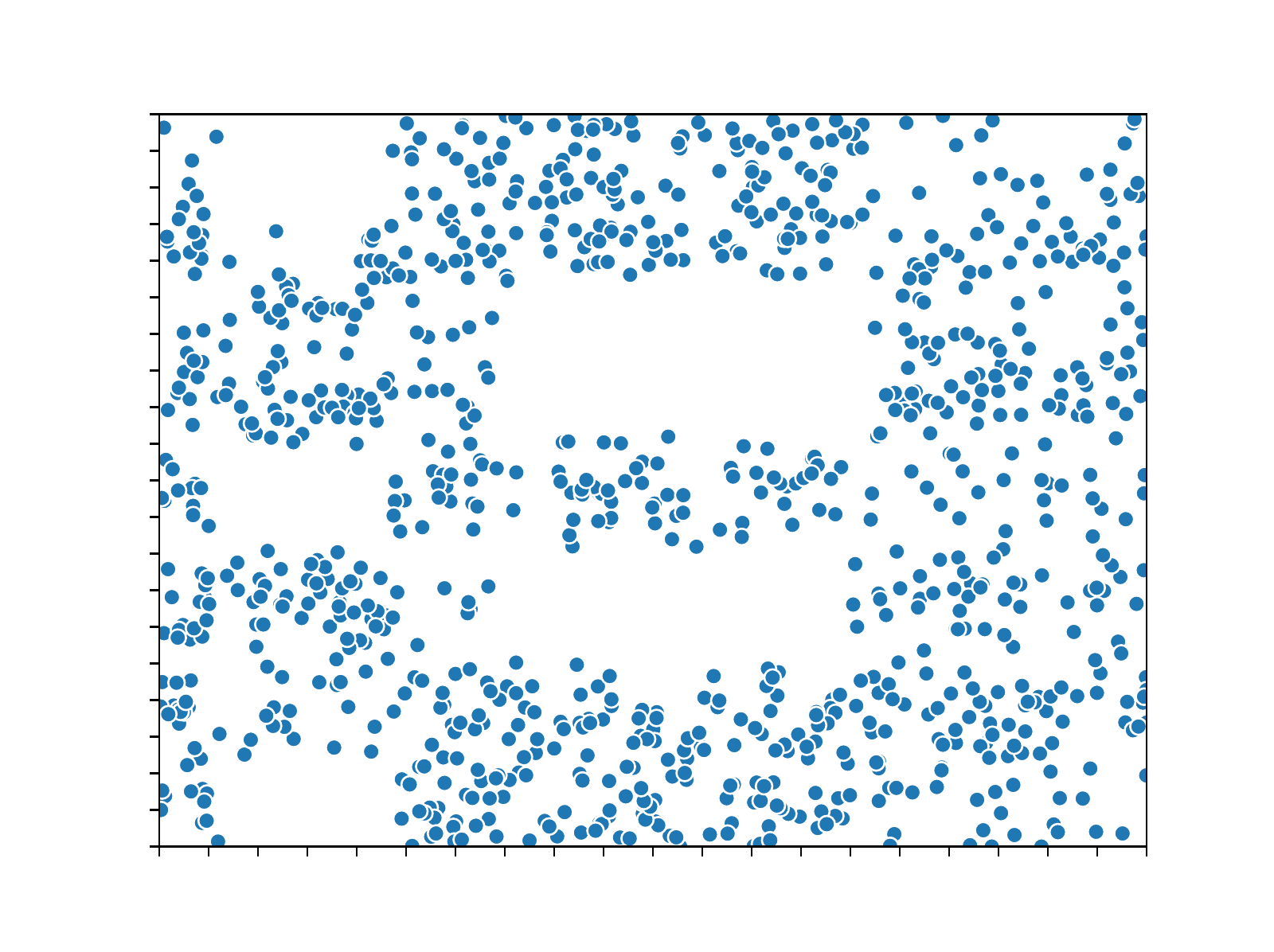}}
\subfigure{\includegraphics[width=.32\columnwidth]{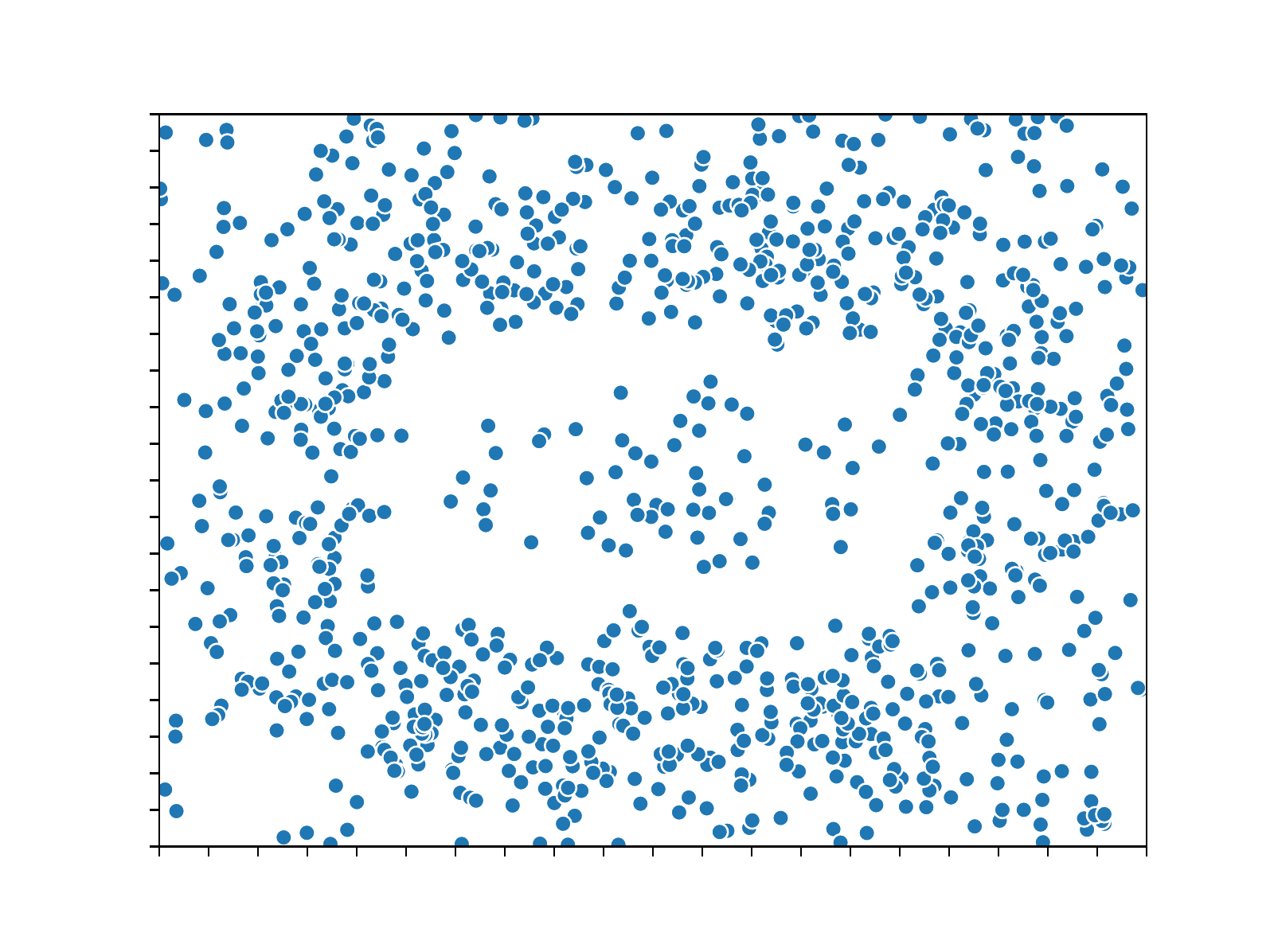}}
\subfigure{\includegraphics[width=.32\columnwidth]{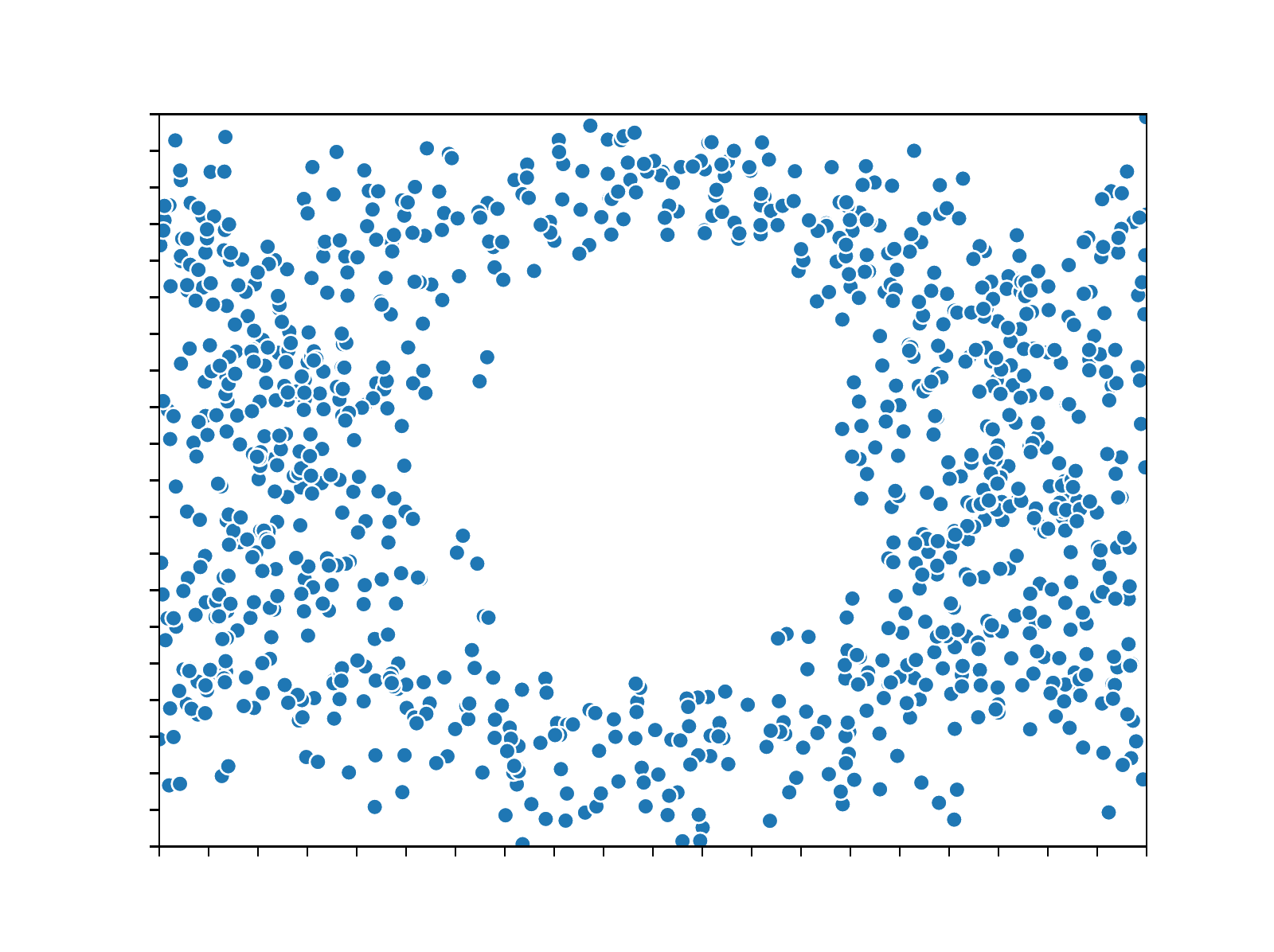}}
\subfigure{\includegraphics[width=.32\columnwidth]{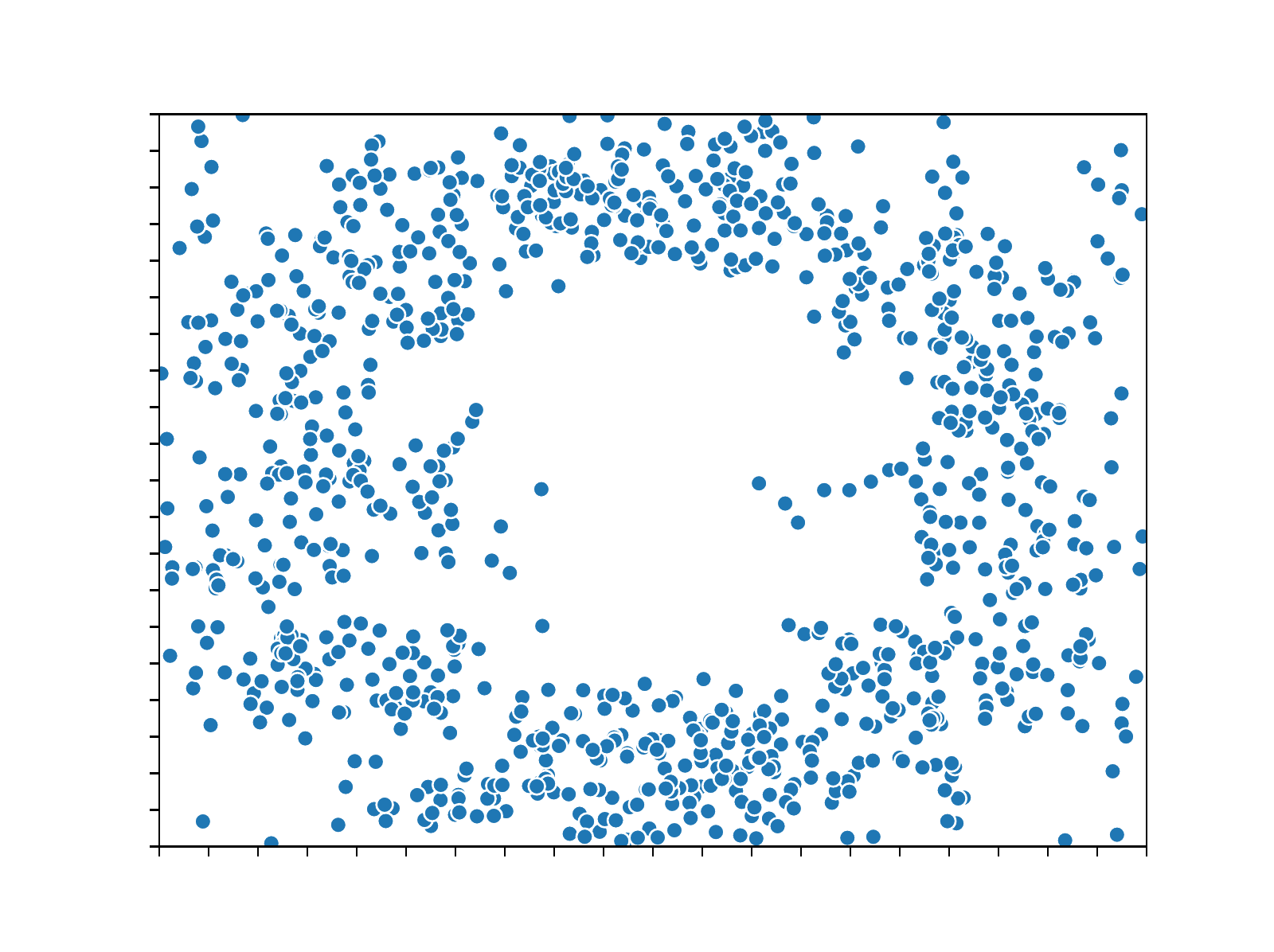}}
\subfigure{\includegraphics[width=.32\columnwidth]{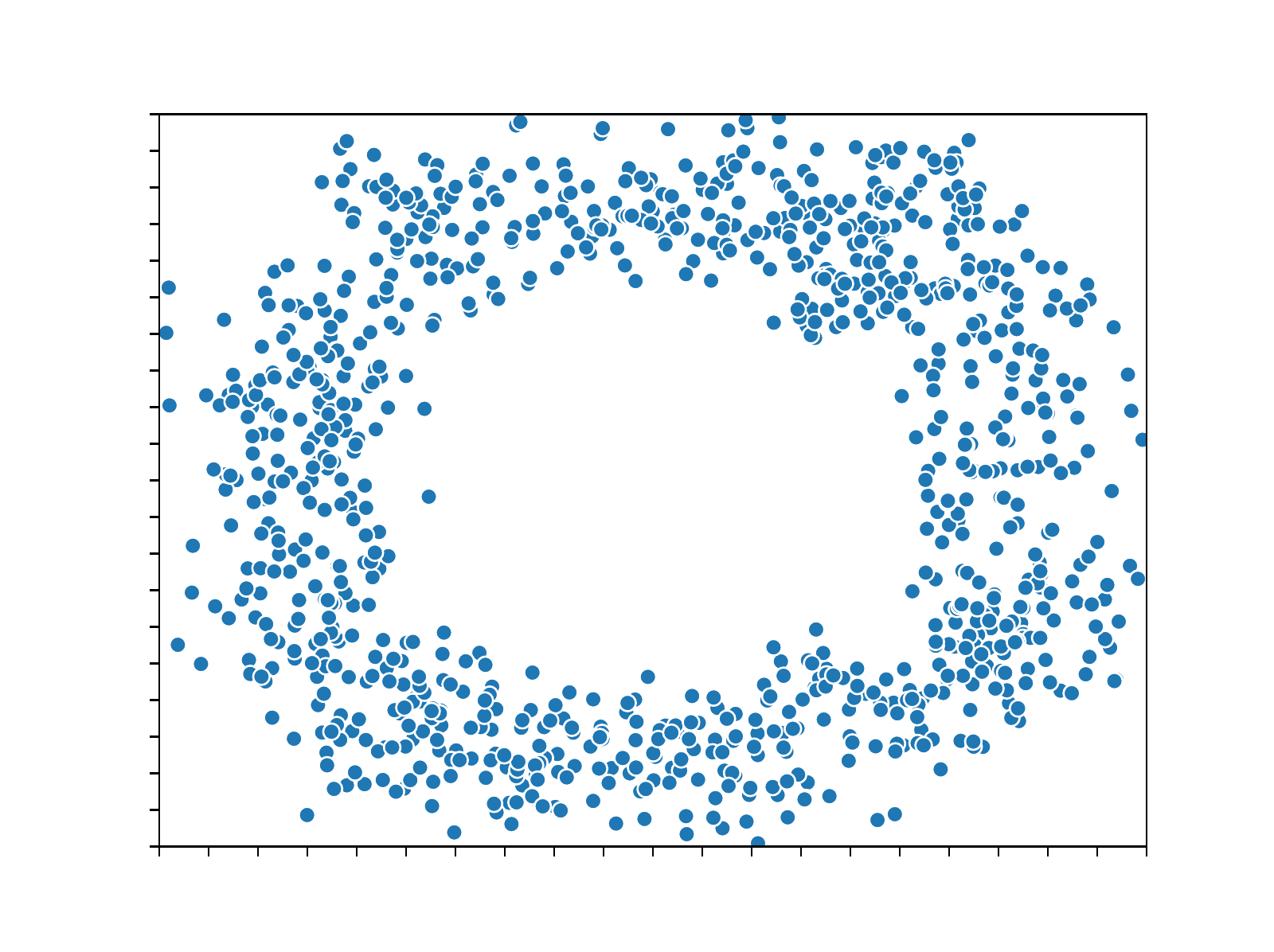}}
\subfigure{\includegraphics[width=.32\columnwidth]{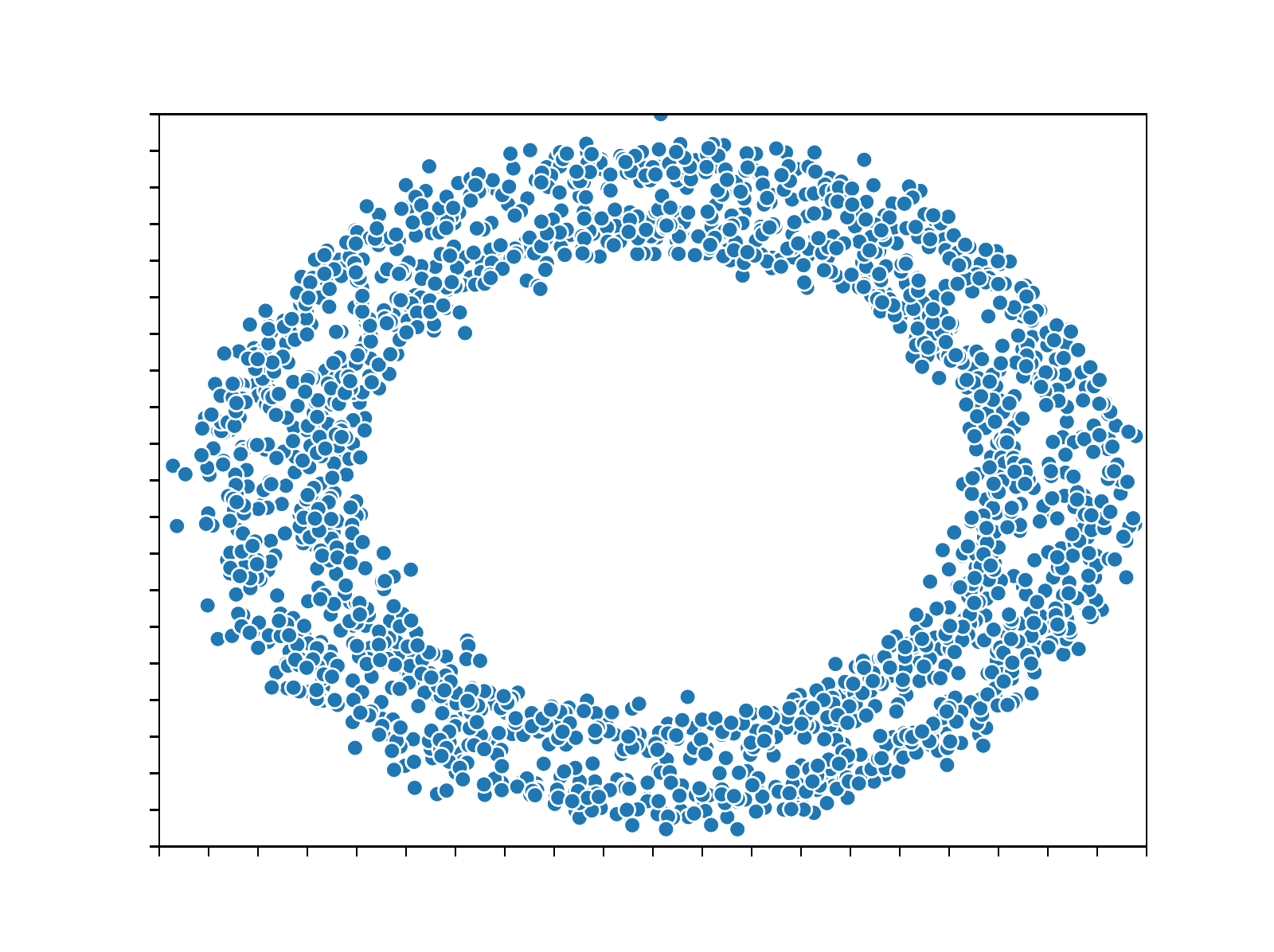}}
\caption{Qualitative synthetic $M'=1500$ samples obtained from the proposed model trained on $M=2000$ samples of a toy $2$-D Moons and Circles datasets (for fixed $K$, $K=11$, from left to right $F \in[1,2,3,4,6]$ -- The rightmost figures represent the ground truth).}\label{fig:2d}
\end{center}
\end{figure*}

By linearity and separability of the multidimensional Fourier transformation it follows that 
\begin{align} 
{\tilde{f}}_{{\boldsymbol X}}({\mathbf x}) &=\sum_{h=1}^F p_H(h)\prod_{n=1}^N{\sum_{k_n=-K}^K}\Phi_{X_n|H=h}[k_n]\text{ }e^{-j2\pi k_n x_n}\nonumber\\
&=\sum_{h=1}^F p_H(h)\prod_{n=1}^N f_{X_n|H}(x_n|h). 
\label{eq:pdf_est}
\end{align}
This generative model can be interpreted as mixture of product distributions \cite{kargas2019learning}. The joint PDF $f_{{\boldsymbol X}}$ is a mixture of $F$ separable component PDFs, i.e., there exists a `hidden' random variable $H$ taking values in $\left\{1,\ldots,F\right\}$ that selects the operational component of the mixture, and given $H$ the random variables $X_1,\ldots,X_N$ become independent (See Figure \ref{NB} for visualization of this model). We have thus shown the following result: 
\begin{Prop}\label{prop:seperable_densities}
 Truncating the multidimensional Fourier series (sampled multivariate characteristic function) of any compactly supported random vector is equivalent to approximating the corresponding multivariate density by a finite mixture of separable densities.  
 
 Thus, by choosing appropriate $K$ and $F$, it is possible to represent and approximate any compactly supported density that it is sufficiently smooth by the proposed model. See Figures \ref{fig:2d}, \ref{toy_weight} where we showcase how each parameter affects the modeling of complex structures in $2$D synthetic datasets. 
\end{Prop}

Conversely, if one {\em assumes} that the sought multivariate density is a finite mixture of separable densities, then it is easy to show that the corresponding characteristic function is likewise a mixture of separable characteristic functions:

\begin{flalign}
\Phi_{{\boldsymbol X}}({\boldsymbol \nu}) &= E\left[e^{j {\boldsymbol \nu}^T {\boldsymbol X}}\right]\nonumber\\
&=E_H\left[E_{{\boldsymbol X}|H}\left[e^{j \nu_1 X_1} \cdots e^{j \nu_N X_N}\right]\right]\nonumber\\
&=E_H\left[ \Phi_{X_1|H}(\nu_1|H) \cdots \Phi_{X_n|H}(\nu_N|H)\right]\nonumber\\
&=\sum_{h=1}^F p_H(h) \prod_{n=1}^N {\Phi}_{X_n|H}(\nu_n|h).
\end{flalign}

If we sample the above on any finite $N$-dimensional grid, we obtain an $N$-way tensor and its polyadic decomposition. Such decomposition is unique, under mild conditions~\cite{sidiropoulos2017tensor}. It follows that: 
\begin{Prop}\label{prop:density_identifiability} A compactly supported multivariate ($N \geq 3$) mixture of separable densities is identifiable from (samples of) its characteristic function, under mild conditions.
\end{Prop}

The above analysis motivates the following course of action. Given a set of realizations $\left\{ {\mathbf x}_m  \right\}_{m=1}^M$,
\begin{enumerate}
\item estimate
\begin{equation}
 \underline{{\boldsymbol{\Phi}}}[{\mathbf k}] = \frac{1}{M}\sum_{m=1}^M e^{j 2 \pi {\mathbf k}^T{{\mathbf x}_{m}}}, \label{eq:sample_avgs}
\end{equation}
\item fit a low-rank model
\begin{equation}
 \underline{{{\boldsymbol{\Phi}}}}[{\mathbf k}] \approx \sum_{h=1}^F p_H(h) \prod_{n=1}^N\Phi_{X_n|H=h}[k_n],   
\end{equation}
\item and invert using
\begin{multline}
{{f}}_{{\boldsymbol X}}({\mathbf x})= \sum_{h=1}^F p_H(h)\prod_{n=1}^N f_{X_n|H}(x_n|h), ~\text{where}~
\\ 
 f_{X_n|H}(x_n|h)=\sum\limits_{k_n=-K}^K\Phi_{X_n|H=h}[k_n]e^{-j2\pi k_n x_n}.
 \label{eq:density_cpd}
\end{multline}
\end{enumerate}

When building any statistical model, identifiability is a fundamental question. 
A statistical model is said to be identifiable when, given a sufficient number of observed data, it is possible to uniquely recover the data-generating distribution. When applying a non-identifiable model, different structures or interpretations may arise from distinct parametrizations that explain the data equally well. Most deep generative models do not address the question of identifiability, and thus may fail to deliver the true latent representations that generate the observations. Our approach is fundamentally different, because it builds on rigorous and controllable Fourier approximation and identifiability of the characteristic tensor.

In the Appendix (Section \ref{SM}), we provide additional statistical insights regarding the proposed methodology, including the asymptotic behavior of the empirical characteristic function and the mean squared error reduction afforded by low-rank tensor modeling in the characteristic function domain. 

Two issues remain. First, uniqueness of CPD only implies that each rank-one factor is unique, but leaves scaling/counter-scaling freedom in $p_H$ and the conditional characteristic functions. To resolve this, we can use the fact that each conditional characteristic function must be equal to $1$ at the origin (zero frequency). Likewise, $p_H$ must be a valid probability mass function. These constraints fix the scaling indeterminacy. 

We note here that, under certain rank conditions (see Section \ref{sec:tensors}) on the Fourier series coefficient tensor, the proposed method ensures that the reconstructed density is positive and integrates to one, as it should. This is due to the uniqueness properties of the Fourier series representation and the CPD: if there exists a density that generates a low-rank characteristic tensor, and that tensor can be uniquely decomposed, the sum of Fourier inverses of its components is unique, and therefore equal to the generating density. Under ideal low-rank conditions, this is true even if we ignore the constraints implied by positivity when we decompose the characteristic tensor in the Fourier domain. This is convenient because strictly enforcing those in the Fourier domain would entail cumbersome spectral factorization-type (positive semidefinite) constraints.
We therefore propose the following formulation:
\begin{equation}
\begin{aligned}
\min  & \quad \left \|   \underline{\boldsymbol{\Phi}} - [\![\boldsymbol{\lambda},\mathbf{A}_1,\ldots,\mathbf{A}_N  ]\!] \right \|_F^2 \\
\text{subject to} & \quad \boldsymbol{\lambda}\geq \boldsymbol{0}, {\boldsymbol{1}^{T}\boldsymbol{\lambda} = 1, }  \\
& \quad { \mathbf{A}_n(K+1,:)=\mathbf{1}^{T},~  n=1\ldots N,}
\end{aligned}
\end{equation}
where $\mathbf{A}_n(K+1+k_n,h)$ holds $\Phi_{X_n|H=h}[k_n]$, and $\boldsymbol{\lambda}(h)$ holds $p_H(h)$. 

The second issue is more important. When $N$ is large, instantiating or even allocating memory for the truncated characteristic tensor is a challenge, because its size grows exponentially with $N$. Fortunately, there is a way around this problem. The main idea is that instead of estimating the characteristic tensor of all $N$ variables, we may instead estimate the characteristic tensors of subsets of variables, such as triples, which partially share variables with other triples. The key observation that enables this approach is that the marginal characteristic function of any subset of random variables is also a constrained complex CPD model that inherits parameters from the grand characteristic tensor. Marginalizing with respect to the $n'$-th random variable, we have that  
\begin{align}
   \underline{\boldsymbol{\Phi}} (k_1,\ldots,k_{n'}=0,\ldots,k_N) &=\sum\limits_{h=1}^F \prod\limits_{\substack{n=1\\n\neq n'}}^N\Phi_{X_n|H}[k_n]{\underbrace{\Phi_{X_{n'}|H}[0]}_\text{=1}}\nonumber\\ 
  &=\sum\limits_{h=1}^F \prod\limits_{\substack{n=1\\n\neq n'}}^N\Phi_{X_n|H}[k_n].
\end{align}
Thus, a characteristic function of any subset  of  three random  variables $X_i,X_j,X_\ell$ (triples) can  be written as a third-order tensor, $\underline{\boldsymbol{\Phi}}_{ij\ell}$, of rank $F$. These sub-tensors can be jointly decomposed in a coupled fashion to obtain the sought factors that allow synthesizing the big characteristic tensor, or decomposed independently and `stitched' later under more stringent conditions. Either way, we beat the curse of dimensionality for low-enough model ranks. In addition to affording significant computational and memory reduction, unlike neural network based methods, the above approach allows us to work with fewer and even missing data during the training phase, i.e., only having access to incomplete realizations of the random vector of interest. We estimate lower-order characteristic function values from only those realizations that all three random variables in a given triple appear together. 

In earlier work, we proposed a similar approach for the categorical case where every random variable is finite-alphabet and the task is to estimate the joint probability mass function (PMF) \cite{Kargas2018Kolmogorov}. There we  showed  that every joint PMF of a finite-alphabet random vector can be represented by a naïve Bayes model with a finite number of latent states (rank). If the rank is low, the high dimensional joint PMF is almost surely identifiable from lower-order marginals -- which is reminiscent of Kolmogorov extension. 

In case of continuous random variables, however, the joint PDF can no longer be directly represented by a tensor. One possible solution could be discretization, but this unavoidably leads to discretization error. In this work, what we show is that we can {\em approximately} represent any smooth joint PDF (and evaluate it at any point) using a low-rank tensor {\em in the characteristic function domain}, thereby avoiding discretization loss altogether. 

Our joint PDF model enables easy computation of any marginal or conditional density of subsets of variables of ${\boldsymbol X}$. Using the conditional expectation, the response variable, taken without loss of generality to be the last variable $X_N$, can be estimated in the following way (see detailed derivation in Section ~\ref{SM}.).
\begin{align}
\label{reg}
& E \left[  X_N|X_1, \ldots, X_{N-1} \right]= \frac{1}{c_1} \sum_{h=1}^F \boldsymbol{\lambda}(h)\prod_{n=1}^{N-1}{\sum_{k_n=-K}^K}\nonumber \\
&\quad\quad\mathbf{A}_n(k_n,h)e^{-j2\pi k_n x_n}{\sum_{k_N=-K}^K}c_{2,k_N}\mathbf{A}_N(k_N,h) \\
&\text{where } c_1=\sum_{h=1}^F \boldsymbol{\lambda}(h)\prod_{n=1}^{N-1}{\sum_{k_n=-K}^K}\mathbf{A}_n(k_n,h)e^{-j2\pi k_n x_n},\nonumber \\
&\text{and } c_{2,k_N}=\frac{e^{-j2\pi k_N}}{-j2\pi k_N}+\frac{1-e^{-j2\pi k_N}}{{[-j2\pi k_N}]^2}. \nonumber 
\end{align}

One of the very appealing properties of the proposed approach is that it is a generative model that affords easy sampling. According to Equation~\eqref{eq:density_cpd}, a sample of the multivariate distribution can be generated by first drawing $H$ according to $p_H$ and then independently drawing samples for each variable $X_n$ from the conditional PDF $f_{X_n|H}$. The resulting generative model can be visualized in Figure~\ref{NB}.

\subsection{Algorithm: Coupled Tensor Factorization}
\label{sec:algo}
\begin{algorithm}
   \caption{\textbf{Low-Rank Characteristic Function based Density Estimation (LRCF-DE).}}
   \label{alg:Algo}
\begin{algorithmic}
   \STATE {\textbf{Input}:} A real-valued dataset $D \in\mathbb{R}^{N\times M}$, parameters $F,K$.
   \STATE {\textbf{Output}:} The joint PDF model ${{f}}_{{\boldsymbol X}}$.
   \STATE  Compute $ \underline{\boldsymbol{\Phi}}_{{i j\ell}} \forall i,j,\ell \in \{1,\ldots,N\},~\ell>j>i$ from training data.
   \STATE  Initialize ${\boldsymbol{\lambda}, \mathbf{A}_{1},\ldots,\mathbf{A}_{N}}$ in compliance with their constraints.
   \REPEAT
   \FORALL{$n \in \{1,\ldots,N\}$}
   \STATE Solve the optimization problem defined in (\ref{eqn:ls_prob}).
   \ENDFOR
   \STATE Solve the optimization problem defined in (\ref{eqn:simplex_prob}).
   \UNTIL{convergence criterion satisfied}
   \STATE Assemble the joint PDF as in equation (\ref{eq:pdf_est_factors}).
\end{algorithmic}
\end{algorithm}
We formulate the problem as a coupled complex tensor factorization problem and propose a Block Coordinate Descent algorithm for recovering the latent factors
of the CPD model representing the joint CF. Then, we only need to invert each conditional CF and synthesize the joint PDF. We refer to this approach as Low-Rank Characteristic Function based Density Estimation (LRCF-DE). 

We begin by defining the following coupled tensor factorization problem
\begin{equation}
\begin{aligned}
\min_{\boldsymbol{\lambda}, \mathbf{A}_{1},\ldots,\mathbf{A}_{N}} &
\sum_{i}\sum_{j>i}\sum_{\ell>j} \left \| \underline{\boldsymbol{\Phi}}_{ij\ell}- [\![\boldsymbol{\lambda}, \mathbf{A}_{i},\mathbf{A}_{j}, \mathbf{A}_{\ell} ]\!]  \right \|_F^2               \\ 
\text{subject to}  \quad &\boldsymbol{\lambda}\geq \boldsymbol{0}, \boldsymbol{1}^{T}\boldsymbol{\lambda} = 1, \\ 
&{ \mathbf{A}_n(K+1,:)=\boldsymbol{1}^{T},~  n=1,\ldots, N}.
\label{eq:optimization}
\end{aligned}
\end{equation}
Each lower-dimensional joint CF of triples,   $\underline{\boldsymbol{\Phi}}_{ij\ell}$, can be computed directly from the observed data via sample averaging according to equation \eqref{eq:sample_avgs}.
The formulated optimization problem \eqref{eq:optimization} is non-convex and NP-hard. However it becomes convex with respect to each variable if we fix the remaining ones and can be handled using alternating optimization. By using the mode-$1$ matrix unfolding of each tensor $\underline{\boldsymbol{\Phi}}_{ij\ell}$, the optimization problem with respect to $\mathbf{A}_{i}$ becomes
\begin{equation}
\begin{aligned}
 & \min_{\mathbf{A}_{i}}
\quad \sum_{j \neq i} \sum_{ \ell \neq i, \ell > j} 
\left \| \underline{\boldsymbol{\Phi}}_{ij\ell}^{(1)} - (\mathbf{A}_{\ell} \odot \mathbf{A}_j) \textrm{diag} (\boldsymbol{\lambda}) \mathbf{A}_i^T
\right \|_F^2  \\ 
& \text{subject to} \quad {\mathbf{A}_i(K+1,:)=\boldsymbol{1}^{T}}.
\end{aligned}
\label{eqn:ls_prob}
\end{equation}
The exact update for each factor $\mathbf{A}_i$ can be computed as
\begin{equation} 
{\mathbf{A}_i}\leftarrow{{\mathbf G}_i}^{-1}{\mathbf V}_i,
\end{equation}
where
\begin{equation*}
\begin{aligned}
&{{\mathbf G}_i}=({\boldsymbol{\lambda}}{\boldsymbol{\lambda}}^T) \circledast \sum_{j \neq i} \sum_{ \ell \neq i, \ell > j} \mathbf{Q}_{\ell j}^H{\mathbf Q}_{\ell j}, \\
&{\mathbf V}_i= \textrm{diag}(\boldsymbol{\lambda}) \sum_{j \neq i} \sum_{ \ell \neq i, \ell > j}  \mathbf{Q}_{\ell j}^H\underline{\boldsymbol{\Phi}}_{{ij\ell}}^{(1)},  \\
&{\mathbf Q}_{\ell j} = {\mathbf{A}_{\ell}}\odot{\mathbf{A}_{j}}. 
\end{aligned}
\end{equation*}
For each update, the row of $\mathbf{A}_i$ that corresponds to zero frequency is removed and updating $\mathbf{A}_i$ becomes an unconstrained complex least squares problem. A vector of ones is appended at the same row index after each update $\mathbf{{A}}_i$. Due to role symmetry the same form holds for each factor $\mathbf{A}_n$. 

Now, for the $\boldsymbol{\lambda}$-update we solve the following optimization problem
\begin{equation}
\begin{aligned}
& \min_{\boldsymbol{\lambda}} 
\quad\sum\limits_{i}\sum\limits_{j>i}\sum\limits_{\ell>j} \left \| { \textrm{vec}(\underline{\boldsymbol{\Phi}}_{ij\ell})- ({\mathbf{A}_\ell}\odot \mathbf{A}_j\odot \mathbf{A}_i) \boldsymbol{\lambda}} \right \|_F^2 \\ 
&  { \text{subject to} } \ {\quad \boldsymbol{\lambda}\geq \mathbf{0},~ \mathbf{1}^{T}\boldsymbol{\lambda} = 1. } 
\end{aligned}
\label{eqn:simplex_prob}
\end{equation}
The optimization problem~\eqref{eqn:simplex_prob} is a least squares problem with a probability simplex constraint. We use an ADMM algorithm to tackle it. Towards this end, we reformulate the optimization problem by introducing an auxiliary variable $\hat{\boldsymbol{\lambda}}$ and rewrite the problem equivalently as
\begin{equation*} 
\begin{aligned}
&{ \min_{{\boldsymbol{\lambda}}, \hat{\boldsymbol{\lambda}}}  }  
\quad f  (\hat{\boldsymbol{\lambda}} )+r({\boldsymbol{\lambda}}) \\ 
&{ \text{subject to} }~{{\hat{\boldsymbol{\lambda}}}^T={\boldsymbol{\lambda}}},
\end{aligned}
\end{equation*}
where, $f(\hat{\boldsymbol{\lambda}})= \sum\limits_{i}\sum\limits_{j>i}\sum\limits_{\ell>j}  \| \textrm{vec}(\underline{\boldsymbol{\Phi}}_{ij \ell})- ({\mathbf{A}_\ell}\odot \mathbf{A}_j\odot \mathbf{A}_i) \hat{\boldsymbol{\lambda}}  \|_F^2$ 
and 
$r({\boldsymbol{\lambda}})$ 
is the indicator function for the probability simplex. $C=\{{\boldsymbol{\lambda}}|{\boldsymbol{\lambda}}\geq \mathbf{0}, \mathbf{1}^{T}{\boldsymbol{\lambda}}= 1\}$,
\begin{equation*}
r({\boldsymbol{\lambda}}) =
\begin{cases}
0,& {\boldsymbol{\lambda}}\in C\\
\infty,& {\boldsymbol{\lambda}}\not\in C.
\end{cases}
\end{equation*}
At each iteration $\tau$, we perform the following updates
\begin{equation}
\begin{aligned}
& \hat{\boldsymbol{\lambda}}^{\tau+1}\leftarrow {(\mathbf{G}+\rho \mathbf{I})}^{-1}(\mathbf{V}+\rho({\boldsymbol{\lambda}}^{\tau}+{\mathbf{u}}^{\tau})) \\
&{{\boldsymbol{\lambda}}}^{\tau+1}\leftarrow \mathcal{P}_C({{\boldsymbol{\lambda}}^{\tau}-\hat{\boldsymbol{\lambda}}}^{\tau+1} + \mathbf{u}^{\tau}) \\
&{{\mathbf{u}}}^{\tau+1}\leftarrow {\mathbf{u}}^{\tau}+{{\boldsymbol{\lambda}}^{\tau+1}-\hat{\boldsymbol{\lambda}}}^{\tau+1},\\
\text{where}\\
&\mathbf{G}= \sum_{i}\sum_{j>i}\sum_{\ell>j} \mathbf{Q}_{\ell j i}^H\mathbf{Q}_{\ell ji},\\
&\mathbf{V}= \sum_{i}\sum_{j>i}\sum_{\ell>j} \mathbf{Q}_{\ell j i}^H \textrm{vec}(\underline{\boldsymbol{\Phi}}),\\
&\mathbf{Q}_{\ell ji} = {\mathbf{A}_{\ell }}\odot{\mathbf{A}_{j}\odot{\mathbf{A}_{i}}}.
\end{aligned}
\end{equation}
$\mathcal{P}_C(\mathbf{y})$ denotes the projection operator onto the convex set $C$~-- it computes the Euclidean projection of the real part of a point $\mathbf{y}= \left[y_1,\ldots, y_F\right]^T\in {\mathbb{C}}^F$ onto the probability simplex
\begin{equation*} 
\begin{aligned}
{ \displaystyle \min_{\mathbf{x} \in {\mathbb{R}}^F}\displaystyle } & 
\frac{1}{2} \| \mathbf{x}-\Re(\boldsymbol{y}) \| _F^2 \\ { \text{subject to} } \ & { \quad \mathbf{x}\geq \mathbf{0},~\mathbf{1}^{T}\mathbf{x} = 1, }
\end{aligned}
\end{equation*} 
using the method described in~\cite{wang2013projection}. The overall procedure is described in Algorithm
\ref{alg:Algo}.

As the final step, the factors are assembled from the triples and the joint CF over all variables is synthesized as $\underline{\boldsymbol{\Phi}} = [\![ \boldsymbol{\lambda},\mathbf{A}_1,\ldots,\mathbf{A}_N]\!]$. Given, the model of the joint CF, the corresponding joint PDF model can be recovered at any point as
\begin{equation}
{{f}}_{{\boldsymbol X}}({\mathbf x})=\sum_{h=1}^F \boldsymbol{\lambda}(h)\prod_{n=1}^N{\sum_{k_n=-K}^K}\mathbf{A}_n(k_n,h)e^{-j2\pi k_n x_n}.\label{eq:pdf_est_factors}
\end{equation}

\begin{table*}[!ht]
\begin{center}
\resizebox{0.8\textwidth}{!}{
\begin{tabular}{lccccc}
\toprule
Data set & MoG & KDE & RNADE & MAF & LRCF-DE\\
\midrule  
Red wine    &$11.9\pm0.29$ &$9.9\pm0.16$&$14.41\pm0.16$& $\mathbf{15.2\pm0.09}$&$\mathbf{16.4\pm0.67}$\\
White wine &$16.1\pm1.48$ &$14.8\pm0.12$&$17.1\pm0.26$& $\mathbf{17.3\pm0.20}$&$\mathbf{18.4\pm0.17}$\\
F-O.TP       &$125.4\pm7.79$ & $103.05\pm0.84$ &$\mathbf{152.48\pm5.62}$ &$149.6\pm8.32$&$\mathbf{154.34\pm8.43}$\\
PCB    &$152.9\pm3.88$ &$147.6\pm1.63$&$171.7\pm2.75$& $\mathbf{179.6\pm1.62}$&$\mathbf{194.4\pm2.43}$\\
Superconductivty &$134.7\pm3.47$ & $127.2\pm2.82$ & $140.2\pm1.03$ & $\mathbf{143.5\pm1.32}$ & $\mathbf{146.1\pm2.31}$\\
Corel Images &$211.7\pm1.04$ & $201.4\pm1.18$ & $\mathbf{223.6\pm0.88}$ &$218.2\pm1.35$&$\mathbf{222.6\pm1.25}$\\
Gas Sensor &$310.3\pm3.47$&$296.48\pm1.62$&$\mathbf{316.3\pm3.57}$&$315.4\pm1.458$&$\mathbf{316.6\pm2.35}$\\
\bottomrule
\end{tabular}}
\end{center}
\caption{Average test-set log-likelihood per datapoint for 5 different
models on UCI datasets; \textbf{higher is better}.}
\label{table:method_evaluation}
\end{table*}

\section{Experiments}
\subsection{Low Dimensional Toy Data}

We first show motivating results from modeling low dimensional datasets and showcase the expressivity of the proposed model as well as the significance of each parameter. We begin by modeling  $M=2000$ samples from the Weight-Height dataset. In Figure~\ref{toy_weight},  we present $M'=1500$ synthetic samples obtained from the proposed model for different smoothing parameters $K\in[1,2,3,4,5]$ and ranks $F \in[2,4,6,8,10]$. By judiciously selecting the parameter search space, our approach yields an expressive generative model that can well-represent the data.

Following the same procedure, we now visualize $M=2000$ samples from the $2$-D Moons and Circles datasets. We fix the number of smoothing coefficients $K$, $K=11$, and visualize synthetic $M'=1500$ samples obtained from the proposed model for different approximation ranks $F \in[1,2,3,4,6]$ in Figure~\ref{fig:2d}. The results show that our model is able to capture complex structures and properties of the data for an appropriate choice of rank $F$.

\subsection{Real Data}

We test the proposed approach on datasets (see a brief description of the datasets in Table~\ref{table:info}) obtained from the UCI machine learning repository~\cite{lichman2013uci}.

\begin{table}[t]
\begin{center}
\resizebox{0.4\textwidth}{!}{
\begin{tabular}{lcc}
\toprule
Data set & N & M\\
\midrule
Red wine &11&1599\\
White wine &11&4898\\
First-order theorem proving (F-O.TP) &51&6118\\
Polish companies bankruptcy (PCB) &64&10503\\
Superconductivty &81&21263\\
Corel Images &89&68040\\
Gas Sensor Array Drift (Gas Sensor) &128&13910\\
\bottomrule
\end{tabular}}
\end{center}
\caption{Dataset information.}
\label{table:info}
\end{table}

For each dataset we randomly hide $20\%$ of data (testing set) and consider the remaining entries as observed information (training set). The parameters, which include the tensor rank $F$ and the smoothing parameter $K$, are chosen using cross-validation. We use  $20\%$ of the training data as validation data, where we seek to find the optimal parameter values maximizing the average log-likelihood of the validation samples. Once the hyperparameters are chosen, we train the model using all the training data (including the validation data) and measure its performance
on the testing set. We compare our approach against standard baselines described in section~\ref{sec:related_work}.

Evaluating the quality of density models is an open and difficult problem \cite{theis2015note}. Following the approach in~\cite{uria2013rnade,papamakarios2017masked}, we calculate and report the average log-likelihood of unseen data samples (testing set), further averaged over 5 random data splits. The results are shown in Table~\ref{table:method_evaluation}.
LRCF-DE has a higher average test sample log likelihood on almost all datasets. Overall, we observe that our method outperforms the baselines in $4$ datasets and is comparable to the winning method in the remaining ones.

 Following the derivation in  Equation~\eqref{reg}, we test the proposed model in several regression tasks. We evaluate and report the Mean Absolute Error (MAE) in estimating $X_N$ for the unseen data samples in Table~\ref{table:mae} and additional results for multi-output regression are presented in Table~\ref{table:multio}. Overall, we observe that LRCF-DE outperforms the baselines on almost all datasets, and performs comparable to the winning method in the remaining ones. 
 
 We have to stress again the fact that neural network based density estimation methods evaluate multivariate densities point-wise. These methods cannot impute more than a few missing elements in the input as grid search becomes combinatorial. Due to the interpretation of the approximation of the sought density as a finite mixture of separable densities and the coupled tensor factorization approach, our method allows us to easily work with missing data during both training and testing. Here, we showcase the results of LRCF-DE against MAF for simultaneously predicting the last two random variables of each dataset given the remaining ones.
\begin{table}
\begin{center}
\resizebox{0.45\textwidth}{!}{
\begin{tabular}{lccccc}
\toprule
Data set & MoG & KDE & RNADE & MAF & LRCF-DE\\
\midrule
Red wine    &1.28&1.13&0.66&0.63&0.56\\
White wine &1.79&1.31&0.80 &0.75&0.59\\
F-O.TP  &1.86&1.46&0.63& 0.52&0.48\\
PCB  &5.6& 7.73&  4.43& 4.52& 3.85\\
Superconductivty     &18.56&19.96&16.46& 16.38&16.53\\
Corel Images    &0.53&0.93&0.27&0.27&0.28\\
Gas Sensor     &29.7&35.3&26.8&26.2&26.7\\
\bottomrule
\end{tabular}}
\end{center}
\caption{MAE for regression tasks.}
\label{table:mae}
\end{table}
\begin{table}[H]
\begin{center}
\resizebox{0.45\textwidth}{!}{
\begin{tabular}{lcc}
\toprule
Data set & LRCF-DE & MAF\\
\midrule
Red wine &0.82&0.91\\
White wine &0.93&0.97\\
First-order theorem proving (F-O.TP) &0.69&0.72\\
Polish companies bankruptcy (PCB) &4.97&5.46\\
Superconductivty &20.84&20.72\\
Corel Images &1.36&1.59\\
Gas Sensor Array Drift (Gas Sensor) &25.7&26.1\\
\bottomrule
\end{tabular}}
\end{center}
\caption{MAF for multi-output regression tasks.}
\label{table:multio}
\end{table}

\begin{figure}
\centering
\subfigure{
    \begin{tabular}{@{}cccccccc@{}}
    \includegraphics[width=.022\textwidth]{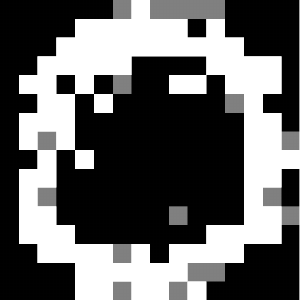} &
     \hspace{-0.4cm}
    \includegraphics[width=.022\textwidth]{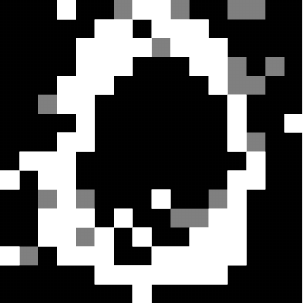} &
     \hspace{-0.4cm}
    \includegraphics[width=.022\textwidth]{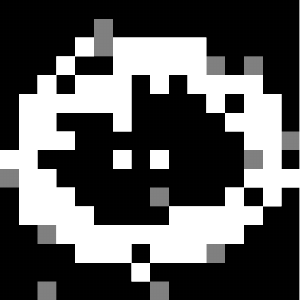} &
     \hspace{-0.4cm}
    \includegraphics[width=.022\textwidth]{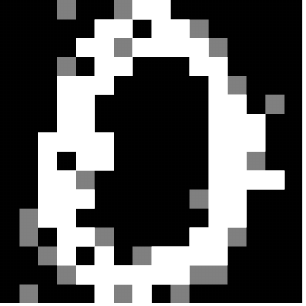} &
    \hspace{-0.4cm}
    \includegraphics[width=.022\textwidth]{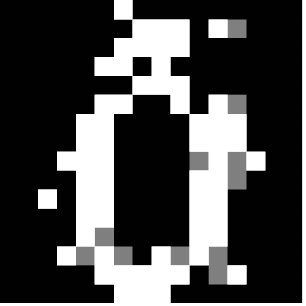} &
    \hspace{-0.4cm}
    \includegraphics[width=.022\textwidth]{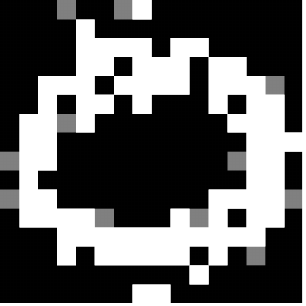} &
    \hspace{-0.4cm}
    \includegraphics[width=.022\textwidth]{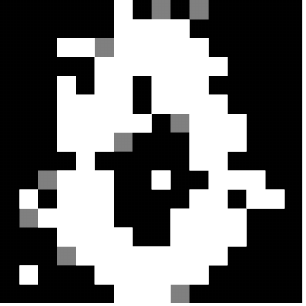} &
    \hspace{-0.4cm}
    \includegraphics[width=.022\textwidth]{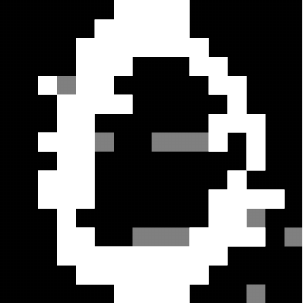} 
    \\
    \includegraphics[width=.022\textwidth]{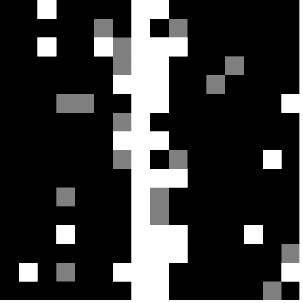} &
      \hspace{-0.4cm}
    \includegraphics[width=.022\textwidth]{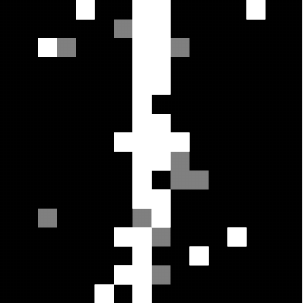} &
      \hspace{-0.4cm}
    \includegraphics[width=.022\textwidth]{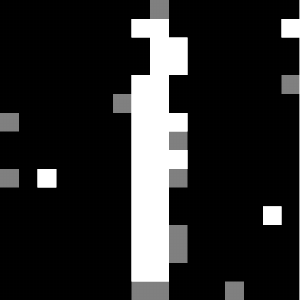} &
      \hspace{-0.4cm}
    \includegraphics[width=.022\textwidth]{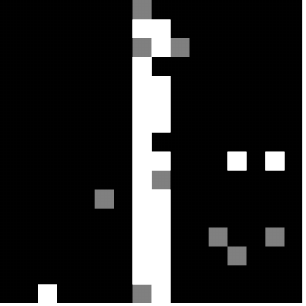} &
      \hspace{-0.4cm}
    \includegraphics[width=.022\textwidth]{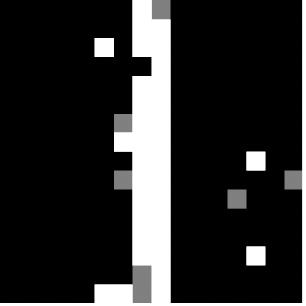} &
      \hspace{-0.4cm}
    \includegraphics[width=.022\textwidth]{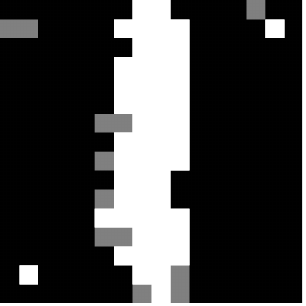} &
      \hspace{-0.4cm}
    \includegraphics[width=.022\textwidth]{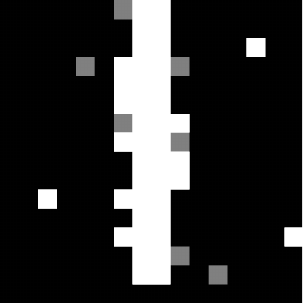} &
      \hspace{-0.4cm}
    \includegraphics[width=.022\textwidth]{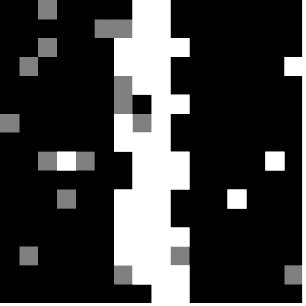}    
    \\
     \includegraphics[width=.022\textwidth]{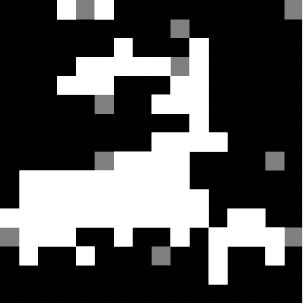} &
     \hspace{-0.4cm}
    \includegraphics[width=.022\textwidth]{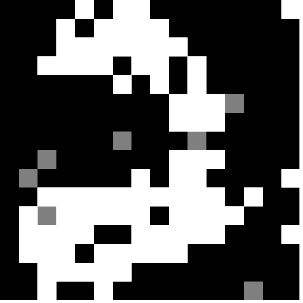} &
     \hspace{-0.4cm}
    \includegraphics[width=.022\textwidth]{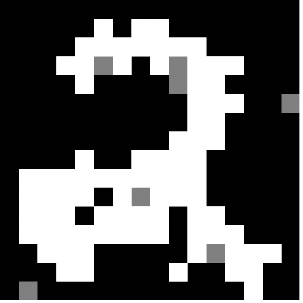} &
     \hspace{-0.4cm}
    \includegraphics[width=.022\textwidth]{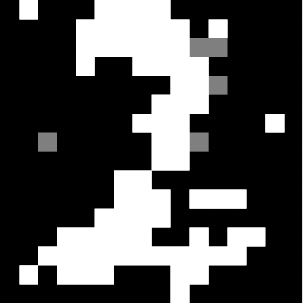} &
     \hspace{-0.4cm}
    \includegraphics[width=.022\textwidth]{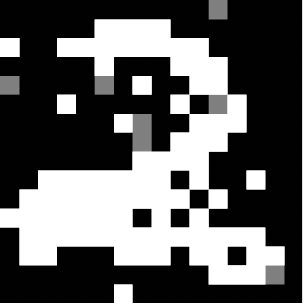} &
    \hspace{-0.4cm}
    \includegraphics[width=.022\textwidth]{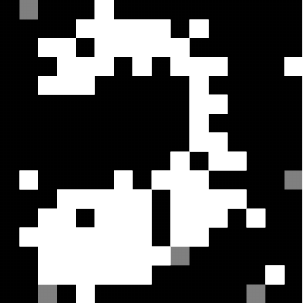} &
     \hspace{-0.4cm}
    \includegraphics[width=.022\textwidth]{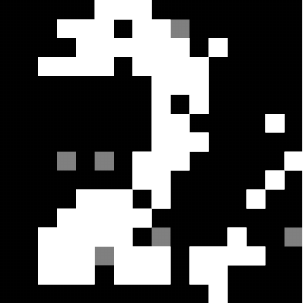} &
     \hspace{-0.4cm}
    \includegraphics[width=.022\textwidth]{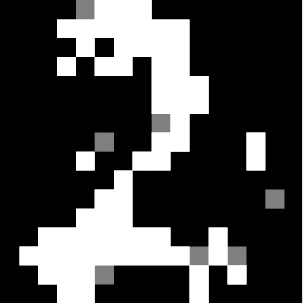}
    \\
     \includegraphics[width=.022\textwidth]{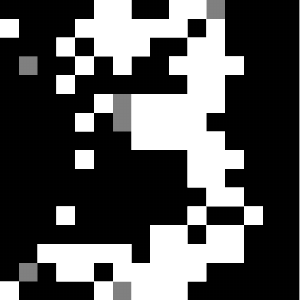} &
     \hspace{-0.4cm}
    \includegraphics[width=.022\textwidth]{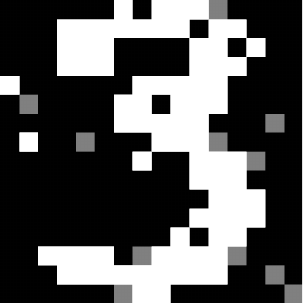} &
     \hspace{-0.4cm}
    \includegraphics[width=.022\textwidth]{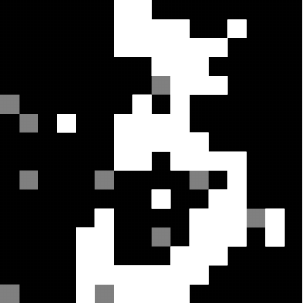} &
     \hspace{-0.4cm}
    \includegraphics[width=.022\textwidth]{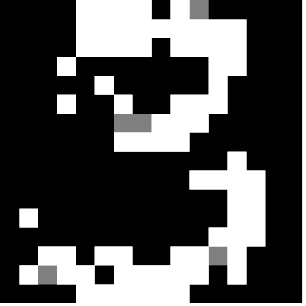} &
     \hspace{-0.4cm}
    \includegraphics[width=.022\textwidth]{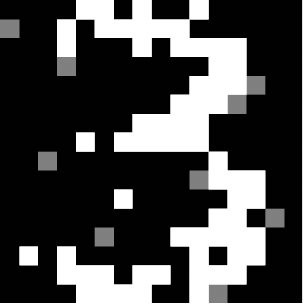} &
    \hspace{-0.4cm}
    \includegraphics[width=.022\textwidth]{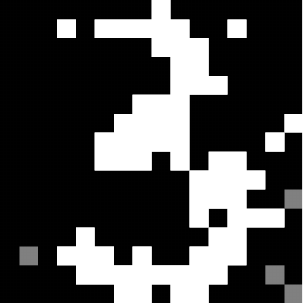} &
     \hspace{-0.4cm}
    \includegraphics[width=.022\textwidth]{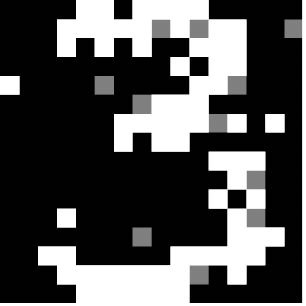} &
     \hspace{-0.4cm}
    \includegraphics[width=.022\textwidth]{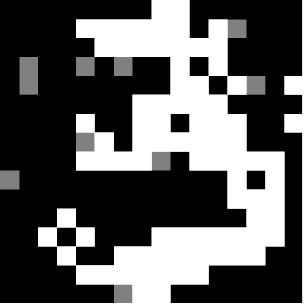}
    \\
    \includegraphics[width=.022\textwidth]{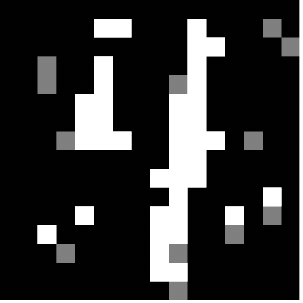} &
     \hspace{-0.4cm}
    \includegraphics[width=.022\textwidth]{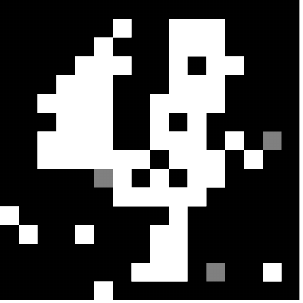} &
     \hspace{-0.4cm}
    \includegraphics[width=.022\textwidth]{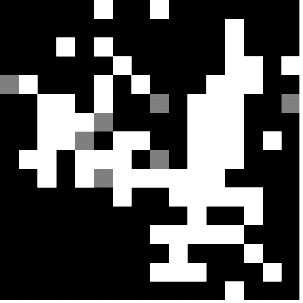} &
     \hspace{-0.4cm}
    \includegraphics[width=.022\textwidth]{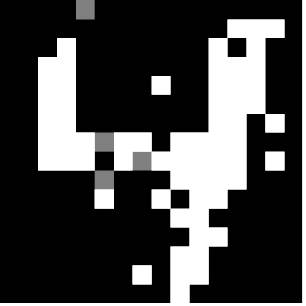} &
     \hspace{-0.4cm}
    \includegraphics[width=.022\textwidth]{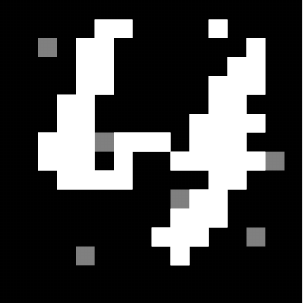} &
    \hspace{-0.4cm}
    \includegraphics[width=.022\textwidth]{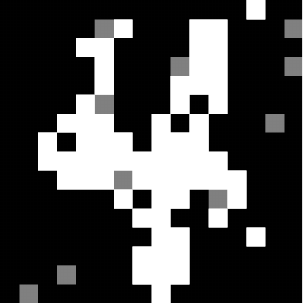} &
     \hspace{-0.4cm}
    \includegraphics[width=.022\textwidth]{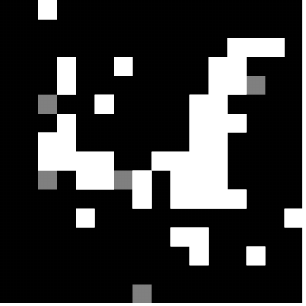} &
     \hspace{-0.4cm}
    \includegraphics[width=.022\textwidth]{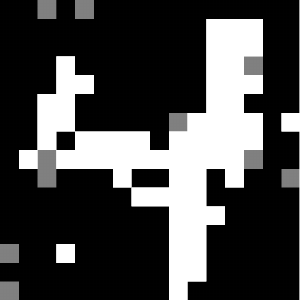}
    \\
    \includegraphics[width=.022\textwidth]{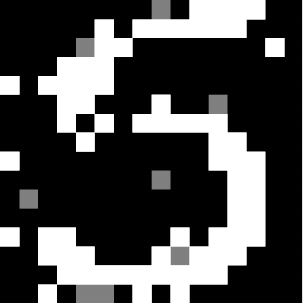} &
     \hspace{-0.4cm}
    \includegraphics[width=.022\textwidth]{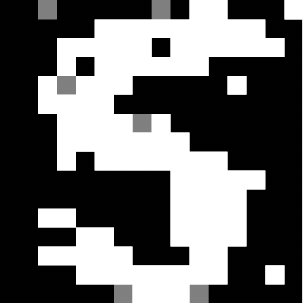} &
     \hspace{-0.4cm}
    \includegraphics[width=.022\textwidth]{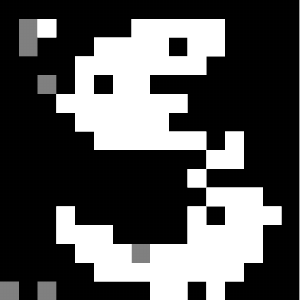} &
     \hspace{-0.4cm}
    \includegraphics[width=.022\textwidth]{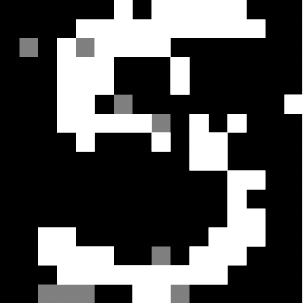} &
     \hspace{-0.4cm}
    \includegraphics[width=.022\textwidth]{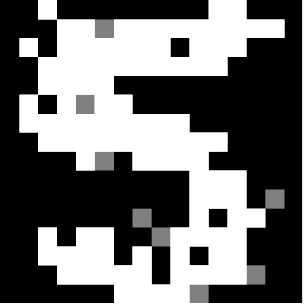} &
         \hspace{-0.4cm}
    \includegraphics[width=.022\textwidth]{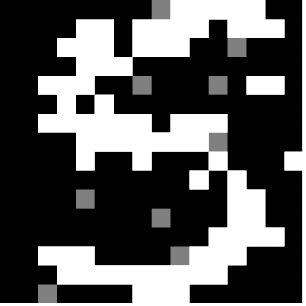} &
     \hspace{-0.4cm}
    \includegraphics[width=.022\textwidth]{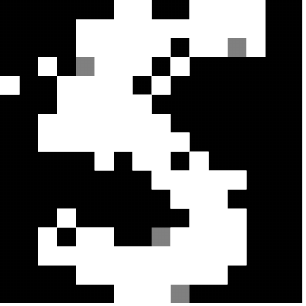} &
     \hspace{-0.4cm}
    \includegraphics[width=.022\textwidth]{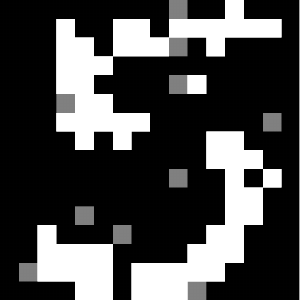}
    \\
     \includegraphics[width=.022\textwidth]{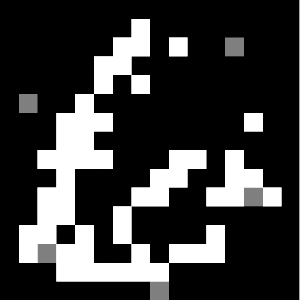} &
     \hspace{-0.4cm}
    \includegraphics[width=.022\textwidth]{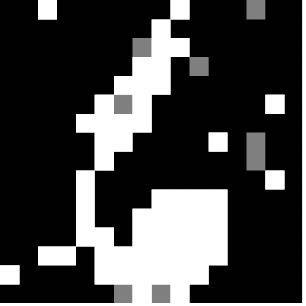} &
     \hspace{-0.4cm}
    \includegraphics[width=.022\textwidth]{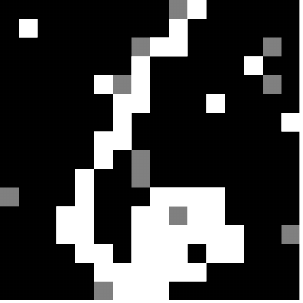} &
     \hspace{-0.4cm}
    \includegraphics[width=.022\textwidth]{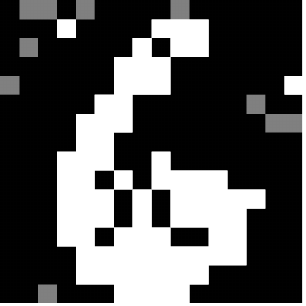} &
     \hspace{-0.4cm}
    \includegraphics[width=.022\textwidth]{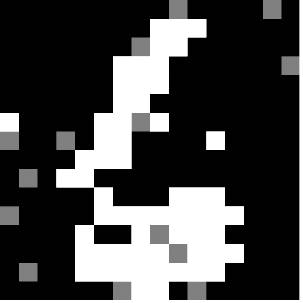} &
    \hspace{-0.4cm}
    \includegraphics[width=.022\textwidth]{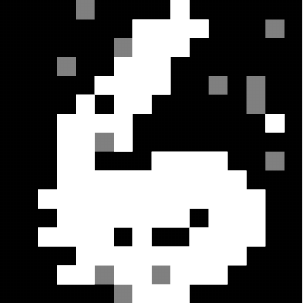} &
     \hspace{-0.4cm}
    \includegraphics[width=.022\textwidth]{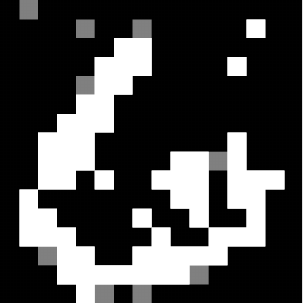} &
     \hspace{-0.4cm}
    \includegraphics[width=.022\textwidth]{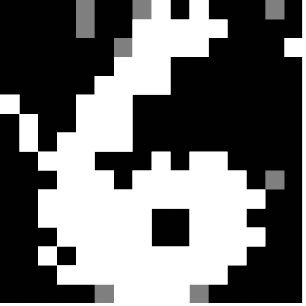}
    \\
     \includegraphics[width=.022\textwidth]{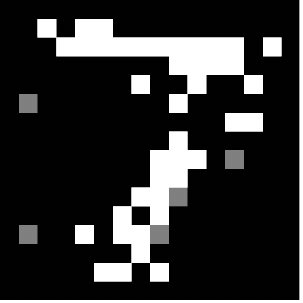} &
     \hspace{-0.4cm}
    \includegraphics[width=.022\textwidth]{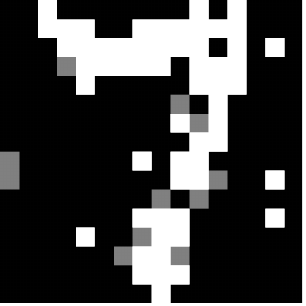} &
     \hspace{-0.4cm}
    \includegraphics[width=.022\textwidth]{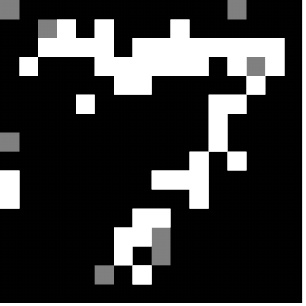} &
     \hspace{-0.4cm}
    \includegraphics[width=.022\textwidth]{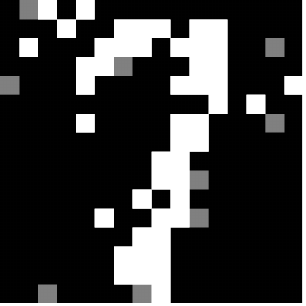} &
     \hspace{-0.4cm}
    \includegraphics[width=.022\textwidth]{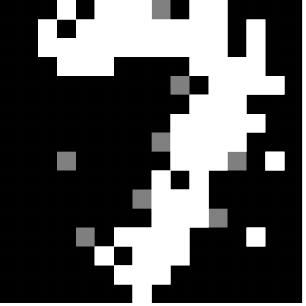} &
         \hspace{-0.4cm}
    \includegraphics[width=.022\textwidth]{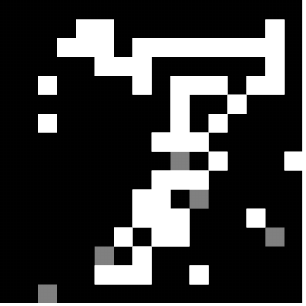} &
     \hspace{-0.4cm}
    \includegraphics[width=.022\textwidth]{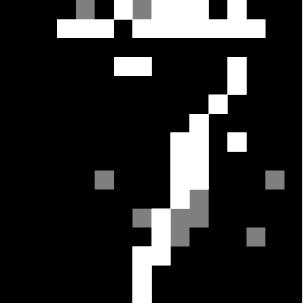} &
     \hspace{-0.4cm}
    \includegraphics[width=.022\textwidth]{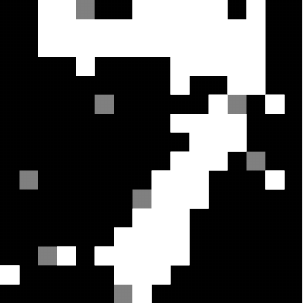}
    \\
     \includegraphics[width=.022\textwidth]{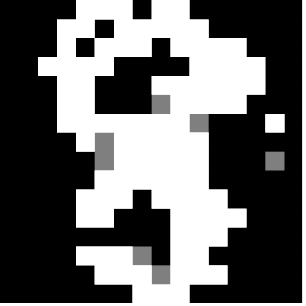} &
     \hspace{-0.4cm}
    \includegraphics[width=.022\textwidth]{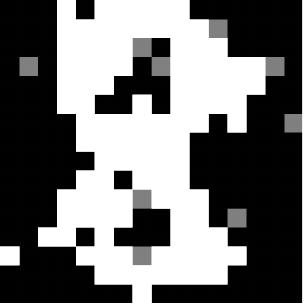} &
     \hspace{-0.4cm}
    \includegraphics[width=.022\textwidth]{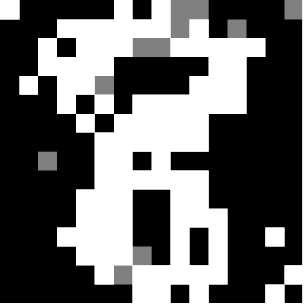} &
     \hspace{-0.4cm}
    \includegraphics[width=.022\textwidth]{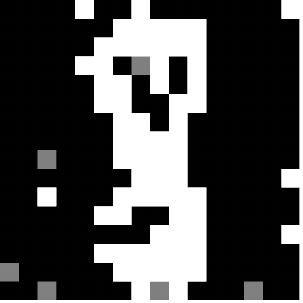} &
     \hspace{-0.4cm}
    \includegraphics[width=.022\textwidth]{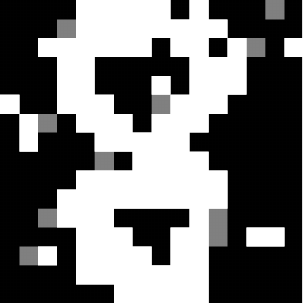} &
    \hspace{-0.4cm}
    \includegraphics[width=.022\textwidth]{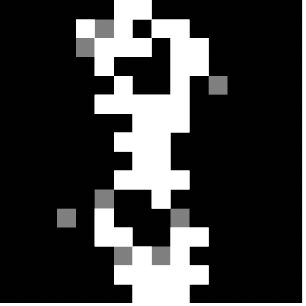} &
     \hspace{-0.4cm}
    \includegraphics[width=.022\textwidth]{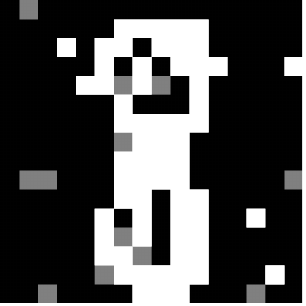} &
     \hspace{-0.4cm}
    \includegraphics[width=.022\textwidth]{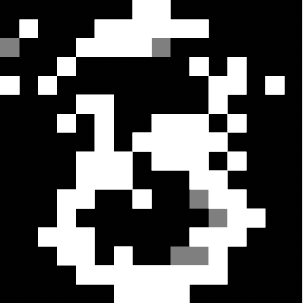} 
    \\
    \includegraphics[width=.022\textwidth]{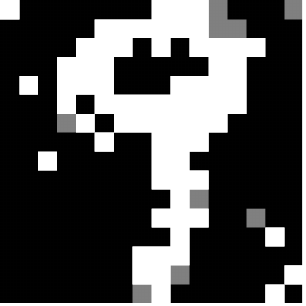} &
     \hspace{-0.4cm}
    \includegraphics[width=.022\textwidth]{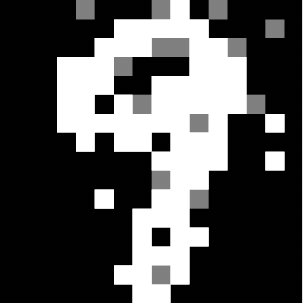} &
     \hspace{-0.4cm}
    \includegraphics[width=.022\textwidth]{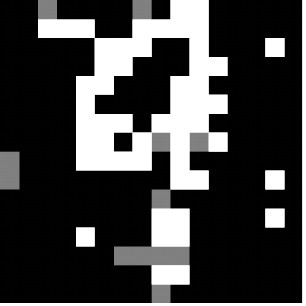} &
     \hspace{-0.4cm}
    \includegraphics[width=.022\textwidth]{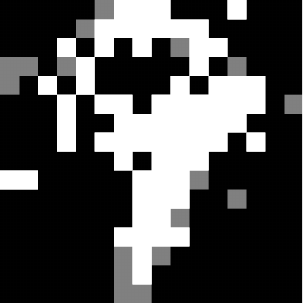} &
      \hspace{-0.4cm}
    \includegraphics[width=.022\textwidth]{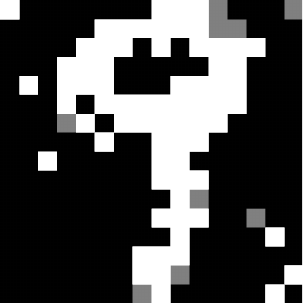} &
         \hspace{-0.4cm}
    \includegraphics[width=.022\textwidth]{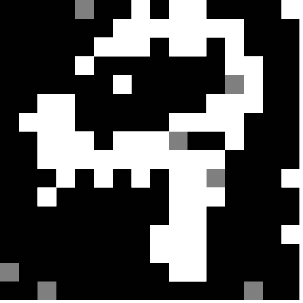} &
     \hspace{-0.4cm}
    \includegraphics[width=.022\textwidth]{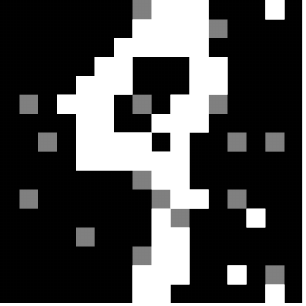} &
      \hspace{-0.4cm}
    \includegraphics[width=.022\textwidth]{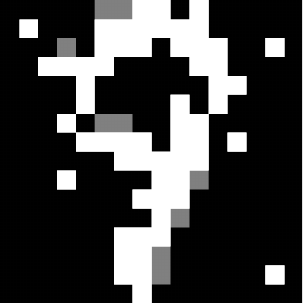}
  \end{tabular}
  }
\hfill
\subfigure{\begin{tabular}{@{}cccccccc@{}}
    \includegraphics[width=.022\textwidth]{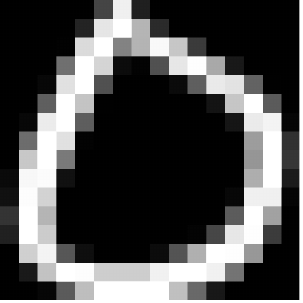} &
     \hspace{-0.4cm}
    \includegraphics[width=.022\textwidth]{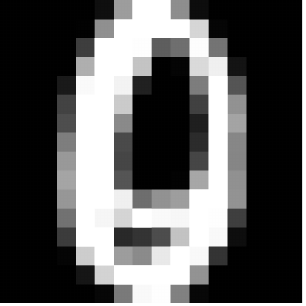} &
     \hspace{-0.4cm}
    \includegraphics[width=.022\textwidth]{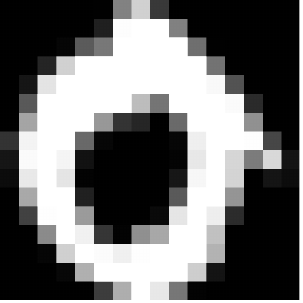} &
     \hspace{-0.4cm}
    \includegraphics[width=.022\textwidth]{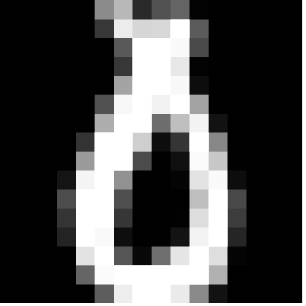} &
     \hspace{-0.4cm}
    \includegraphics[width=.022\textwidth]{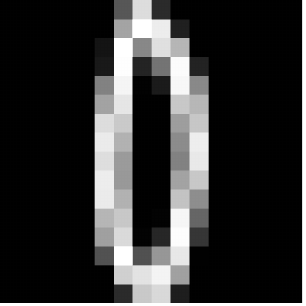} &
     \hspace{-0.4cm}
    \includegraphics[width=.022\textwidth]{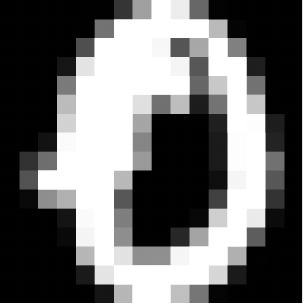} &
     \hspace{-0.4cm}
    \includegraphics[width=.022\textwidth]{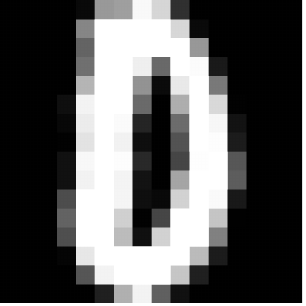} &
      \hspace{-0.4cm}
    \includegraphics[width=.022\textwidth]{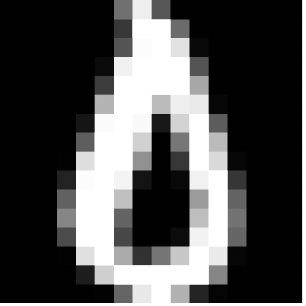}
    \\
    \includegraphics[width=.022\textwidth]{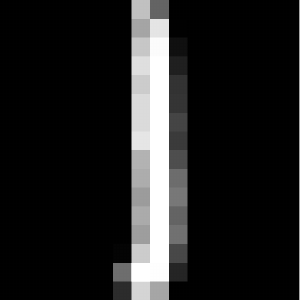} &
      \hspace{-0.4cm}
    \includegraphics[width=.022\textwidth]{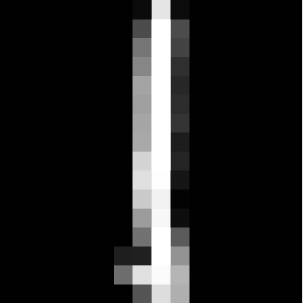} &
      \hspace{-0.4cm}
    \includegraphics[width=.022\textwidth]{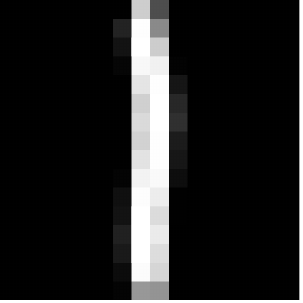} &
      \hspace{-0.4cm}
    \includegraphics[width=.022\textwidth]{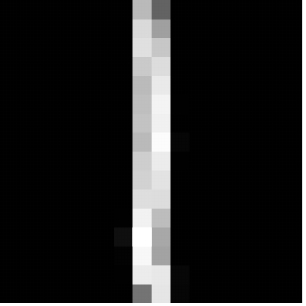} &
      \hspace{-0.4cm}
    \includegraphics[width=.022\textwidth]{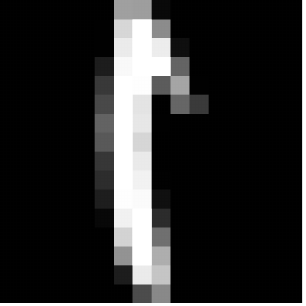} &
     \hspace{-0.4cm}
    \includegraphics[width=.022\textwidth]{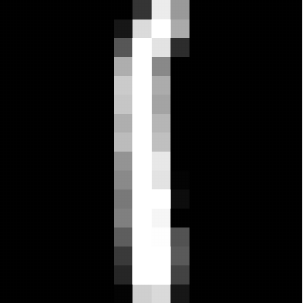} &
      \hspace{-0.4cm}
    \includegraphics[width=.022\textwidth]{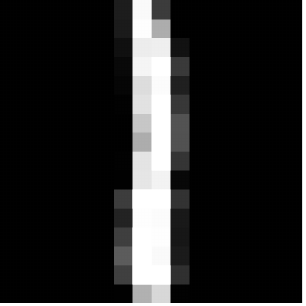} &
      \hspace{-0.4cm}
    \includegraphics[width=.022\textwidth]{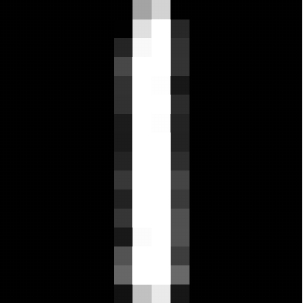}
    \\
     \includegraphics[width=.022\textwidth]{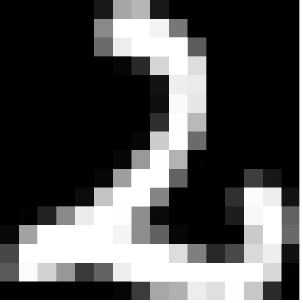} &
     \hspace{-0.4cm}
    \includegraphics[width=.022\textwidth]{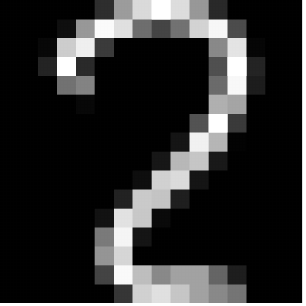} &
     \hspace{-0.4cm}
    \includegraphics[width=.022\textwidth]{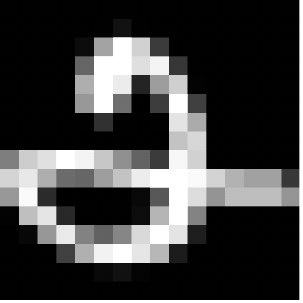} &
     \hspace{-0.4cm}
    \includegraphics[width=.022\textwidth]{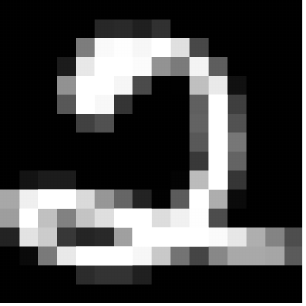} &
     \hspace{-0.4cm}
    \includegraphics[width=.022\textwidth]{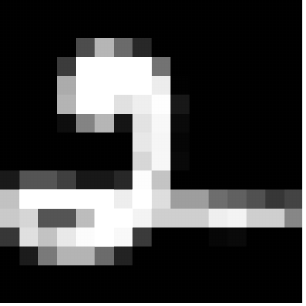} &
         \hspace{-0.4cm}
    \includegraphics[width=.022\textwidth]{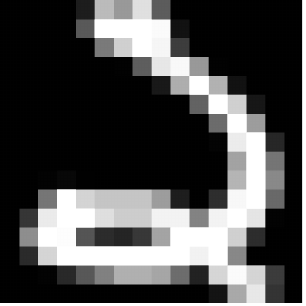} &
     \hspace{-0.4cm}
    \includegraphics[width=.022\textwidth]{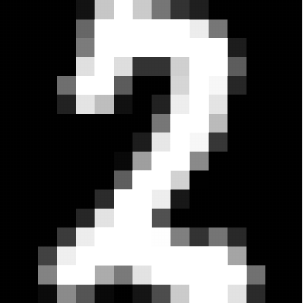} &
     \hspace{-0.4cm}
    \includegraphics[width=.022\textwidth]{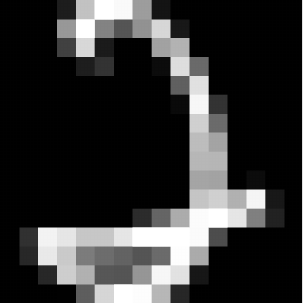} 
     \\
       \includegraphics[width=.022\textwidth]{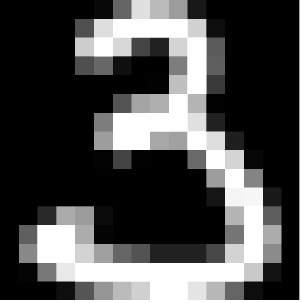} &
     \hspace{-0.4cm}
    \includegraphics[width=.022\textwidth]{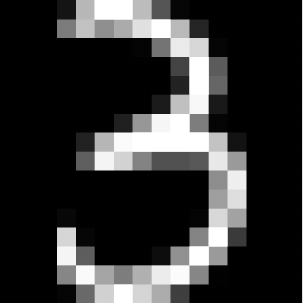} &
     \hspace{-0.4cm}
    \includegraphics[width=.022\textwidth]{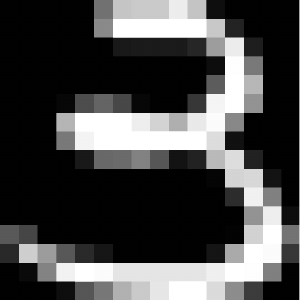} &
     \hspace{-0.4cm}
    \includegraphics[width=.022\textwidth]{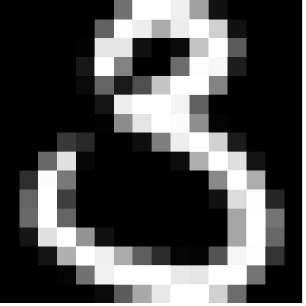} &
     \hspace{-0.4cm}
    \includegraphics[width=.022\textwidth]{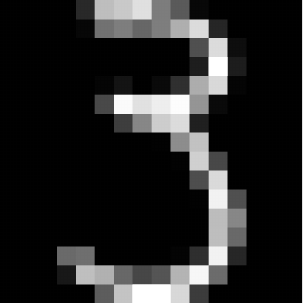} &
         \hspace{-0.4cm}
    \includegraphics[width=.022\textwidth]{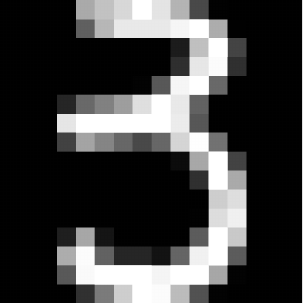} &
     \hspace{-0.4cm}
    \includegraphics[width=.022\textwidth]{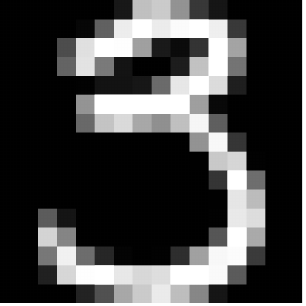} &
     \hspace{-0.4cm}
    \includegraphics[width=.022\textwidth]{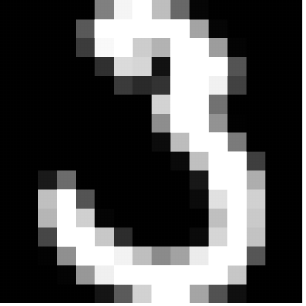}
    \\
      \includegraphics[width=.022\textwidth]{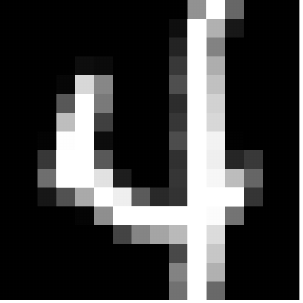}&
     \hspace{-0.4cm}
    \includegraphics[width=.022\textwidth]{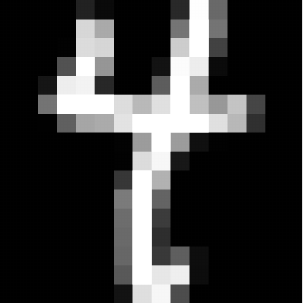} &
     \hspace{-0.4cm}
    \includegraphics[width=.022\textwidth]{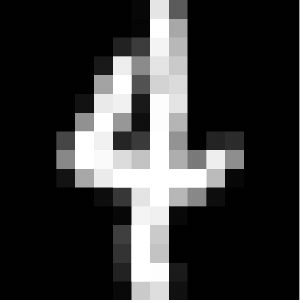} &
     \hspace{-0.4cm}
    \includegraphics[width=.022\textwidth]{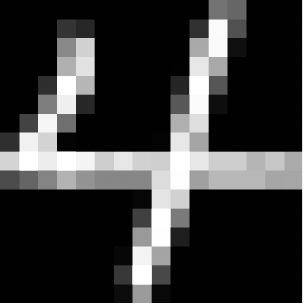} &
     \hspace{-0.4cm}
    \includegraphics[width=.022\textwidth]{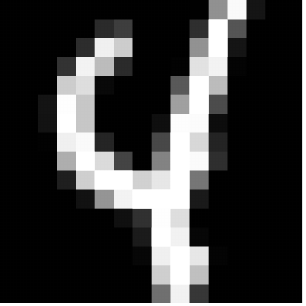} &
    \hspace{-0.4cm}
    \includegraphics[width=.022\textwidth]{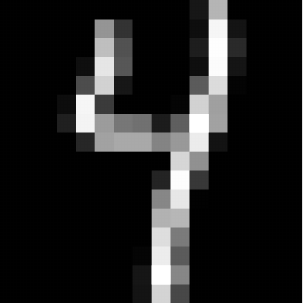} &
     \hspace{-0.4cm}
    \includegraphics[width=.022\textwidth]{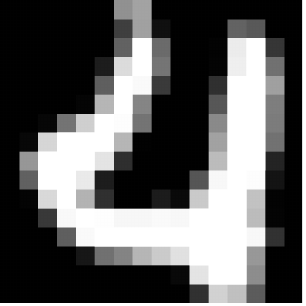} &
     \hspace{-0.4cm}
    \includegraphics[width=.022\textwidth]{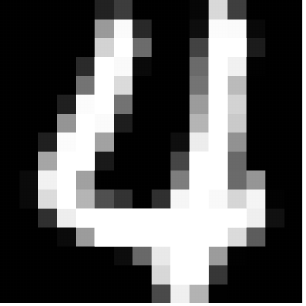}
    \\
     \includegraphics[width=.022\textwidth]{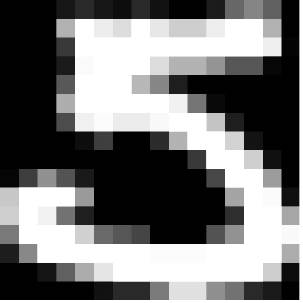} &
     \hspace{-0.4cm}
    \includegraphics[width=.022\textwidth]{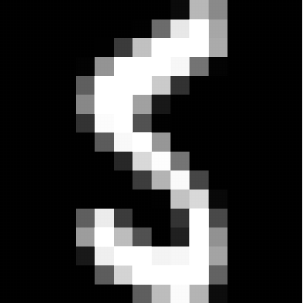} &
     \hspace{-0.4cm}
    \includegraphics[width=.022\textwidth]{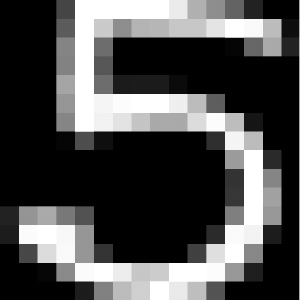} &
     \hspace{-0.4cm}
    \includegraphics[width=.022\textwidth]{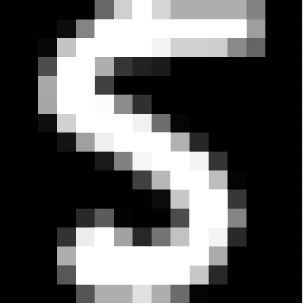} &
     \hspace{-0.4cm}
    \includegraphics[width=.022\textwidth]{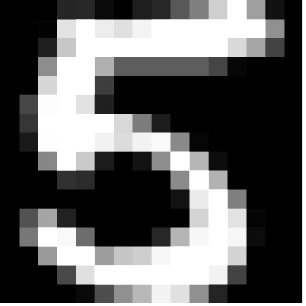} &
         \hspace{-0.4cm}
    \includegraphics[width=.022\textwidth]{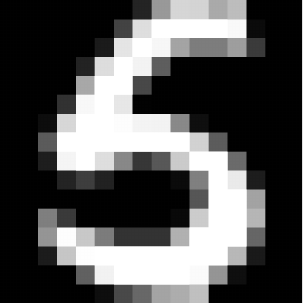} &
     \hspace{-0.4cm}
    \includegraphics[width=.022\textwidth]{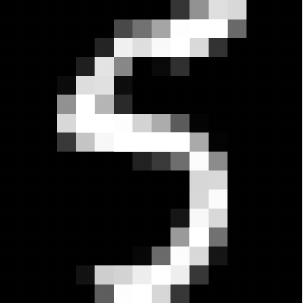} &
     \hspace{-0.4cm}
    \includegraphics[width=.022\textwidth]{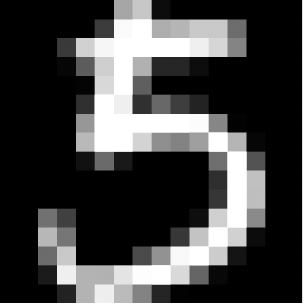}\\
      \includegraphics[width=.022\textwidth]{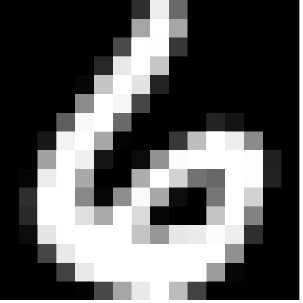} &
     \hspace{-0.4cm}
    \includegraphics[width=.022\textwidth]{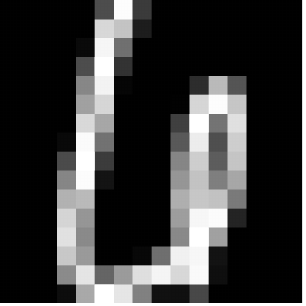} &
     \hspace{-0.4cm}
    \includegraphics[width=.022\textwidth]{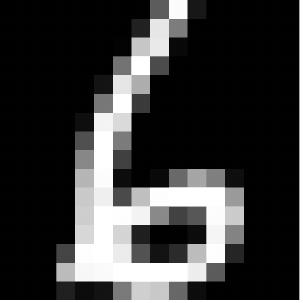} &
     \hspace{-0.4cm}
    \includegraphics[width=.022\textwidth]{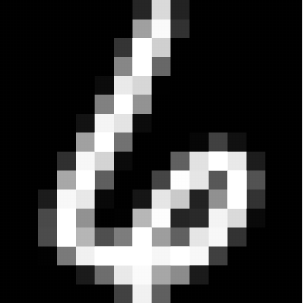} &
     \hspace{-0.4cm}
    \includegraphics[width=.022\textwidth]{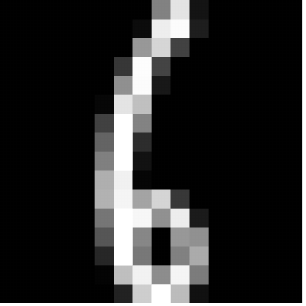} &
    \hspace{-0.4cm}
    \includegraphics[width=.022\textwidth]{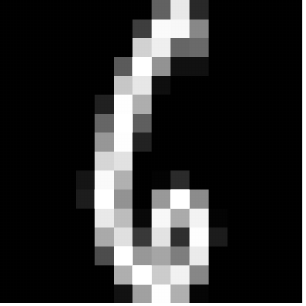} &
     \hspace{-0.4cm}
    \includegraphics[width=.022\textwidth]{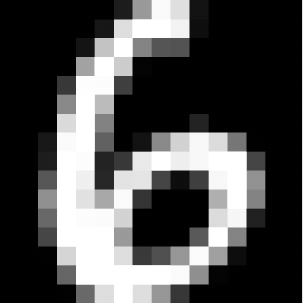} &
     \hspace{-0.4cm}
    \includegraphics[width=.022\textwidth]{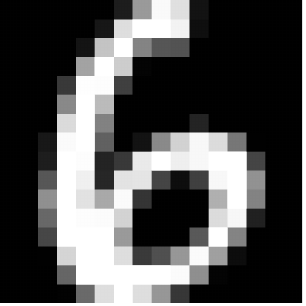}
    \\
      \includegraphics[width=.022\textwidth]{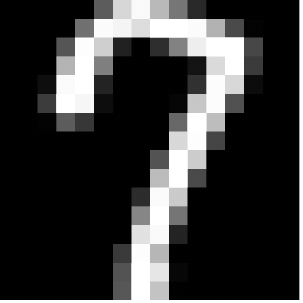} &
     \hspace{-0.4cm}
    \includegraphics[width=.022\textwidth]{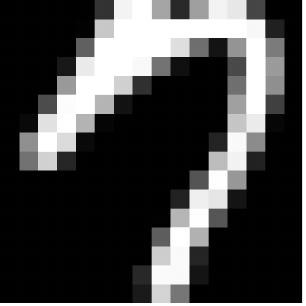} &
     \hspace{-0.4cm}
    \includegraphics[width=.022\textwidth]{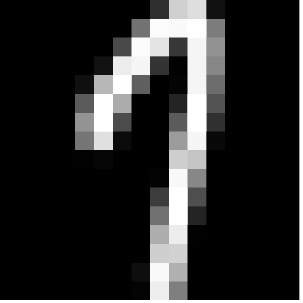} &
     \hspace{-0.4cm}
    \includegraphics[width=.022\textwidth]{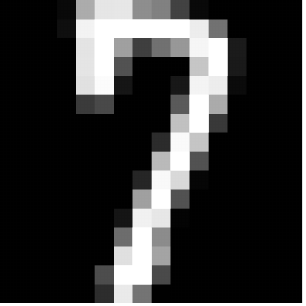} &
     \hspace{-0.4cm}
    \includegraphics[width=.022\textwidth]{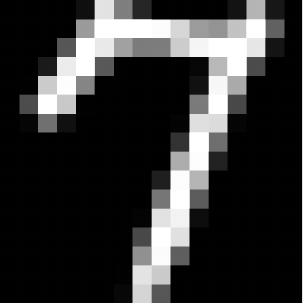} &
    \hspace{-0.4cm}
    \includegraphics[width=.022\textwidth]{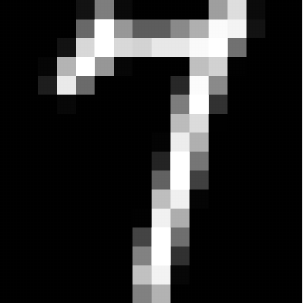} &
     \hspace{-0.4cm}
    \includegraphics[width=.022\textwidth]{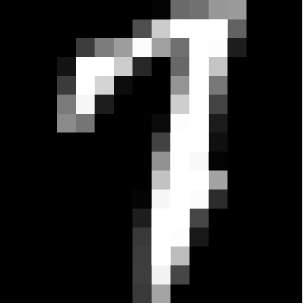} &
     \hspace{-0.4cm}
    \includegraphics[width=.022\textwidth]{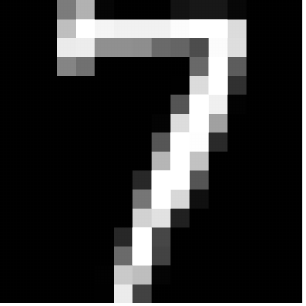} 
    \\
      \includegraphics[width=.022\textwidth]{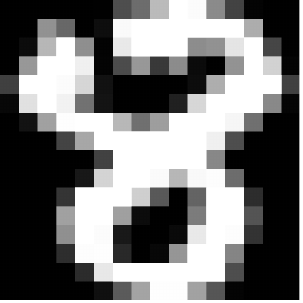} &
     \hspace{-0.4cm}
    \includegraphics[width=.022\textwidth]{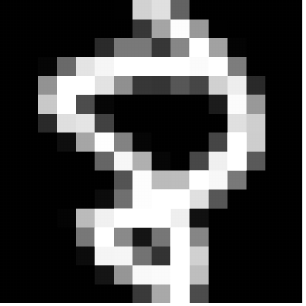} &
     \hspace{-0.4cm}
    \includegraphics[width=.022\textwidth]{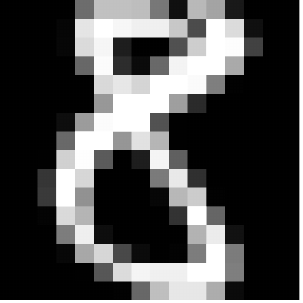} &
     \hspace{-0.4cm}
    \includegraphics[width=.022\textwidth]{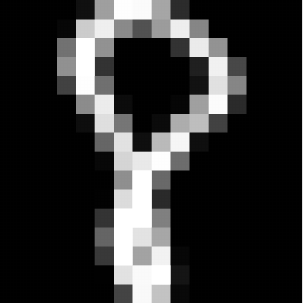} &
     \hspace{-0.4cm}
    \includegraphics[width=.022\textwidth]{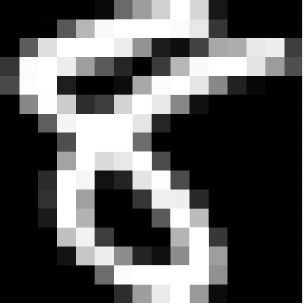} &
         \hspace{-0.4cm}
    \includegraphics[width=.022\textwidth]{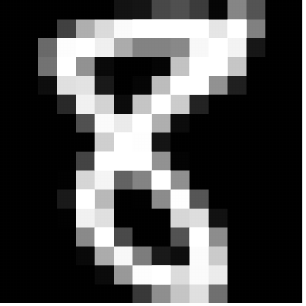} &
     \hspace{-0.4cm}
    \includegraphics[width=.022\textwidth]{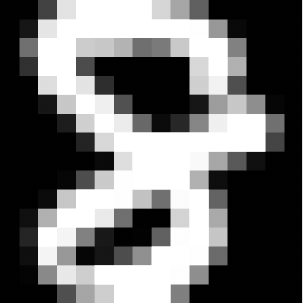} &
     \hspace{-0.4cm}
    \includegraphics[width=.022\textwidth]{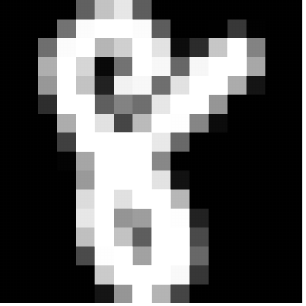}
    \\
    \includegraphics[width=.022\textwidth]{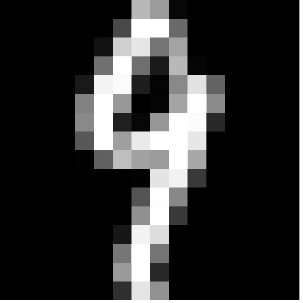} &
     \hspace{-0.4cm}
    \includegraphics[width=.022\textwidth]{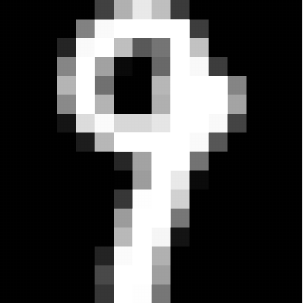} &
     \hspace{-0.4cm}
    \includegraphics[width=.022\textwidth]{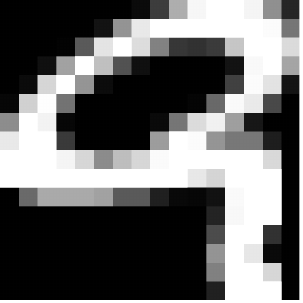} &
     \hspace{-0.4cm}
    \includegraphics[width=.022\textwidth]{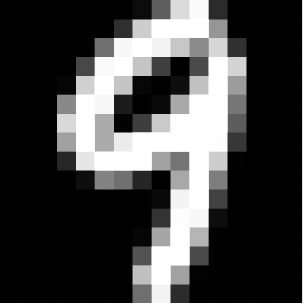} &
      \hspace{-0.4cm}
    \includegraphics[width=.022\textwidth]{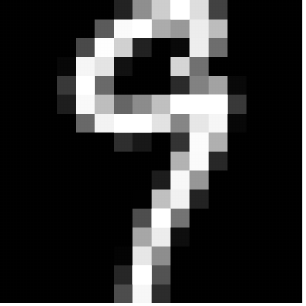} &
         \hspace{-0.4cm}
    \includegraphics[width=.022\textwidth]{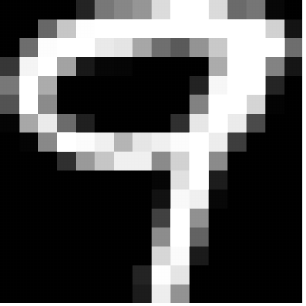} &
     \hspace{-0.4cm}
    \includegraphics[width=.022\textwidth]{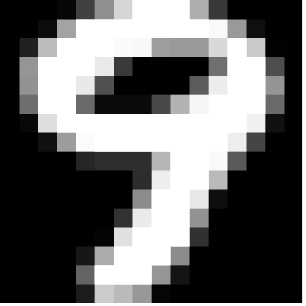} &
      \hspace{-0.4cm}
    \includegraphics[width=.022\textwidth]{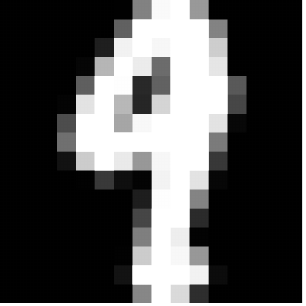}
  \end{tabular}
  }
\caption{The first eight columns correspond to class-conditional synthetic samples (generated by LRCF-DE) and the rest correspond to real samples from the USPS dataset.}
\label{fig:sampling_data}
\end{figure}

As our last experiment, we train LRCF-DE to learn the joint distribution of grayscale images from the USPS dataset \cite{lecun1990handwritten}, which contains $9298$ images of handwritten digits of size $16 \times 16 \rightarrow N=256$. The number of examples for each digit is shown in Table~\ref{table:sample-table}. We sample from the resulting $256$-dimensional model, and provide visualization of the generated data. We fix the tensor rank to $F=8$ and the smoothing parameter to $K=15$, and draw $8$ random samples of each digit (class). The resulting samples are shown in Figure \ref{fig:sampling_data}, and they are very pleasing -- in light of the fact that our model is ``agnostic'': designed for general-purpose density estimation, not specifically for realistic-looking image synthesis. It is possible to incorporate image modeling domain knowledge in the design of LRCF-DE (such as correlation between adjacent pixel values), but this is beyond the scope of this paper. 

\section{Conclusions}
In this work, we have revisited the classic problem of non-parametric density estimation from a fresh perspective -- through the lens of complex Fourier series approximation and tensor modeling, leading to a low-rank characteristic function approach. We showed that any compactly supported density can be well-approximated by a finite {\em characteristic tensor} of leading complex Fourier coefficients as long as the coefficients decay sufficiently fast. We posed density  estimation as a constrained (coupled) tensor factorization problem and proposed a Block Coordinate Descent algorithm, which under certain conditions enables learning the true data-generating distribution. Results on real data have demonstrated the utility and promise of this novel approach compared to both standard and recent density estimation techniques.

\begin{table}
\begin{center}
\resizebox{0.48\textwidth}{!}{
\begin{tabular}{lccccccccccc}
\toprule
 &\textbf{0}&\textbf{1}&\textbf{2}&\textbf{3}&\textbf{4}&\textbf{5}&\textbf{6}&\textbf{7}&\textbf{8}&\textbf{9}& Total\\
\midrule
Samples &$1553$&$1269$&$929$&$824$&$852$&$716$&$834$&$792$&$708$&$821$&$9298$\\
\bottomrule
\end{tabular}}
\end{center}
\caption{Images of handwritten digits - USPS dataset information.}
\label{table:sample-table}
\end{table}

\section{Appendix}
\label{SM}
\subsection{Bias, variance, consistency of the empirical characteristic function}

\noindent In this appendix we summarize important properties of empirical characteristic functions as statistical estimators of the corresponding characteristic functions. We refer the reader to \cite{ushakov2011selected} for proofs and additional results.  

By linearity of expectation, it is easy to see that the empirical characteristic function is an unbiased estimator of the corresponding characteristic function, i.e., \begin{equation*}
E\left[ {\widehat{\Phi}}_{{\boldsymbol X}}({\boldsymbol \nu}) \right] = \Phi_{{\boldsymbol X}}({\boldsymbol \nu}),
\end{equation*}
for all ${\boldsymbol \nu}$ and $M \geq 1$. For the remainder of this section, we assume that $\left\{ {\mathbf x}_m \right\}_{m=1}^M$ is i.i.d. in $m$. The variance of the  empirical characteristic function estimate can be shown \cite{ushakov2011selected} to be    
\begin{align*}
{\rm{Var}}[{\widehat{\Phi}}_{{\boldsymbol X}}({\boldsymbol \nu})]&=  E\left[{\left|{\widehat{\Phi}}_{{\boldsymbol X}}({\boldsymbol \nu}) -E\left[ {\widehat{\Phi}}_{{\boldsymbol X}}({\boldsymbol \nu}) \right]\right|}^2 \right]\\
&=  E\left[{\left|{\widehat{\Phi}}_{{\boldsymbol X}}({\boldsymbol \nu})-\Phi_{{\boldsymbol X}}({\boldsymbol \nu})\right|}^2 \right]\\
&= \frac{1}{M}\left(1-{\left|\Phi_{{\boldsymbol X}}({\boldsymbol \nu})\right|}^2\right).
\end{align*}
Note that $0\leq\big{|}{\Phi}_{{\boldsymbol X}}({\boldsymbol \nu})\big{|}\leq1$, and therefore ${\rm{Var}}[{\widehat{\Phi}}_{{\boldsymbol X}}({\boldsymbol \nu})]\leq \frac{1}{M}$. It follows that 
\begin{equation*}
\lim_{M \to \infty} E\left[{\left|{\widehat{\Phi}}_{{\boldsymbol X}}({\boldsymbol \nu})-\Phi_{{\boldsymbol X}}({\boldsymbol \nu})\right|}^2 \right] = 0,
\end{equation*}
i.e., for any fixed ${\boldsymbol \nu}$, ${\widehat{\Phi}}_{{\boldsymbol X}}({\boldsymbol \nu})$ converges to $\Phi_{{\boldsymbol X}}({\boldsymbol \nu})$ in the mean-squared sense. By the strong law of large numbers, it also converges almost surely for any fixed ${\boldsymbol \nu}$. Furthermore, for any fixed positive $T<\infty$
\begin{equation*}
\lim_{M \to \infty}\displaystyle\text{sup }_{\|\boldsymbol{{\nu}}\|_2 \leq  {T}}
\big{|}{\widehat{\Phi}}_{{\boldsymbol X}}({\boldsymbol \nu})-{\Phi}_{{\boldsymbol X}}({\boldsymbol \nu})\big{|}=0,
\end{equation*}
almost surely. It can also be shown \cite{ushakov2011selected} that for any increasing sequence $T_M$ such that $\lim_{M \to \infty} \frac{\log(T_M)}{M}=0$, it holds 
\begin{equation*}
\lim_{M \to \infty} \displaystyle\text{sup }_{\|\boldsymbol{{\nu}}\|_2 \leq  {T}_M}
\big{|}{\widehat{\Phi}}_{{\boldsymbol X}}({\boldsymbol \nu})-{\Phi}_{{\boldsymbol X}}({\boldsymbol \nu})\big{|} = 0, 
\end{equation*}
almost surely. In our context, we only use a sampled and truncated version of the characteristic function (corresponding to a truncated multivariate Fourier series), hence $T$ is always finite -- we do not need the latter result. 

It is also worth noting that the covariance of different samples of the empirical characteristic function (corresponding to different values of ${\boldsymbol \nu}$) goes to zero $\sim \frac{1}{M}$, and so does the covariance of its real and imaginary parts. As a result, for large $M$, the errors in the different elements of the characteristic tensor are approximately uncorrelated, with uncorrelated real and imaginary parts. This suggests that when we fit a model to the empirical characteristic function, it makes sense to use a least squares approach. Another motivation for this comes from Parseval's theorem: minimizing integrated squared error in the Fourier domain corresponds to minimizing integrated squared error between the corresponding multivariate distributions. This is true in particular when we limit the support of the distribution to a hypercube and use the samples of the characteristic function that correspond to the multivariate Fourier series, thereby replacing the multivariate integral in the Fourier domain by a multivariate sum. 

\subsection{Low-rank denoising: reduction of the mean squared error}

Our use of a low-rank model in the characteristic tensor domain is primarily motivated by the need to avoid the ``curse of dimensionality'': using a rank-$F$ model with $2K+1$ Fourier coefficients per mode parametrizes the whole $N$-dimensional multivariate density using just $FN(2K+1)$ coefficients, and avoids instantiating and storing a tensor of size $(2K+1)^N$, which is close to impossible even for moderate $N$. However, there is also a variance benefit that comes from this low-rank parametrization. We know from \cite{DonohoOptimalShrinkage} that for a square $L \times L$ matrix of rank $F$ observed in zero-mean white noise of variance $\sigma^2$, low-rank denoising attains mean squared error $c LF \sigma^2$ asymptotically in $L$, where $c$ is a small constant. In practice the asymptotics kick in even for relatively small $L$ \cite{DonohoOptimalShrinkage}. Contrast this to the raw $L^2 \sigma^2$ if one does not use the low-rank property. 

For an $N$-way tensor of rank $F$, assume $N$ is even, $K_n=K$, $\forall N$ (for simplicity of exposition), and ``unfold'' the characteristic tensor into a $K^{N/2} \times K^{N/2}$ matrix. In practice we use $F$ {\em far} less than $K^{N/2}$, and thus the resulting matrix will be {\em very} low rank. Invoking \cite{DonohoOptimalShrinkage}, low-rank tensor modeling will yield a reduction in mean squared error by a factor of at least $\frac{F}{K^{N/2}}$. We say at least, because this low-rank matrix structure is implied but does not imply low-rank tensor structure, which is much stronger. Note also that mean squared error in the characteristic tensor domain translates to mean squared error between the corresponding distributions, by virtue of Parseval's theorem.  

\subsection{Derivation of (\ref{reg})}
\noindent In this appendix we present the derivation of (\ref{reg}), which is used to solve  regression  tasks.
\begin{align*}
& E   \left[  X_N|X_1, \ldots, X_{N-1} \right] \nonumber \\
&= \int_{0}^{1} x_N {{f}}_{{X_N|X_1 ,\ldots, X_{N-1}}}({x_N|x_1, \ldots, x_{N-1}}) dx_N \nonumber \\
&= \int_{0}^{1} x_N \frac{{{f}}_{{X_1,\ldots,X_{N}}}({x_1,\ldots, x_{N}})}{{{f}}_{{X_1,\ldots,X_{N-1}}}({x_1,\ldots,x_{N-1}})} dx_N\nonumber \\
&= \frac{1}{c_1}\int_{0}^{1} x_N {{f}}_{{X_1,\ldots,X_{N}}}({x_1,\ldots,x_{N}}) dx_N  \nonumber \\
&= \frac{1}{c_1}\int_{0}^{1} x_N \sum_{h=1}^F \boldsymbol{\lambda}(h)\prod_{n=1}^N{\sum_{k_n=-K}^K}\mathbf{A}_n(k_n,h)e^{-j2\pi k_n x_n}dx_N  \nonumber \\
&= \frac{1}{c_1} \sum_{h=1}^F \boldsymbol{\lambda}(h)\prod_{n=1}^{N-1}{\sum_{k_n=-K}^K}\mathbf{A}_n(k_n,h)e^{-j2\pi k_n x_n}\nonumber \\
&\qquad\qquad\qquad\qquad {\sum_{k_N=-K}^K}\mathbf{A}_N(k_N,h)\int_{0}^{1} x_N e^{-j2\pi k_N x_N}dx_N  \nonumber \\
&= \frac{1}{c_1} \sum_{h=1}^F \boldsymbol{\lambda}(h)\prod_{n=1}^{N-1}{\sum_{k_n=-K}^K}\mathbf{A}_n(k_n,h)e^{-j2\pi k_n x_n}\nonumber \\
&\qquad\qquad\qquad\qquad\qquad\qquad {\sum_{k_N=-K}^K}c_{2,k_N}\mathbf{A}_N(k_N,h),  \nonumber \\
\end{align*}
where 
\begin{align*}
c_1&={{{f}}_{{X_1,\ldots,X_{N-1}}}({x_1,\ldots,x_{N-1}})} \nonumber \\
&=\sum_{h=1}^F \boldsymbol{\lambda}(h)\prod_{n=1}^{N-1}{\sum_{k_n=-K}^K}\mathbf{A}_n(k_n,h)e^{-j2\pi k_n x_n}, \nonumber \\
\end{align*}
and 
\begin{align*}
c_{2,k_N}&=\frac{e^{-j2\pi k_N}}{-j2\pi k_N}+\frac{1-e^{-j2\pi k_N}}{{[-j2\pi k_N}]^2}. \nonumber \\
\end{align*}

\bibliographystyle{IEEEtran}
\bibliography{references}

\end{document}